\RequirePackage{tikz}
\documentclass[sn-apa,iicol]{sn-jnl}
\jyear{2021}%

\usepackage{hhline}
\usepackage{cleveref}
\usepackage{natbib}
\usepackage{amsmath}
\usepackage{amsthm}

\newtheorem{lemma}{\textbf{Lemma}}
\usepackage{amssymb}
\usepackage{tabularx}
\usepackage[caption=false,font=footnotesize]{subfig}
\usepackage{hyperref}
\usepackage{url}
\usepackage{graphicx}
\usepackage{multirow}
\usepackage{siunitx}
\usepackage{diagbox}
\usepackage{pgfplots}
\pgfplotsset{width=7cm,compat=1.8}
\usepackage{tikz}
\usetikzlibrary{shapes,arrows}
\tikzstyle{block} = [rectangle, fill, fill=blue!10, 
    text width=12em, text centered, minimum height=2em]
\tikzstyle{empty} = [rectangle, fill, fill=white, 
     text centered]
\tikzstyle{line} = [draw, -latex']
\usetikzlibrary{calc}

\usepackage{soul}
\usepackage{cancel}

\newcommand{\normtwo}[1]{\left\lVert#1\right\rVert_2}

\newcommand{\removethis}[1]{}

\newcommand{\mR}{\mathcal{R}}
\newcommand{\mO}{\mathcal{O}}
\newcommand{\mH}{\mathcal{H}}
\newcommand{\mW}{\mathcal{W}}

\newcommand{\mV}{\mathcal{V}}

\newcommand{\mS}{\mathcal{S}}
\newcommand{\mP}{\mathcal{P}}

\newcommand{\mQ}{\mathcal{Q}}

\newcommand{\mF}{\mathcal{F}}
\newcommand{\mJ}{\mathcal{J}}

\newcommand{\vp}{\mathbf{p}}
\newcommand{\vpxy}[2]{\mathbf{p}_{#1}^{#2}}

\newcommand{\vf}{\mathbf{f}}
\newcommand{\vfx}[1]{\mathbf{f}_{#1}}
\newcommand{\vd}{\mathbf{d}}
\newcommand{\vdx}[1]{\mathbf{d}_{#1}}

\newcommand{\vn}{\mathbf{n}}

\newcommand{\vDelta}{\boldsymbol{\Delta}}

\newcommand{\vpi}{\boldsymbol{\pi}}
\newcommand{\vx}{\mathbf{x}}
\newcommand{\vq}{\mathbf{q}}
\newcommand{\vg}{\mathbf{g}}
\newcommand{\vy}{\mathbf{y}}

\newcommand{\vB}{\mathbf{B}}
\newcommand{\vD}{\mathbf{D}}
\newcommand{\vzero}{\mathbf{0}}
\newcommand{\vC}{\mathbf{C}}

\newcommand{\vA}{\mathbf{A}}

\newcommand{\vu}{\mathbf{u}}
\newcommand{\va}{\mathbf{a}}
\newcommand{\vb}{\mathbf{b}} 

\newcommand{\ve}{\mathbf{e}}

\newcommand{\vtheta}{\boldsymbol{\theta}}
\newcommand{\vlambda}{\boldsymbol{\lambda}}
\newcommand{\vPx}[1]{\mathbf{P}_{#1}}
\newcommand{\vPxy}[2]{\mathbf{P}_{#1, #2}}

\makeatletter
\def\Cline#1#2{\@Cline#1#2\@nil}
\def\@Cline#1-#2#3\@nil{%
  \omit
  \@multicnt#1%
  \advance\@multispan\m@ne
  \ifnum\@multicnt=\@ne\@firstofone{&\omit}\fi
  \@multicnt#2%
  \advance\@multicnt-#1%
  \advance\@multispan\@ne
  \leaders\hrule\@height#3\hfill
  \cr}
\makeatother

\newtheoremstyle{named}{}{}{\itshape}{}{\bfseries}{.}{.5em}{\thmnote{#3's }#1}

\newtheorem{remark}{Remark}

\usepackage{array}
\newcolumntype{L}{>{\arraybackslash}m{3cm}}

\usepackage[algo2e,linesnumbered]{algorithm2e}

\SetKwInOut{Parameter}{Parameter}
\SetKwInOut{TaskInput}{TaskInput}
\SetKwInOut{Returns}{Return}
\SetKwInOut{Input}{Input}

\begin{document}

\title[RLSS]{RLSS: Real-time, Decentralized, Cooperative, Networkless Multi-Robot Trajectory Planning using Linear Spatial Separations}


\author*[1]{\fnm{Bask{\i}n} \sur{\c{S}enba\c{s}lar}}\email{baskin.senbaslar@usc.edu}

\author[2]{\fnm{Wolfgang} \sur{H\"{o}nig}}\email{hoenig@tu-berlin.de}

\author[3]{\fnm{Nora} \sur{Ayanian}}\email{nora\_ayanian@brown.edu}

\affil*[1]{\orgdiv{Department of Computer Science}, \orgname{University of Southern California}, \orgaddress{\street{941 Bloom Walk}, \city{Los Angeles}, \postcode{90089}, \state{CA}, \country{USA}}}

\affil[2]{\orgdiv{Department of Electrical Engineering and Computer Science}, \orgname{Technical University of Berlin}, \orgaddress{\street{Marchstr. 23}, \city{Berlin}, \postcode{10587}, \country{Germany}}}

\affil[3]{\orgdiv{Department of Computer Science and School of Engineering}, \orgname{Brown University}, \orgaddress{\street{115 Waterman Street}, \city{Providence}, \postcode{02912}, \state{RI}, \country{USA}}}

\abstract{Trajectory planning for multiple robots in shared environments is a challenging problem especially when there is limited communication available or no central entity.
In this article, we present {R}eal-time planning using {L}inear {S}patial {S}eparations, or RLSS%
%RLSS (short for \underline{R}eal-time Planning using \underline{L}inear \underline{S}patial \underline{S}eparations)
: a real-time decentralized trajectory planning algorithm for cooperative multi-robot teams in static environments.
%RLSS accounts for the robots' dynamic limits, is decentralized, requires sensing the positions of robots and obstacles only, and requires no communication between robots. 
The algorithm requires relatively few robot capabilities, namely sensing the positions of robots and obstacles without higher-order derivatives and the ability of distinguishing robots from obstacles. There is no communication requirement and the robots' dynamic limits are taken into account.
RLSS generates and solves convex quadratic optimization problems that are kinematically feasible and guarantees collision avoidance if the resulting problems are feasible.
% The major components of RLSS are: selecting a goal position, finding a discrete path towards the goal position, and smoothing a trajectory on top of the path by solving a kinematically feasible convex quadratic program.
We demonstrate the algorithm's performance in real-time in simulations and on physical robots. 
We compare RLSS to two state-of-the-art planners and show empirically that RLSS does avoid deadlocks and collisions in forest-like and maze-like environments, significantly improving prior work, which result in collisions and deadlocks in such environments.
}


\maketitle

\section*{Acknowledgment}
This work was supported by National Science Foundation awards IIS-1724399, IIS-1724392, and CPS-1837779. B. \c{S}enba\c{s}lar was supported by a University of Southern California Annenberg Fellowship.
Wolfgang Hönig was partially funded by the Deutsche Forschungsgemeinschaft (DFG, German Research Foundation) - 448549715.

\section{Introduction}

Effective navigation of multiple robots 
in cluttered environments %%%
is key to emerging industries such as warehouse automation~\citep{kiva}, autonomous driving~\citep{autdrive}, and automated intersection management~\citep{intman}.
One of the core challenges of navigation systems in such domains is trajectory planning.
Planning safe trajectories for multiple robots is especially challenging when there is no central entity that plans all robots' trajectories a priori or re-plans mid-execution if there is a fault.
In some cases, such a central entity is undesirable because of the communication link that must be maintained between each robot and the central entity. If the map is not known a priori, building and relaying the observed map of the environment to the central entity through the communication channel adds further challenges.
In some cases, it is impractical to have such a central entity because it cannot react to robot trajectory tracking errors and map updates fast enough due to communication and computation delays.
In practice, robots must operate safely even if there is limited communication available.
This necessitates the ability for robots to plan in a decentralized fashion, where each robot plans a safe trajectory for itself while operating in environments with other robots and obstacles.

\begin{figure}
    \centering
    \includegraphics[width=\linewidth]{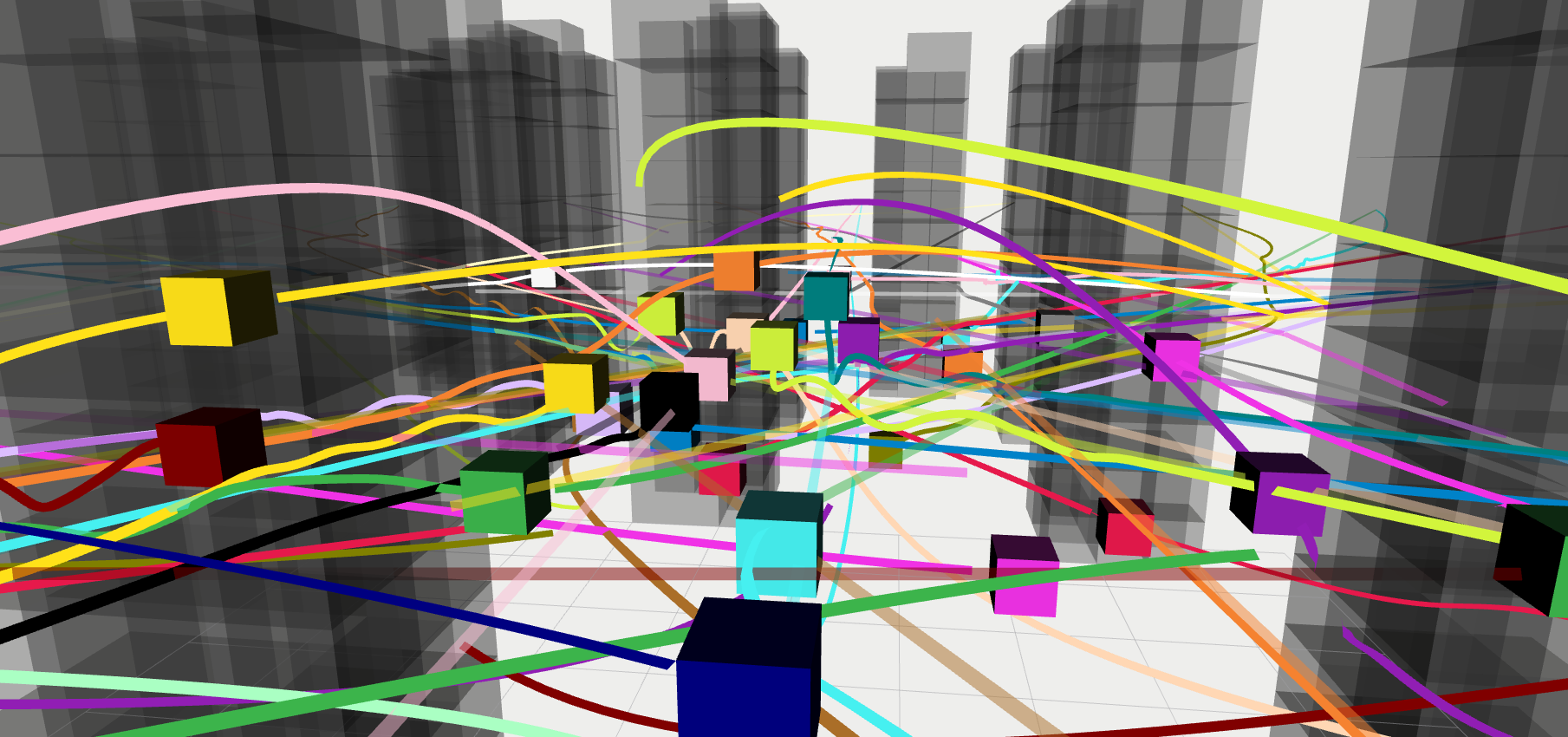}
    \caption{RLSS runs in real-time in dense environments. Each robot plans a trajectory by itself using only the position information of other robots and obstacles.}
    \label{Figure:RLSS}
\end{figure}

Decentralized algorithms delegate the computation of trajectories:
each robot plans for itself and reacts to the environment by itself.
In this paper, we introduce RLSS, a real-time decentralized trajectory planning algorithm for multiple robots in a shared environment that requires no communication between robots and requires relatively few sensing capabilities: each robot senses the relative positions of other robots and obstacles along with their geometric shapes in the environment, and is able to distinguish robots from obstacles.
RLSS requires relatively few robot capabilities than most state-of-the-art decentralized planners used for multi-robot navigation, which typically require communication~\citep{luis2020dmpc, tordesillas2020mader, wang2021dpmc}, higher order derivative estimates~\citep{park2021rsfc, wang2021dpmc}, or predicted trajectories of objects~\citep{park2021rsfc}.
However, the ability to distinguish robots from obstacles is not required by some state-of-the-art algorithms~\citep{zhou2017bvc}, which require modelling obstacles as robots.
RLSS is cooperative in the sense that we assume each robot stays within its current cell of a tessellation of the space until the next planning iteration.
We assume obstacles are static, which is required to guarantee collision avoidance.
%With its minimal requirements, our algorithm is a new baseline for algorithms that require communication, higher order derivative estimates, or predicted trajectories of objects.

RLSS explicitly accounts for the dynamic limits of the robots and enforces safety with hard constraints, reporting when they cannot be enforced, thus guaranteeing collision avoidance when it succeeds.
The planning algorithm can be guided with desired trajectories, thus it can be used in conjunction with centralized planners.
If no centralized planner or plan is available, RLSS can be used on its own, without any central guidance, by setting the desired trajectories to line segments directly connecting robots' start positions to their goal positions.

There are $4$ stages in RLSS. \begin{enumerate}
\item Select a goal position to plan toward on the desired trajectory;
\item Plan a discrete path toward the selected goal position; 
\item Formulate and solve a kinematically feasible convex optimization problem to compute a safe smooth trajectory guided by the discrete plan;
\item Check if the trajectory obeys the dynamic limits of the robot, and temporally rescale the trajectory if not.
\end{enumerate}
RLSS works in the receding horizon fashion: it plans a long trajectory, executes it for a short duration, and replans at the next iteration.
It utilizes separating hyperplanes, i.e., linear spatial separations, between robot shapes, obstacles, and sweep geometries (the subsets of space swept by robots while traversing straight line segments) to enforce safety during trajectory optimization.

We demonstrate, through simulation and experiments on physical robots, that RLSS works in dense environments in real-time (Fig.~\ref{Figure:RLSS}).
We compare our approach to two state-of-the-art receding horizon decentralized multi-robot trajectory planning algorithms in 3D.
In the first, introduced by~\cite{zhou2017bvc}, robots plan for actions using a model predictive control-style optimization formulation while enforcing that each robot stays inside its buffered Voronoi cell at every iteration.
We refer to this method as BVC.
The original BVC formulation works only for discrete single-integrator dynamics and environments without obstacles.
We extend the BVC formulation to discrete time-invariant linear systems with position outputs and environments with obstacles, and call the extended version eBVC (short for extended BVC).
RLSS and eBVC have similar properties: they require no communication between robots, and require position sensing of other objects in the environment.
However, unlike RLSS, eBVC does not require that robots are able to distinguish robots from obstacles, as it treats each obstacle as a robot.
In the second, introduced by~\cite{park2021rsfc}, robots plan for trajectories by utilizing relative safe navigation corridors, which they execute for a short duration, and replan.
We refer to this method as RSFC.
RSFC requires no communication between robots, and utilizes positions as well as velocities of the objects in the environment, thus requires more sensing capabilities than RLSS.
We demonstrate empirically that RLSS results in no deadlocks or collisions in our experiments in forest-like and maze-like environments, while eBVC is prone to deadlocks and RSFC results in collisions in such environments.
However, RLSS results in slightly longer navigation durations compared to both eBVC and RSFC.

The contribution of this work can be summarized as follows:
\begin{itemize}
    \item A carefully designed, numerically stable, and effective real-time decentralized planning algorithm for multiple robots in shared environments with static obstacles with relatively few requirements: no communication between robots, position-only sensing of robots and obstacles, and ability to distinguish robots from obstacles.
    \item An extension (eBVC) of a baseline planner (BVC) to environments with obstacles and a richer set of dynamics than only single-integrators.
    \item The first comparison of more than two state-of-the-art communication-free decentralized multi-robot trajectory planning algorithms, namely, RSFC, eBVC, and RLSS, in complicated forest-like and maze-like environments, some of which containing more than $2000$ obstacles.
\end{itemize}

\section{Related Work}
The pipeline of RLSS contains three stages (discrete planning, safe navigation corridor construction, and trajectory optimization) that are employed by several existing single-robot trajectory planning algorithms.
~\cite{richter2013planning} present a single-robot trajectory planning method for aggressive quadrotor flight which utilizes RRT*~\citep{karaman2010rrtstar} to find a kinematically feasible discrete path and formulates an unconstrained quadratic program over polynomial splines guided by the discrete path.
Collisions are checked after optimization, and additional decision points are added and optimization is re-run if there is collision.
~\cite{chen2016planning} present a method that utilizes OcTree representation~\citep{hornung2013octomap} of the environment during discrete search.
They find a discrete path using unoccupied grid cells of the OcTree.
Then, they maximally inflate unoccupied grid cells to create a safe navigation corridor that they use as constraints in the subsequent polynomial spline optimization.
~\cite{liu2017planning} uses Jump Point Search (JPS)~\citep{harabor2011jps} as the discrete planner, and construct safe navigation corridors that are used as constraints in the optimization stage.
Our planning system  uses these three stages (discrete planning, safe navigation corridor construction, and corridor-constrained  optimization) and extends it to multi-robot scenarios in a decentralized way.
We handle robot-robot safety by cooperatively computing a linear partitioning of the environment, and enforcing that each robot stays within its own cell of the partition.

%Here, we discuss different approaches for solving the multiple robot trajectory planning problem.
We categorize multi-robot trajectory planning algorithms first on where computation is done, since the location of computation changes the properties of the algorithms.
There are two main strategies to solve the multi-robot trajectory planning problem: centralized and decentralized.
Centralized algorithms compute trajectories for all robots on a central entity with global information; 
these trajectories are communicated back to the robots for them to execute. 
In the decentralized strategy, each robot runs an algorithm on-board to compute its own trajectory; 
these algorithms may utilize direct communication between robots.
Centralized algorithms often provide strong theoretical guarantees on completeness or global optimality because i) they have the complete knowledge about the environment beforehand, ii) they generally run on powerful centralized computers, and iii) there are generally less restrictive time limits on their execution.
Decentralized algorithms generally forgo such strong theoretical guarantees in favor of fast computation because i) they are often used when complete information of the environment is not known beforehand, and ii) they have to work in real-time on-board.

\subsection{Centralized Algorithms}\label{Section:CentralizedAlgorithms}
We further categorize centralized algorithms into those that do not consider robot dynamics during planning and those that do.

\textbf{Centralized without Dynamics:}
If planning can be abstracted to agents moving along edges in a graph synchronously, we refer to the multi-agent path finding (MAPF) problem.
Some variants of the MAPF problem are NP-Hard to solve optimally~\citep{MRPP2013yu}.
For NP-Hard variants, there are many optimal~\citep{sharon2015cbs, bcp2019stuckey}, bounded suboptimal~\citep{barer2014ecbs}, and suboptimal~\citep{solovey2013drrt, ma2019prioritized} algorithms that perform well in some environments.
Trajectories generated by planning algorithms that do not model robot dynamics may not be followed perfectly by real robots, resulting in divergence from the plan.

\textbf{Centralized with Dynamics:}
These algorithms deliver smooth control or output trajectories that are executable by real robots. 
Trajectories are usually generated by formulating the problem under an optimization framework, and communicated to the robots for them to execute. 
\cite{tang2016safe} propose an approach that combines a discrete motion planning algorithm with trajectory optimization that works only in obstacle-free environments. 
Another approach combines a MAPF solver with trajectory optimization to plan trajectories for hundreds of quadrotors in environments with obstacles, while ensuring that the resulting trajectories are executable by the robots~\citep{honig2018quadswarms}.
\cite{park2020relbern} combine a MAPF solver with trajectory optimization and provide executability and feasibility guarantees.
\cite{desai2020IGT} propose an approach utilizing position-invariant geometric trees to find kinodynamically feasible trajectories. 

Centralized algorithms can provide theoretical completeness and global optimality guarantees under the assumptions of perfect prior knowledge and execution of trajectories.
However, centralized replanning is required when the environment is discovered during operation or the robots deviate from the planned trajectories.
This necessitates continuous communication between the central computer and the individual robots.
In some cases, solving the initial problem from scratch may be required.
If the communication between the robots and the central computer cannot be maintained, or solving the initial problem from scratch cannot be done in real-time, the robots may collide with each other or obstacles.
% Furthermore, centralized algorithms require a prior map of the environment, which is not always available.

\subsection{Decentralized Algorithms}
When the requirements of the centralized strategies cannot be satisfied, decentralization of the trajectory planning among robots is required. In decentralized planning, robots assess the state of the environment and plan for themselves using their on-board capabilities. 
We categorize decentralized algorithms into reactive and long horizon algorithms.

\textbf{Decentralized Reactive:} In reactive algorithms, each robot computes the next action to execute based on the state of the environment without considering the actions that might follow it.
One such approach is Optimal Reciprocal Collision Avoidance (ORCA)~\citep{alonsomora2013orca}. 
In ORCA, each robot chooses a velocity vector that is as close as possible to a desired velocity vector such that collisions between each pair of robots are provably avoided.
It requires robots to i) be cooperatively executing the same algorithm, ii) obey single integrator dynamics, and iii) either sense other robots' velocities or receive them through communication. 
While ORCA handles dynamic changes in the environment and is completely distributed, it fails to avoid deadlocks in environments with obstacles~\citep{rte-dars2018}.
Safety barrier certificates guarantee collision avoidance by computing a safe control action that is as close as possible to a desired control action~\citep{wang2017safety}. 
Similarly to ORCA, the robots using safety barrier certificates may get stuck in deadlocks. 
Most learning-based approaches fall into the body of reactive strategies as well.
Here, neural networks are trained to compute the next action to execute, given a robot's immediate neighborhood and goal\footnote{Note that some learning-based approaches train neural networks to mimic the behavior of a global long horizon planner. Because of this, even if they output a single action, they abstractly consider a longer horizon evolution of the system. We place these approaches to the reactive algorithms category because they output a single action at each iteration.}.
For instance, PRIMAL~\citep{sartoretti2019primal} is a computationally efficient suboptimal online MAPF solver, but it only works on grids, and does not consider robot dynamics.
GLAS~\citep{riviere2020glas}, which considers robot dynamics, can guarantee safety for some robot dynamics by combining the network output with a safety term.
It performs better than ORCA in terms of deadlocks, nevertheless results in deadlocks in dense environments.
Another approach employs graph neural networks (GNNs) that allow communication between robots to avoid collisions~\citep{li2020gnn}, but results in deadlocks in dense environments similar to GLAS.
~\cite{nn2021batra} propose a method that outputs direct motor commands for a quadrotor using the positions and velocities of objects in the its immediate neighborhood, but results in collisions between robots.

\textbf{Decentralized Long Horizon:} In long horizon algorithms, robots generate a sequence of actions or long trajectories instead of computing a single action to execute.
These approaches employ receding horizon planning: they plan long trajectories, execute them for a short duration, and replan.
\cite{zhou2017bvc} present a model predictive control (MPC) based approach in which robots plan for a sequence of actions while enforcing each robot to stay within its buffered Voronoi cell.
It requires no inter-robot communication and depends on sensing of other robots' positions only. 
Another MPC-based approach approximates robots' controller behaviors under given desired states as a linear system~\citep{luis2020dmpc}.
A smooth B\'ezier curve for each robot is computed by solving an optimization problem, in which 
%In the optimization problem, 
samples of desired states are drawn from the curve and fed into the model of the system.
The approach requires communication of future states for collision avoidance.
Another MPC-based approach %\cite{wang2021dpmc} presents another MPC-based approach.
generates plans using motion primitives to compute time-optimal trajectories, then trains a neural network to approximate the behavior of the planner, which is used during operation for fast planning~\citep{wang2021dpmc}.
It requires sensing or communicating the full states of robots.
\cite{tordesillas2020mader} present a method for dynamic obstacle avoidance, handling asynchronous planning between agents in a principled way.
However, the approach requires instant communication of planned trajectories between robots to ensure safety.
\cite{park2021rsfc} plan a piecewise B\'ezier curve by formulating an optimization problem.
It utilizes relative safe flight corridors (RSFCs) for collision avoidance and requires no communication between robots, but requires position and velocity sensing.
In \cite{peterson2021TWTL++}'s approach, robot tasks are defined as time window temporal logic specifications.
Robots plan trajectories to achieve their tasks while avoiding each other.
The method provably avoids deadlocks, but requires communication of planned trajectories between robots.

RLSS falls into this body of algorithms.
It requires sensing only the positions of other robots and obstacles, and does not require any communication between robots.
The work of~\cite{zhou2017bvc} (BVC) is the only other approach with these properties.
Different from their approach, RLSS uses discrete planning to avoid local minima, plans piecewise B\'ezier trajectories instead of MPC-style input planning, and does not constrain the full trajectory to be inside a constraining polytope. These result in better deadlock and collision avoidance as we show in the evaluation section.

In terms of inter-robot collision avoidance, RLSS belongs to a family of algorithms that ensure multi-robot collision avoidance without inter-robot communication by using the fact that each robot in the team runs the same algorithm.
In these algorithms, robots share the responsibility of collision avoidance by computing feasible action sets that would not result in collisions with others when others compute their own feasible action sets using the same algorithm.
Examples include BVC~\citep{zhou2017bvc}, which partitions position space between robots and makes sure that each robot stays within its own cell in the partition, ORCA~\citep{alonsomora2013orca}, which computes feasible velocity sets for robots such that there can be no collisions between robots as long as each robot chooses a velocity command from their corresponding sets, and SBC~\citep{wang2017safety}, which does the same with accelerations.

BVC and RLSS also belong to a family of decentralized multi-robot trajectory planning algorithms that utilize mutually computable separating hyperplanes for inter-robot safety as defined by~\cite{senbaslar2022async}.
These algorithms do not require any inter-robot communication and utilize only position/geometry sensing between robots.
The inter-robot safety is enforced by making sure that each pair of robots can compute the \emph{same separating hyperplane} and enforce themselves to be in the different sides of this hyperplane at every planning iteration.
Separating hyperplanes that can be mutually computed by robots using only position/geometry sensing and no communication are defined as mutually computable.
BVC uses Voronoi hyperplanes and RLSS uses support vector machine hyperplanes, both of which are mutually computable.

RLSS is an extension to our previous conference paper~\citep{rte-dars2018}. We extend it conceptually by
\begin{itemize}
    \item supporting robots with any convex geometry instead of only spherical robots; and
    \item decreasing the failure rate of the algorithm from $3\%$ in complex environments to $0.01\%$ by providing important modifications to the algorithm. These modifications include i) changing the discrete planning to best-effort $A^*$ search, thus allowing robots to plan towards their goals even when goals are not currently reachable; ii) increasing the numerical stability of the algorithm by running discrete planning at each iteration instead of using the trajectory of the previous iteration when it is collision free to define the homotopy class, and adding a \emph{preferred distance cost term} to the optimization problem in order to create a safety zone between objects when possible; and iii) ensuring the kinematic feasibility of the optimization problem generated at the trajectory optimization stage.
\end{itemize}
We extend our previous work empirically by
\begin{itemize}
    \item applying our algorithm to a heterogeneous team of differential drive robots in 2D;
    \item applying our algorithm to quadrotors in 3D; and
    \item comparing the performance of our algorithm to stronger baselines.
\end{itemize}

A preliminary version of RLSS previously appeared in a workshop~\citep{senbaslar2021rlss}.
In the workshop version, the RLSS optimization stage fails about $\SI{3}{\%}$ of the time, and switches to a soft optimization formulation when that happens.
In the current version, the optimization %formulation
rarely (close to $1$ in every $10000$ iterations in our experiments) fails thanks to the newly added preferred distance cost term, changed goal selection stage, and changed discrete planning stage that ensures kinematic feasibility of the optimization problem.
Hence, there is no longer a different soft optimization formulation.
In the workshop version, we compare RLSS against \cite{luis2020dmpc}'s work, which is based on distributed MPC, which requires communication of planned trajectories between robots.
Herein, we compare our approach to two planners that do not require communication between robots as this provides a more comparable baseline.

\section{Problem Statement}\label{Section:ProblemStatement}

We first define the multi-robot trajectory planning problem,
then, we indicate the specific case of the problem we solve.

Consider a team of $N$ robots for which trajectories must be computed at time $t=0$. The team can be heterogeneous, i.e., each robot might be of a different type or shape. 
Let $\mR_i$ be the collision shape function of robot $i$ such that $\mR_i(\vp)$ is the collision shape of robot $i$ located at position $\vp \in \mathbb R^d, d\in\{2,3\}$, i.e., the subset of space occupied by robot $i$ when placed at position $\vp$.
We define $\mR_i(\vp) = \{\vp + \vx \mid \vx \in \mR_{i,0}\}$ where $\mR_{i,0} \subset \mathbb{R}^d$ is the the space occupied by robot $i$ when placed at the origin.
Note that we do not model robot orientation. If a robot can rotate, the collision shape should contain the union of spaces occupied by the robot for each possible orientation at a given position; which is minimally a hypersphere if all orientations are possible.

We assume the robots are differentially flat~\citep{murray1995differential}, i.e., the robots' states and inputs can be expressed in terms of output trajectories and a finite number of derivatives thereof.
Differential flatness is common for many kinds of mobile robots, including differential drive robots~\citep{campion1996structural}, car-like robots~\citep{murray1993CarLike}, omnidirectional robots~\citep{jiang2013omnidirectional}, and quadrotors~\citep{mellinger2011snap}.
When robots are differentially-flat, their dynamics can be accounted for by i) imposing $C^c$ continuity on the trajectories for the required $c$ (i.e. continuity up to $c^{th}$ degree of derivative), and ii) imposing constraints on the maximum $k^{th}$ derivative magnitude of trajectories for any required $k$.
For example, quadrotor states and inputs can be expressed in terms of the output trajectory, and its first (velocity), second (acceleration) and third (jerk) derivatives~\citep{mellinger2011snap}.
Hence, to account for quadrotor dynamics, continuity up to jerk should be enforced by setting $c=3$, and velocity, acceleration, and jerk of the output trajectory should be limited appropriately by setting upper bounds for $k=1$, $2$, and $3$.

Let $\mO(t) = \{\mQ \subset \mathbb{R}^d\}$ be the set of obstacles in the environment at time $t$. 
The union $\mQ(t) = \cup_{\mQ \in \mO(t)}\mQ$ denotes the occupied space at time $t$.
Let $\mW_i \subseteq \mathbb R^d$ be the workspace that robot $i$ should stay inside of.

Let $\vdx{i}(t): [0, T_i] \to \mathbb R^d$ be the desired trajectory of duration $T_i$ that robot $i$ should follow as closely as possible. 
%is the duration of the desired trajectory.
Define $\vdx{i}(t) = \vdx{i}(T_i)\ \forall t \geq T_i$.
We do not require that this trajectory is collision-free or even dynamically feasible to track by the robot. 
% However, a better trajectory should yield better results in terms of energy consumption or time until the goal is reached.
If no such desired trajectory is known, it can be initialized, for example, with a straight line to a goal location.

The intent is that each robot $i$ tracks a Euclidean trajectory $\vfx{i}(t) : [0, T]\to \mathbb R^d$ such that $\vfx{i}(t)$ is collision-free, executable according to the robot's dynamics, is as close as possible to the desired trajectory $\vdx{i}(t)$, and ends at $\vdx{i}(T_i)$.
Here, $T$ is the navigation duration of the team.
We generically define the multi-robot trajectory planning problem as the problem of finding trajectories $\vfx{1}, \ldots, \vfx{N}$ and navigation duration $T$ that optimizes the following formulation:
% \begin{align}
%     &\min_{\vfx{1}, \ldots, \vfx{N}, T} \sum_{i=1}^{N}\int_{0}^{T} \normtwo{\vfx{i}(t) - \vdx{i}(t)} dt \text{ s.t. }\label{Formulation:Cost} \\
%     &\ \ \ \vfx{i}(t) \in C^{c_i} & &\mkern-98mu\forall i\in\{1,\ldots,N\}\label{Formulation:ContinuityConstraint}\\
%     &\ \ \ \vfx{i}(T) = \vdx{i}(T_i) & &\mkern-98mu\forall i \in\{1,\ldots,N\}\label{Formulation:EndPointConstraint}\\
%     &\ \ \ \frac{d^c\vfx{i}(0)}{dt^c} = \frac{d^c\vpxy{i}{0}}{dt^c} & &\mkern-113mu\forall i \forall c\in\{0, \ldots, c_i\}\label{Formulation:InitialPointConstrant}\\
%     &\ \ \ \mR_i(\vfx{i}(t)) \cap \mQ(t) = \emptyset & &\mkern-71mu\forall i ,\forall t\in[0, T]\label{Formulation:ObstacleAvoidance}\\
%     &\ \ \ \mR_i(\vfx{i}(t)) \cap \mR_j(\vfx{j}(t))\!=\! \emptyset & &\mkern-109mu\forall j\!\neq\! i,\forall t\in[0,T]\label{Formulation:RobotAvoidance}\\
%     &\ \ \ \mR_i(\vfx{i}(t)) \in \mW_i & &\mkern-71mu\forall i\forall t\in[0, T]\label{Formulation:Workspace}\\
%     &\ \ \ \underset{t \in [0, T]}{\max}\normtwo{\frac{d^k \vfx{i}(t)}{dt^k}} \leq  \gamma_i^k & &\mkern-123mu\forall i,\forall k \in \{1, \ldots, K_i\} \label{Formulation:DynamicLimits}
% \end{align}
\begin{align}
    \rlap{$\displaystyle \min_{\vfx{1}, \ldots, \vfx{N}, T} \sum_{i=1}^{N}\int_{0}^{T} \normtwo{\vfx{i}(t) - \vdx{i}(t)}dt$ s.t.}\label{Formulation:Cost} \\
    &\vfx{i}(t) \in C^{c_i} &&\forall i\in\{1,\ldots,N\}\label{Formulation:ContinuityConstraint}\\
    &\vfx{i}(T) = \vdx{i}(T_i) &&\forall i \in\{1,\ldots,N\}\label{Formulation:EndPointConstraint}\\
    &\frac{d^c\vfx{i}(0)}{dt^c} = \frac{d^c\vpxy{i}{0}}{dt^c} &&\forall i \forall c\in\{0, \ldots, c_i\}\label{Formulation:InitialPointConstrant}\\
    &\mR_i(\vfx{i}(t)) \cap \mQ(t) = \emptyset &&\forall i\forall t\in[0, T]\label{Formulation:ObstacleAvoidance}\\
    &\mR_i(\vfx{i}(t)) \cap \mR_j(\vfx{j}(t))\!=\! \emptyset &&\forall j\!\neq\! i\forall t\in[0,T]\label{Formulation:RobotAvoidance}\\
    &\mR_i(\vfx{i}(t)) \in \mW_i && \forall i\forall t\in[0, T]\label{Formulation:Workspace}\\
    &\underset{t \in [0, T]}{\max}\normtwo{\frac{d^k \vfx{i}(t)}{dt^k}}\!\leq\!\gamma_i^k &&\forall i\forall k \in \{1, \ldots, K_i\} \label{Formulation:DynamicLimits}
\end{align}
where $c_i$ is the order of derivative up to which the trajectory of the $i^{th}$ robot must be continuous, $\vpxy{i}{0}$ is the initial position of robot $i$ (derivatives of which are the initial higher order state components, e.g., velocity, acceleration, etc.), $\gamma_i^k$ is the maximum $k^{th}$ derivative magnitude that the $i^{th}$ robot can execute, and $K_i$ is the maximum derivative degree that robot $i$ has a derivative magnitude limit on.
%With these, each robot tries to come up with a safe trajectory $\boldsymbol{f_i}(t)$ that is as close as possible to the original trajectory $\boldsymbol{o_i}(t)$.

The cost \eqref{Formulation:Cost} of the optimization problem is a metric for the total deviation from the desired trajectories; it is the sum of position distances between the planned and the desired trajectories.
\eqref{Formulation:ContinuityConstraint} enforces that the trajectory of each robot is continuous up to the required degree of derivatives. \eqref{Formulation:EndPointConstraint} enforces that planned trajectories end at the endpoints of desired trajectories.
\eqref{Formulation:InitialPointConstrant} enforces that the planned trajectories have the same initial position and higher order derivatives as the initial states of the robots. % in the beginning. 
\eqref{Formulation:ObstacleAvoidance} and \eqref{Formulation:RobotAvoidance} enforce robot-obstacle and robot-robot collision avoidance, respectively.
\eqref{Formulation:Workspace} enforces that each robot stays within its defined workspace.
Lastly, \eqref{Formulation:DynamicLimits} enforces that the dynamic limits of the robot are obeyed.

Note that only constraint \eqref{Formulation:RobotAvoidance} stems from multiple robots. 
However, this seemingly simple constraint couples robots' trajectories both spatially and temporally, making the problem much harder.
As discussed in Section~\ref{Section:CentralizedAlgorithms}, solving the multi-agent path finding problem optimally is NP-Hard even for the discrete case while the discrete single-agent path finding problem can be solved optimally with classical search methods in polynomial time.
This curse of dimensionality affects continuous motion planning as well, where even geometric variants are known to be PSPACE-Hard~\citep{hopcroft1984complexity}.

A centralized planner can be used to solve the generic multi-robot trajectory planning problem in one-shot provided that: the current and future obstacle positions are known a priori, computed trajectories can be sent to robots over a communication link, and robots can track these trajectories well enough that they do not violate the spatio-temporal safety of the computed trajectories.
In the present work, we aim to approximately solve this problem in the case where obstacles are static, but not known a priori by a central entity, and there is no communication channel between robots or between robots and a central entity.
Each robot plans its own trajectory in real-time.
They plan at a high frequency to compensate for trajectory tracking errors.

\section{Preliminaries}
We now introduce essential mathematical concepts used herein.

\subsection{Parametric Curves and Splines}
Trajectories are curves $\vf : [0, T]\rightarrow \mathbb R^d$ that are parametrized by time, with duration $T$.
Mathematically, we choose to use splines, i.e. piecewise polynomials, where each piece is a B\'ezier curve defined by a set of control points and a duration. 

A B\'ezier curve $\vf : [0, T] \rightarrow \mathbb R^d$ of degree $h$ is defined by $h+1$ control points $\vPx{0}, \ldots, \vPx{h} \in \mathbb R^d$ as follows:
\begin{align*}
    \vf(t) &= \sum_{i=0}^h \vPx{i} {h \choose i} \left(\frac{t}{T}\right)^i\left(1-\frac{t}{T}\right)^{(h-i)}.
\end{align*}
Any B\'ezier curve $\vf$ satisfies $\vf(0) = \vPx{0}$ and $\vf(T) = \vPx{h}$.
Other points guide the curve from $\vPx{0}$ to $\vPx{h}$.
Since any B\'ezier curve $\vf$ is a polynomial of degree $h$, it is smooth, meaning $\vf\in C^\infty$.

We choose B\'ezier curves as pieces because of their \emph{convex hull property}: the curves themselves lie inside the convex hull of their control points, i.e., $\vf(t) \in ConvexHull\{\vPx{0},\ldots,\vPx{h}\}\ \forall t\in[0, T]$~\citep{farouki2012bernstein}.
Using the convex hull property, we can constrain a curve to be inside a convex region by constraining its control points to be inside the same convex region.
~\cite{tordesillas2020minvo} discuss the conservativeness of the convex hulls of control points of B\'ezier curves and show that the convex hulls are considerably less conservative than those of B-Splines~\citep{piegl1996nurbs}, which are another type of curve with the convex hull property.
Yet, they also show that convex hulls of the control points of the B\'ezier curves can be considerably more conservative compared to the smallest possible convex sets containing the curves.

\subsection{Linear Spatial Separations: Half-spaces, Convex Polytopes, and Support Vector Machines}

A hyperplane $\mH$ in $\mathbb R^d$ can be defined by a normal
vector $\mH_\vn \in \mathbb R^d$ and an offset $\mH_a$ as 
$\mH = \{\vx\in \mathbb{R}^d \mid \mH_\vn^\top\vx + \mH_a = 0\}$. 
A half-space $\tilde{\mH}$ in $\mathbb{R}^d$ is a subset of $\mathbb{R}^d$ that is bounded by a hyperplane such that $\tilde{\mH} = \{\vx\in \mathbb R^d \mid \mH_\vn^\top \vx + \mH_a \leq 0\}$.
A convex polytope is an intersection of a finite number of half-spaces.

Our approach relies heavily on computing safe convex polytopes and constraining spline pieces to be inside these polytopes.
Specifically, we compute hard-margin support vector machine (SVM)~\citep{cortes1995SVM} hyperplanes between spaces swept by robots along line segments and the obstacles/robots in the environment, and use these hyperplanes to create safe convex polytopes for robots to navigate in. 

\section{Assumptions}

Here, we list our additional assumptions about the problem formulation defined in Section~\ref{Section:ProblemStatement} and the capabilities of robots.

We assume that obstacles in the environment are static, i.e., $\mO(t) = \mO\ \forall t\in[0, \infty]$, and convex\footnote{For the purposes of our algorithm, concave obstacles can be described or approximated by a union of finite number of convex shapes provided that the union contains the original obstacle. Using our algorithm with approximations of concave obstacles results in trajectories that avoid the approximations.}.
Many existing, efficient, and widely-used mapping tools, including occupancy grids~\citep{homm2010efficient} and octomaps~\citep{hornung2013octomap}, internally store obstacles as convex shapes; such maps can be updated in real-time using visual or RGBD sensors, and use unions of convex axis-aligned boxes to approximate the obstacles in the environment.

Similarly, we assume that the shapes of the robots are convex.

We assume that the workspace $\mW_i \subseteq \mathbb{R}^d$ is a convex polytope.
It can be set to a bounding box that defines a room that a ground robot must not leave, a half-space that contains vectors with positive $z$ coordinates so that a quadrotor does not hit the ground or simply be set to $\mathbb R^d$. 
A non-convex workspace $\tilde{\mW_i}$ can be modeled by a convex workspace $\mW_i$ such that $\tilde{\mW_i} \subseteq \mW_i$ and a static set of convex obstacles $\tilde{\mO}$ that block portions of the convex workspace so that $\tilde{\mW_i} = \mW_i \backslash \left(\cup_{\mQ\in \tilde{\mO}} \mQ\right)$.

% We assume that if the robot $i$ can perfectly follow the desired trajectory $\vdx{i}(t)$, it is contained in the workspace $\mW_i$ at all times, i.e. $\forall t\in[0, T_i]\ \mR_i(\vdx{i}(t))\in \mW_i$.

To provide a guarantee of collision avoidance, we assume that robots can \emph{perfectly} sense the relative positions of obstacles and robots as well as their shapes in the environment.
The approach we propose does not require sensing of higher order state components (e.g., velocity, acceleration, etc.) or planned/estimated trajectories of objects, as the former is generally a noisy signal which cannot be expected to be sensed perfectly and the latter would require either communication or a potentially noisy trajectory estimation.

RLSS treats robots and obstacles differently.
It enforces that each robot stays within a spatial cell that is disjoint from the cells of other robots until the next planning iteration to ensure robot-robot collision avoidance.
To compute the spatial cell for each robot, RLSS uses positions and shapes of nearby robots, but not obstacles, in the environment.
Therefore, robots must be able to distinguish other robots from obstacles. 
However, we do not require individual identification of robots.

We assume that the team is cooperative in that each robot runs the same algorithm using the same replanning period.

Lastly, we assume that planning is synchronized between robots, meaning that each robot plans at the same time instant. The synchronization assumption is needed for ensuring robot-robot collision avoidance when planning succeeds\footnote{
In physical deployments, asynchronous planning can cause collisions between robots when they are in close proximity to each other.
To handle collisions stemming from asynchronous planning, robot shapes can be artificially inflated according to the maximum possible planning lag between robots (an empirical value)  at the expense of conservativeness.}.

\section{Approach}
Under the given assumptions, we solve the generic multi-robot trajectory planning problem approximately using decentralized receding horizon planning.
Each robot plans a trajectory, executes it for a short period of time, and repeats this cycle until it reaches its goal.
We call each plan-execute cycle an iteration.

We refer to the planning robot as the ego robot, and henceforth drop indices for the ego robot from our notation as the same algorithm is executed by each robot.
Workspace $\mW$ is the convex polytope in which the ego robot must remain.
$\mR$ is the collision shape function of the ego robot such that $\mR(\vp)\subset \mathbb{R}^d$ is the space occupied by the ego robot when it is placed at position $\vp$.
The ego robot is given a desired trajectory $\vd(t):[0, T]\rightarrow \mathbb{R}^d$ that it should follow.
The dynamic limits of the robot are modeled using required derivative degree $c$ up to which the trajectory must be continuous, and maximum derivative magnitudes $\gamma^k$ for required degrees $k \in \{1, \ldots, K\}$.

At every iteration, the ego robot computes a piecewise trajectory $\vf(t)$, which is dynamically feasible and safe up to the safety duration $s$, that it executes for the replanning period $\delta t \leq s$, and fully re-plans, i.e., runs the full planning pipeline, for the next iteration.
\emph{The only parameter that is shared by all robots is the replanning period $\delta t$.}

Without  loss of generality, we assume that  navigation begins at $t=0$, and at the start of planning iteration $u$, the current timestamp is $\tilde{T} = u\delta t$.

RLSS fits into the planning part of the classical robotics pipeline using perception, planning, and control.
The inputs from perception for the ego robot are:
\begin{itemize}
    \item $\mS$: Shapes of other robots\footnote{Practically, if the ego robot cannot sense a particular robot $j$ because it is not within the sensing range of the ego robot, robot $j$ can be omitted by the ego robot.
    As long as the sensing range of the ego robot is more than the maximum distance that can be travelled by the ego robot and robot $j$ in duration $\delta t$, omitting robot $j$ does not affect the behavior of the algorithm.}. $\mS_j \in \mS$ where $j \in \{1,\ldots, i-1, i+1,\ldots,N\}$ is the collision shape of robot $j$ sensed by the ego robot such that $\mS_j \subseteq \mathbb{R}^d$.
    \item $\mO$: The set of obstacles in the environment, where each obstacle $\mQ\in\mO$ is a subset of $\mathbb{R}^d$.
    \item $\vp$: Current position of the ego robot, from which derivatives up to required degree of continuity can be computed\footnote{In reality, all required derivatives of the position would be estimated using a state estimation method. We use this definition of $\vp$ for notational convenience.}.
\end{itemize}

We define $\hat{\mO} = \cup_{\mQ\in\mO}\mQ$ as the space occupied by the obstacles, and $\hat{\mS} = \cup_{\mS'\in\mS}\mS'$ as the space occupied by the robot shapes.
Robots sense the set of obstacles and the set of robot shapes and use those sets in practice.
We use spaces occupied by obstacles and robots for brevity in notation.

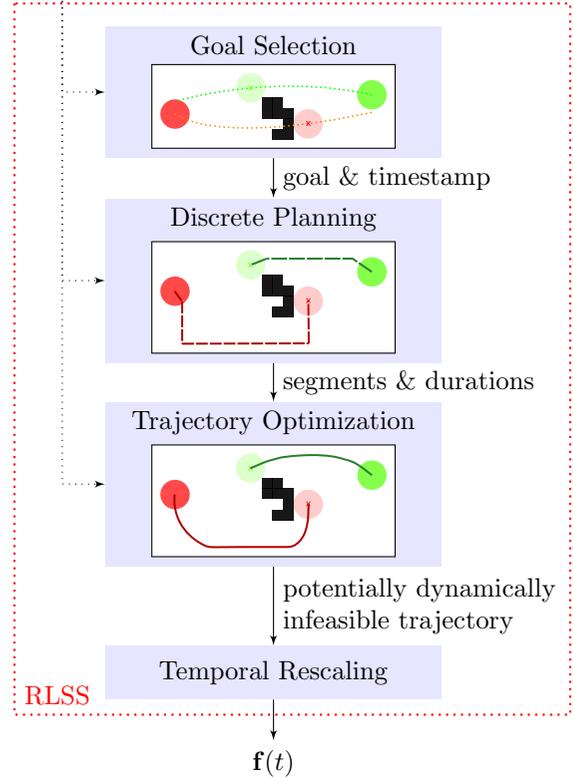
\begin{figure}
\centering
\begin{tikzpicture}
    \node [block] (goalselection) {Goal Selection\\\vspace*{0.1cm} \fcolorbox{black}{white}{\includegraphics[width=3cm]{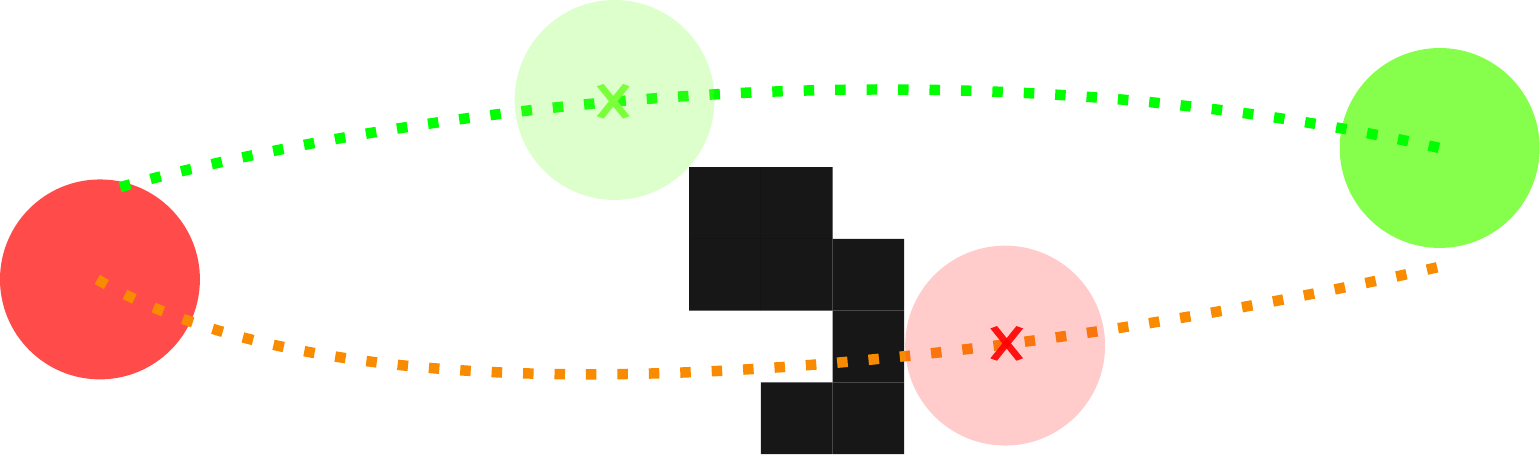}}};
    \node [block, below of = goalselection, node distance=2.5cm] (discretesearch) {Discrete Planning\\\vspace*{0.1cm} \fcolorbox{black}{white}{\includegraphics[width=3cm]{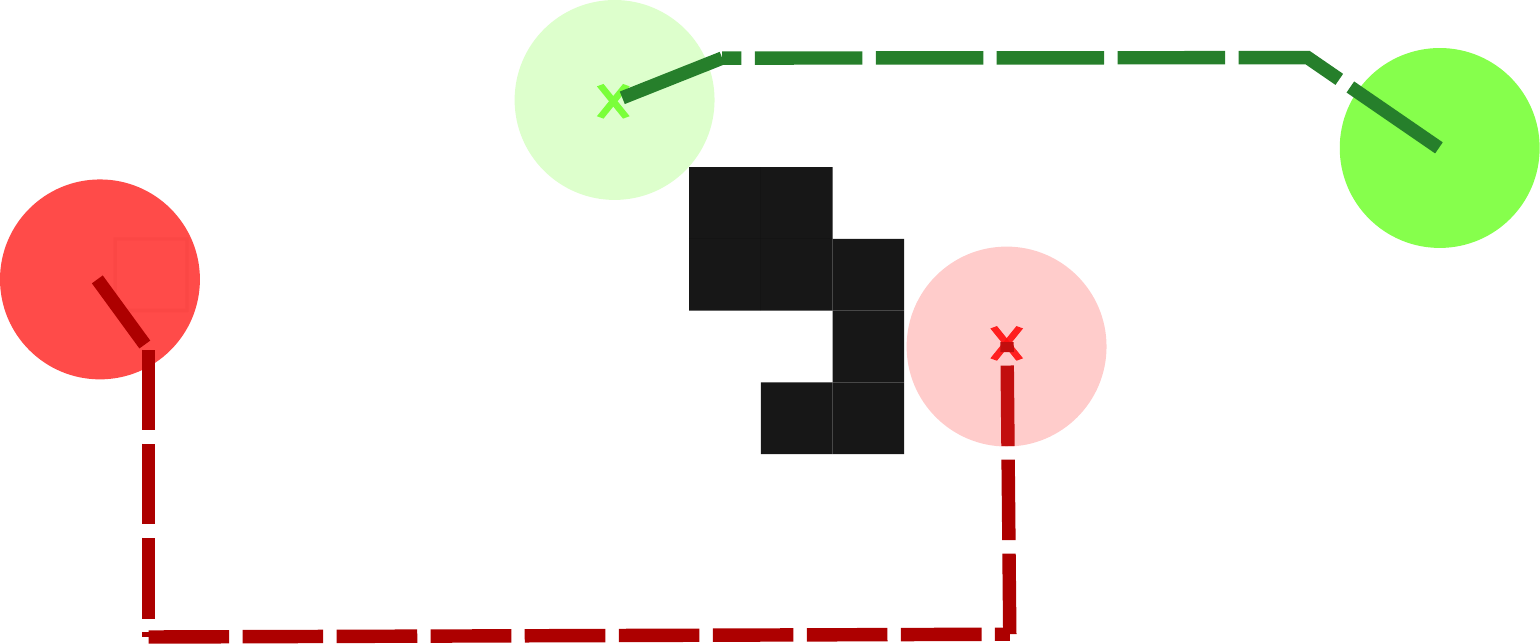}}};
    \node [block, below of = discretesearch, node distance=2.7cm] (trajopt) {Trajectory Optimization\\\vspace*{0.1cm} \fcolorbox{black}{white}{\includegraphics[width=3cm]{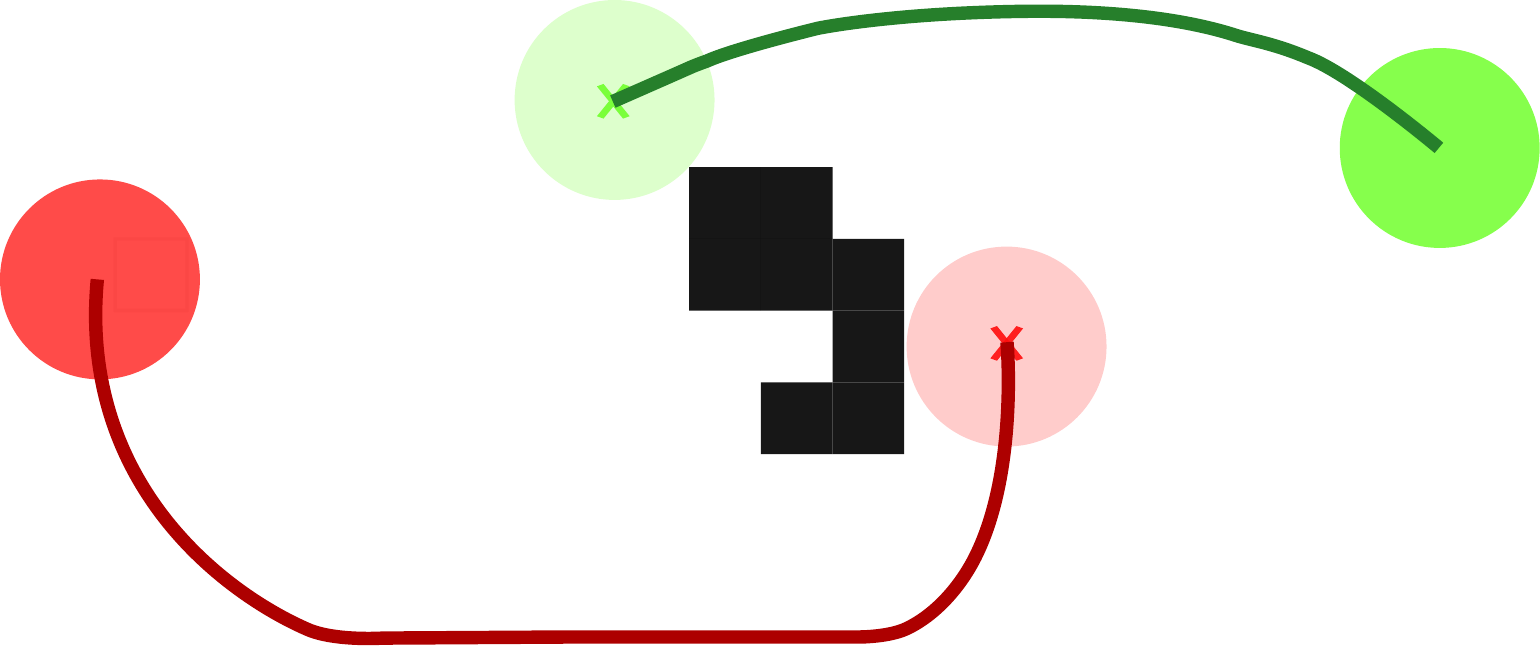}}};
    \node [block, below of = trajopt, node distance=2.5cm] (tempres) {Temporal Rescaling};
    \node [empty, below of = tempres, node distance = 1.2cm] (end) {$\vf(t)$};
    \node [empty] at (-2.8, 1.5) (sense) {};
    \node [empty, node distance = 0.5cm] at (-1.2, 1.6) (obstacles) {$\mS, \mO, \vp$ (inputs from sensing)};
    \node[text width=4cm, color = red] at (-1.3,-8.0) {RLSS};
    \path [line] (goalselection) -- node [right, align=left] {goal \& timestamp} (discretesearch);
    \path [line] (discretesearch) -- node [right, align=left] {segments \& durations} (trajopt);
   \path [line] (trajopt) -- node [right, align=left] {potentially dynamically\\ infeasible trajectory} (tempres);
    \path [line] (tempres) -- (end);
    \draw[red,thick,dotted] ($(goalselection.north west)+(-1.2,0.3)$) rectangle ($(tempres.south east)+(1.7,-0.2)$);
    \path [line, dotted]  (sense) |- (goalselection);
    \path [line, dotted] (sense) |- (discretesearch);
    \path [line, dotted] (sense) |- (trajopt);
\end{tikzpicture}
\caption{RLSS planning pipeline. Based on the sensed robots $\mS$, sensed obstacles $\mO$, and current position $\vp$, the ego robot computes the trajectory $\vf(t)$ that is dynamically feasible and safe up to time $s$.}
\label{Figure:PlanningPipeline}
\end{figure}

There are $4$ main stages of RLSS: 1) goal selection, 2) discrete planning, 3) trajectory optimization, and 4) temporal rescaling.
The planning pipeline is summarized in Fig.~\ref{Figure:PlanningPipeline}.
At each planning iteration, the ego robot executes the four stages to plan the next trajectory.
In the goal selection stage, a goal position and the corresponding timestamp of the goal position on the desired trajectory $\vd(t)$ is selected.
In the discrete planning stage, a discrete path from robot's current position toward the selected goal position is computed and durations are assigned to each segment.
%Assigning durations to discrete segments is done at this stage as well.
In the trajectory optimization stage, discrete segments are smoothed to a piecewise trajectory.
In the temporal rescaling stage, the dynamic limits of the robot are checked and duration rescaling is applied if necessary. 

Next, we describe each stage in detail. 
Each stage has access to the workspace $\mW$, the collision shape function $\mR$, the desired trajectory $\vd(t)$ and its duration $T$, the maximum derivative magnitudes $\gamma^k$, and the derivative degree $c$ up to which the trajectory must be continuous.
We call these task inputs.
They describe the robot shape, robot dynamics, and the task at hand; these  are not parameters that can be tuned freely and they do not change during navigation.

\subsection{Goal Selection}

\begin{figure}
    \centering
     \subfloat[Desired Trajectories]{%
       \includegraphics[width=0.49\linewidth]{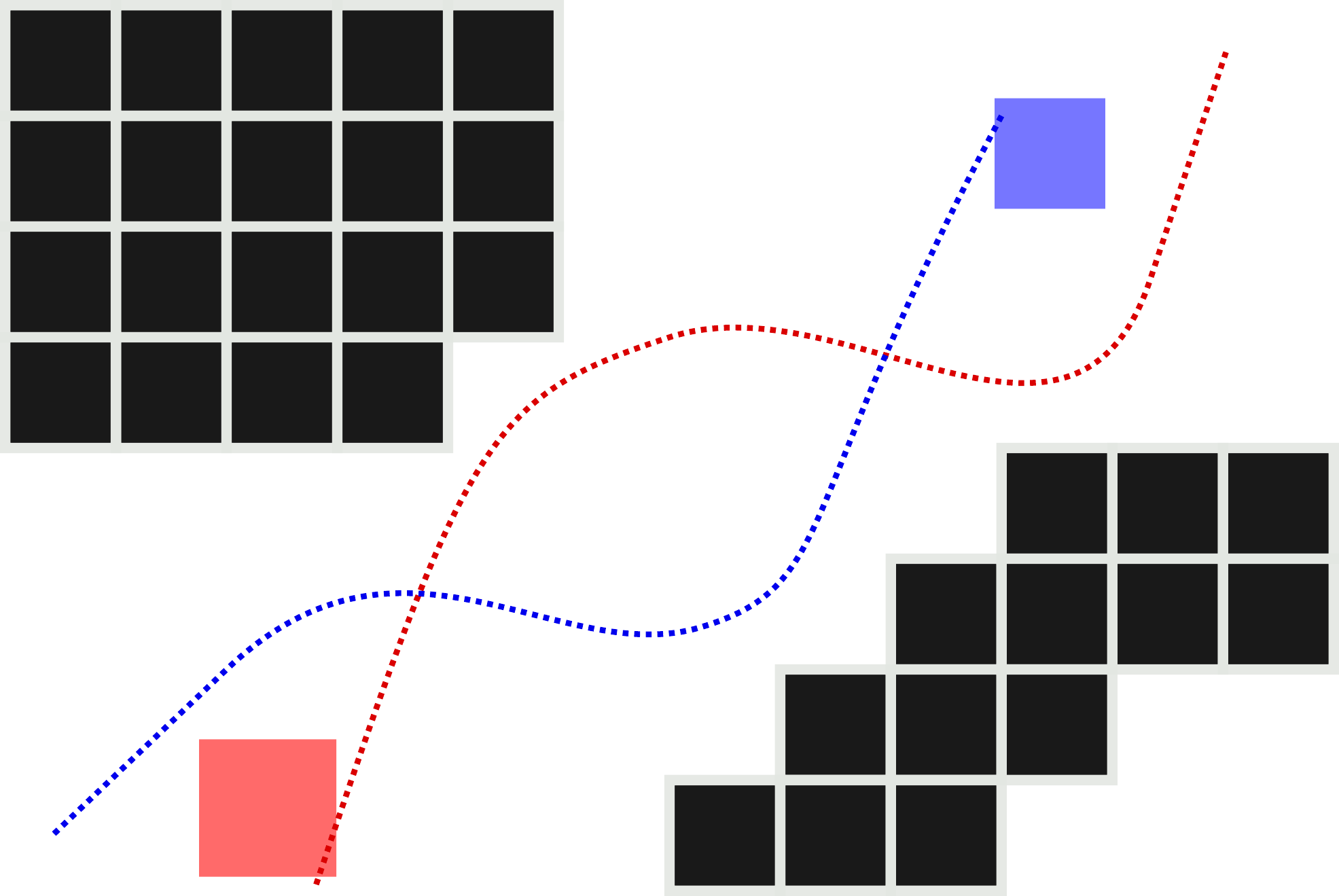}
     }
     \subfloat[Red Goal Selection]{%
       \includegraphics[width=0.49\linewidth]{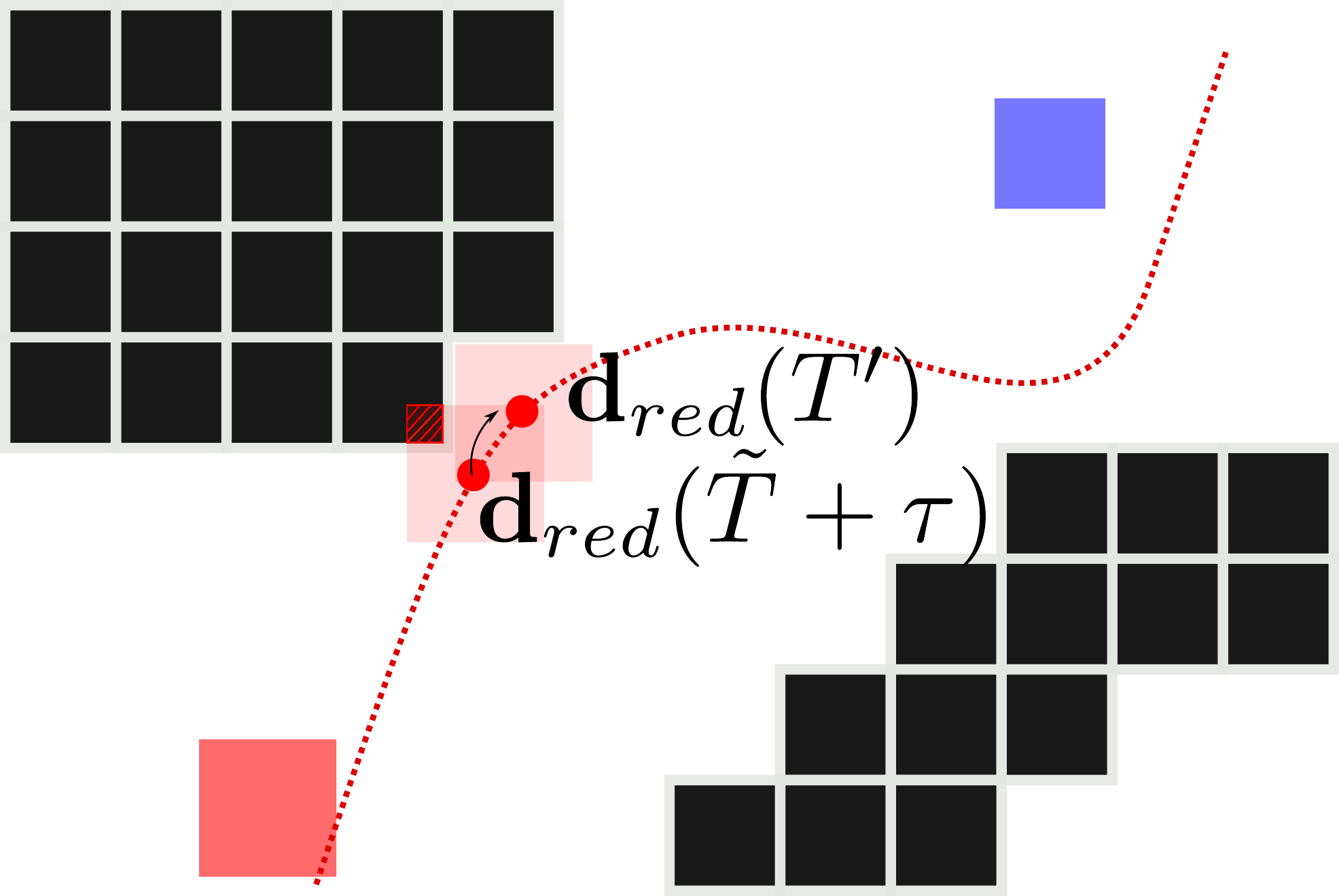}
     }
     \hfill
    \subfloat[Blue Goal Selection]{%
       \includegraphics[width=0.49\linewidth]{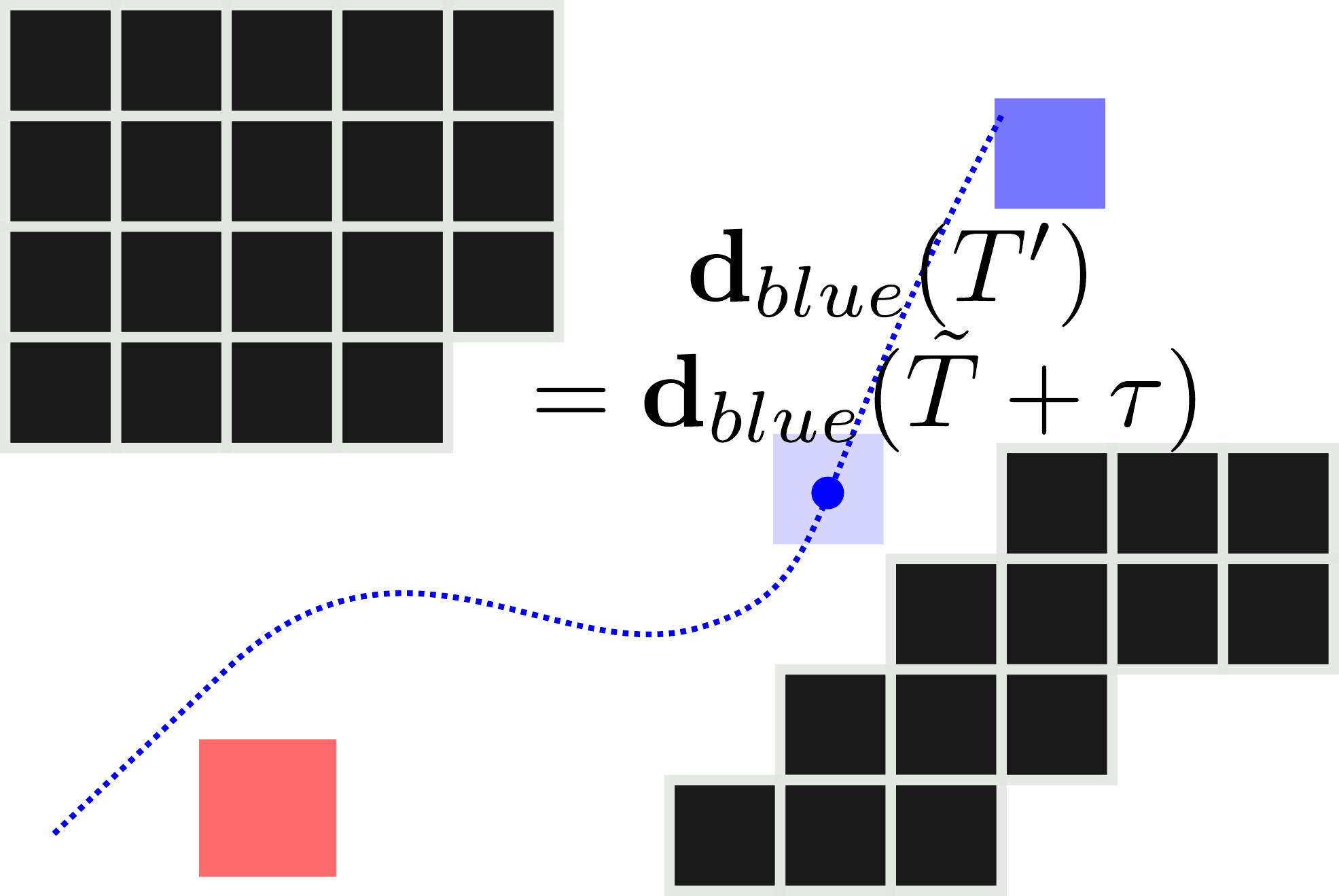}
     }
     \subfloat[Selected Goals]{%
       \includegraphics[width=0.49\linewidth]{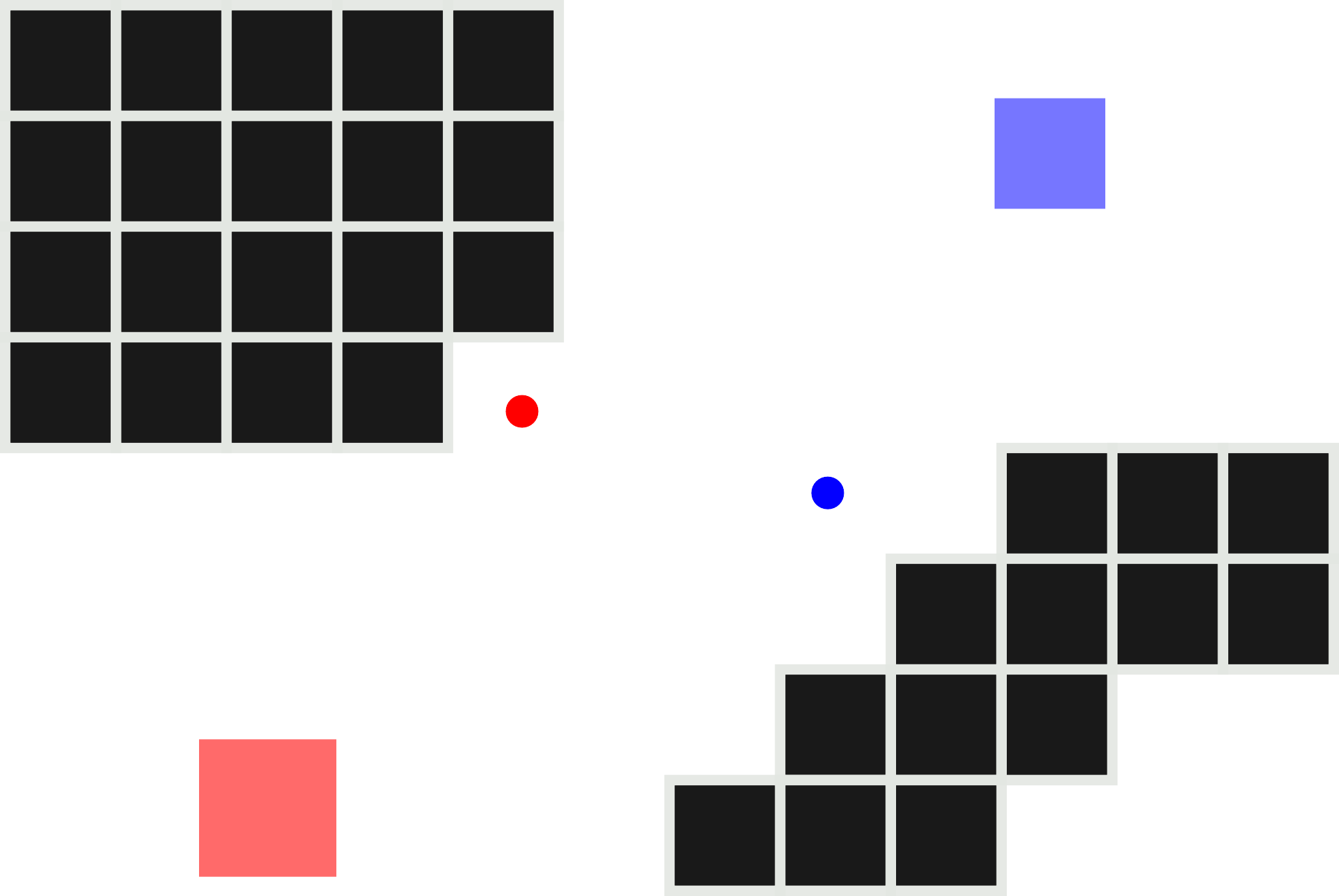}
     }
     \caption{\emph{Goal Selection.} a) Blue and red squares are robots, while obstacles are black boxes. The desired trajectories $\vdx{red}(t)$ and $\vdx{blue}(t)$ of each robot are given as dotted lines. Safety distance $D$ is set to $0$ for clarity. b) The desired trajectory of the red robot is not collision-free at timestamp $\tilde{T} + \tau$. It selects its goal timestamp $T'$ (and hence its goal position) by solving~\eqref{Equation:GoalSelection}, which is the closest timestamp to $\tilde{T} + \tau$ when the robot, when placed on the desired trajectory, would be collision free. c) Since the desired trajectory of blue robot is collision free at timestamp $\tilde{T} + \tau$, it selects its goal timestamp $T' = \tilde{T} + \tau$ after solving~\eqref{Equation:GoalSelection}. d) Selected goal positions are shown.}
     \label{Figure:GoalSelection}
\end{figure}

At the goal selection stage (Algorithm~\ref{Algorithm:GoalSelection}), we find a goal position $\vg$ on the desired trajectory $\vd(t)$ and a timestamp by which it should be reached.
These are required in the subsequent discrete planning stage, which is a goal-oriented search algorithm.
%that is should be (or should have been) reached at.

Goal selection has two parameters: the desired planning horizon $\tau$ and safety distance $D$.
It uses the robot collision shape function $\mR$, desired trajectory $\vd(t)$ and its duration $T$, and workspace $\mW$ from the task inputs.
The inputs of goal selection are the shapes of other robots $\mS$, obstacles in the environment $\mO$, current position $\vp$, and the current timestamp $\tilde{T}$.

At the goal selection stage, the algorithm finds the timestamp $T'$ that is closest to $\tilde{T} + \tau$ (i.e., the timestamp that is one planning horizon away from the current timestamp) when the robot, if placed on the desired trajectory at $T'$, is at  least safety distance $D$ away from all objects in the environment. 
We use the safety distance $D$ as a heuristic to choose goal positions that have free volume around them in order not to command robots into tight goal positions.
Note that goal selection only chooses a single point on the desired trajectory that satisfies the safety distance; the actual trajectory the robot follows will be planned by the rest of the algorithm.
Formally, the problem we solve in the goal selection stage is given as follows:
\begin{equation}
    \begin{aligned}
        T' = \arg&\min_{t}  \lvert t-(\tilde{T}+\tau)\rvert \ s.t.\\
        &t\in[0, T]\\
        &\text{min-dist}(\mR(\vd(t)), \hat{\mO} \cup \hat{\mS} \cup \partial\mW) \geq D 
    \hspace{-10pt}\end{aligned}
    \label{Equation:GoalSelection}
\end{equation}
where $\partial\mW$ is the boundary of workspace $\mW$, and min-dist returns the minimum distance between two sets.

We solve~\eqref{Equation:GoalSelection} using linear search on timestamps starting from $\tilde{T} + \tau$ with small increments and decrements.

Figure~\ref{Figure:GoalSelection}, demonstrates the goal selection procedure for a particular instance.

If there is no safe point on the desired trajectory, i.e. if the robot is closer than $D$ to objects when it is placed on any point on the desired trajectory, we return the current position and timestamp.
This allows us to plan a safe stopping trajectory.

\begin{algorithm}[bt]
    \Input{$\mS:$ Set of robot shapes}
    \Input{$\mO:$ Set of obstacles}
    \Input{$\vp:$ Current position}
    \Input{$\tilde{T}:$ Current timestamp}
    \TaskInput{$\mR:$ Collision shape function}
    \TaskInput{$\vd(t), T:$ Desired trajectory and its duration}
    \TaskInput{$\mW:$ Workspace}
    \Parameter{$\tau:$ Desired planning horizon}
    \Parameter{$D:$ Safety distance}
    \Returns{Goal and timestamp by which it should be reached}
    $T' \gets$ Solve~\eqref{Equation:GoalSelection} with linear search\;
    
    \eIf{\eqref{Equation:GoalSelection} is infeasible}{
        \Return ($\vp$, $\tilde{T}$)}{
        $\vg \gets \vd(T')$\;
        
        \Return ($\vg$, $T'$)}
\caption{GOAL-SELECTION}
\label{Algorithm:GoalSelection}
\end{algorithm}

Note that while the selected goal position has free volume around it, it may not be reachable by the ego robot.
For example, the goal position may be encircled by obstacles or other robots.
Therefore, we use a best-effort search method during discrete planning (as described in Section~\ref{Section:DiscretePlanning}) that plans a path \emph{towards} the goal position.

The goal and the timestamp are used in the subsequent discrete planning stage as suggestions.

\subsection{Discrete Planning}\label{Section:DiscretePlanning}

\begin{algorithm}[bt]
\Input{$\vg:$ Goal position}
\Input{$T':$ The timestamp that goal position should be (or should have been) reached at}
\Input{$\mS:$ Set of robot shapes}
\Input{$\mO:$ Set of obstacles}
\Input{$\vp:$ Current position}
\Input{$\tilde{T}:$ Current timestamp}
\TaskInput{$\mR: $ Collision shape function}
\TaskInput{$\mW: $ Workspace}
\TaskInput{$\gamma^1:$ Maximum velocity the robot can execute}
\Parameter{$\sigma: $ Step size of search grid}
\Parameter{$s:$ Duration up to which computed trajectory must be safe. $s \geq \delta t$ must hold.}
\Returns{Discrete path and duration assignments to segments}
\setcounter{AlgoLine}{0}
$\mF \gets \mW \backslash(\hat{\mO} \cup \hat{\mS})$\; 
\label{Line:FreeSpace}

$actions \gets \text{BEST-EFFORT-}A^*(\vp, \vg, \mR, \mF, \sigma)$\;\label{Line:BestEffortAStar}

$\{\ve_1, \ldots, \ve_L\} \gets \text{EXTRACT-SEGMENTS} (actions)$\label{Line:ExtractSegments}

Prepend $\ve_1$ to $\{\ve_1, \ldots, \ve_L\}$ so that $\ve_0 = \ve_1$.\; \label{Line:PrepentFirstPoint}

$totalLength \gets \sum_{i=1}^L \normtwo{\ve_i - \ve_{i-1}}$\; \label{Line:TotalLength}

$fDuration \gets \max\left(T' - \tilde{T}, \frac{totalLength}{\gamma^1}\right)$\; \label{Line:FeasibleDuration}

$T_1 \gets s$\label{Line:FirstSegment}

\For{$i=2 \to L$} {\label{Loop:DurationAssignmentStart}
    $T_i \gets fDuration\frac{\normtwo{\ve_i-\ve_{i-1}}}{totalLength}$
}\label{Loop:DurationAssignmentEnd}

\Return $\{\ve_0, \ldots, \ve_L\}, \{T_1, \ldots, T_L\}$
\caption{DISCRETE-PLANNING}
\label{Algorithm:DiscretePlanning}
\end{algorithm}

Discrete planning (Algorithm~\ref{Algorithm:DiscretePlanning}) performs two main tasks: i) it finds a collision-free discrete path from the current position $\vp$ towards the goal position $\vg$, and ii) it assigns durations to each segment of the discrete path.
The discrete path found at the discrete planning stage represents the homotopy class of the final trajectory. Trajectories in the same homotopy class can be smoothly deformed into one another without intersecting obstacles~\citep{bhattacharya2010homotopy}. The subsequent trajectory optimization stage computes a smooth trajectory within the homotopy class.
The trajectory optimization stage utilizes the discrete path to i) generate obstacle avoidance constraints and ii) guide the computed trajectory by adding distance cost terms between the discrete path and the computed trajectory.
It uses the durations assigned by the discrete planning stage as the piece durations of the piecewise trajectory it computes.

Finding a discrete path from the start position $\vp$ to the goal position $\vg$ is done with best-effort $A^*$ search (Line~\ref{Line:BestEffortAStar}, Algorithm~\ref{Algorithm:DiscretePlanning}), which we define as follows.
If there are valid paths from $\vp$ to $\vg$, we find the least-cost path among such paths.
If no such path exists, we choose the least-cost path to the position that has the lowest heuristic value (i.e., the position that is heuristically closest to $\vg$).
This modification of $A^*$ search is done due to the fact that the goal position may not always be reachable, since the goal selection stage does not enforce reachability.

The ego robot plans its path in a search grid where the grid has hypercubic cells (i.e. square in 2D, cube in 3D) with edge size $\sigma$, which we call the step size of the search.
The grid shifts in the environment with the robot in the sense that the robot's current position always coincides with a grid center.
Let $\mF = \mW \backslash (\hat{\mO} \cup \hat{\mS})$ be the free space within the workspace (Line~\ref{Line:FreeSpace}, Algorithm~\ref{Algorithm:DiscretePlanning}).
We do not map free space $\mF$ to the grid.
Instead, we check if the robot shape swept along any segment on the grid is contained in $\mF$ or not, to decide if a movement is valid.
This allows us to i) model obstacles and robot shapes independently from the grid's step size $\sigma$, and ii) shift the grid with no additional computation since we do not store occupancy information within the grid.

The states during the search have two components: position $\vpi$ and direction $\vDelta$.
Robots are allowed to move perpendicular or diagonal to the grid.
This translates to $8$ directions in $2$-dimension, $26$ directions in $3$-dimension.
Goal states are states that have position $\vg$ and any direction.
We model directions using vectors $\vDelta$ of $d$ components where each component is in $\{-1, 0, 1\}$.
When the robot moves $1$ step along direction $\vDelta$, its position changes by $\sigma\vDelta$.
The initial state of the search is the ego robot's current position $\vp$ and direction $\mathbf{0}$.

\begin{table}
    \centering
    \caption{Discrete search actions and their costs.}
    \begin{tabular}{|c|L|c|}
         \hline
         Action & \centering Description & Cost \\
         \hline
         ROTATE & Change current direction to a new direction. & 1\\
         \hline
         FORWARD & Move forward from the current position $\vpi$ along current direction $\vDelta$ by the step size $\sigma$. It is only available when $\vDelta \neq \mathbf{0}$. & $\normtwo{\vDelta}$\\
         \hline
         REACHGOAL &  Connect the current position $\vpi$ to the goal position $\vg$ where step size of the grid is $\sigma$. & $1 + \frac{\normtwo{\vpi - \vg}}{\sigma}$\\
         \hline
    \end{tabular}
    \label{Table:DiscreteSearchActions}
\end{table}

There are $3$ actions in the search formulation: ROTATE, FORWARD, and REACHGOAL, summarized in Table~\ref{Table:DiscreteSearchActions}.
ROTATE action has a cost of $1$.
The cost of the FORWARD action is the distance travelled divided by the step size (i.e. cells travelled), which is equal to the size $\normtwo{\vDelta}$ of the direction vector.
REACHGOAL has the cost of one rotation plus cells travelled from $\vpi$ to goal position $\vg$: $1 + \frac{\normtwo{\vpi - \vg}}{\sigma}$.
One rotation cost is there because it is almost surely required to do one rotation before going to goal from a cell.
ROTATE actions in all directions are always valid whenever the current state is valid.
FORWARD and REACHGOAL actions are valid whenever the robot shape $\mR$ swept along the movement is contained in free space $\mF$.

For any state ($\vpi, \vDelta$), we use the Euclidean distance from position $\vpi$ to the goal position $\vg$ divided by step size $\sigma$ (i.e. cells travelled when $\vpi$ is connected to $\vg$ with a straight line) as the admissible heuristic.

\begin{lemma}
In the action sequence of the resulting plan 
\begin{itemize}
\item[1.] each ROTATE action must be followed by at least one FORWARD action,
\item[2.] the first action cannot be a FORWARD action,
\item[3.] and no action can appear after REACHGOAL action.
\end{itemize}
\begin{proof}[Proof Sketch]\renewcommand{\qedsymbol}{}
${}$

% \begin{itemize}
1. After each ROTATE action, a FORWARD action must be executed in a least-cost plan because i) a ROTATE action cannot be the last action since goal states accept any direction and removing any ROTATE action from the end would result in a valid lower cost plan, ii) the REACHGOAL action cannot appear after ROTATE action because REACHGOAL internally assumes robot rotation and removing the ROTATE action would result in a valid lower cost plan, and iii) there cannot be consecutive ROTATE actions in a least-cost path as each rotation has the cost of $1$ and removing consecutive rotations and replacing them with a single rotation would result in a valid lower cost plan. 

2. The first action cannot be a FORWARD action since initial direction is set to $\mathbf{0}$ and FORWARD action is available only when $\vDelta \neq \mathbf{0}$.

3. No action after a REACHGOAL action can appear in a least cost plan because REACHGOAL connects to the goal position, which is a goal state regardless of the direction. Removing any action after REACHGOAL action would result in a valid lower cost plan.
% \end{itemize}
\end{proof}
\label{Lemma:ActionSequence}
\end{lemma}

By Lemma~\ref{Lemma:ActionSequence}, the action sequence can be described by the following regular expression in POSIX-Extended Regular Expression Syntax:
\scalebox{0.98}{\parbox{.5\linewidth}{
\begin{align*}
    (\text{(ROTATE)}\text{(FORWARD)}^+)^*\text{(REACHGOAL)}^{\{0,1\}}
\end{align*}
}}

\begin{figure}
    \centering
     \subfloat[Goal Positions]{%
       \includegraphics[width=0.49\linewidth]{4-selected-goals.pdf}
     }
     \subfloat[Red Discrete Planning]{%
       \includegraphics[width=0.49\linewidth]{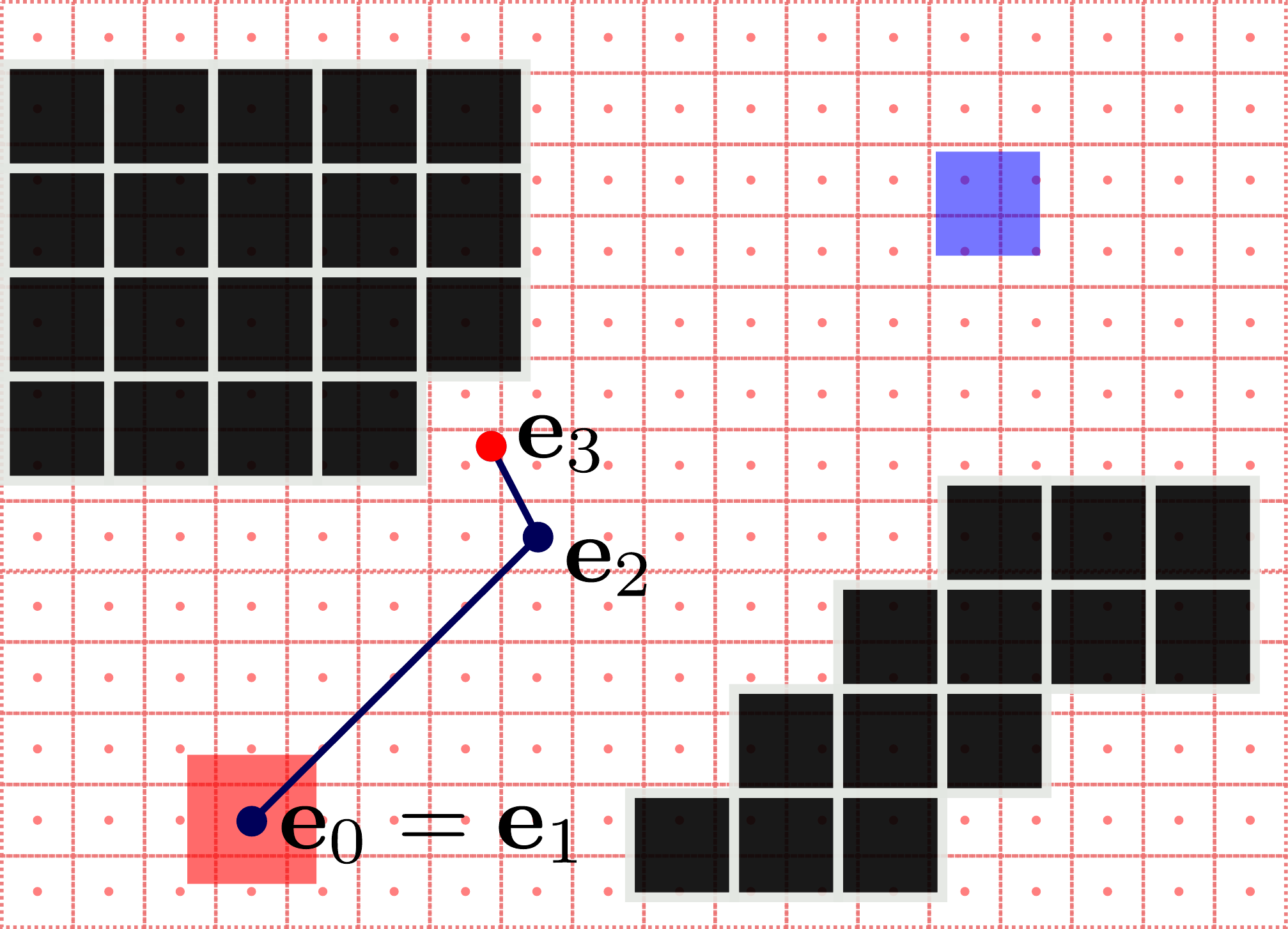}
     }
     \hfill
    \subfloat[Blue Discrete Planning]{%
       \includegraphics[width=0.49\linewidth]{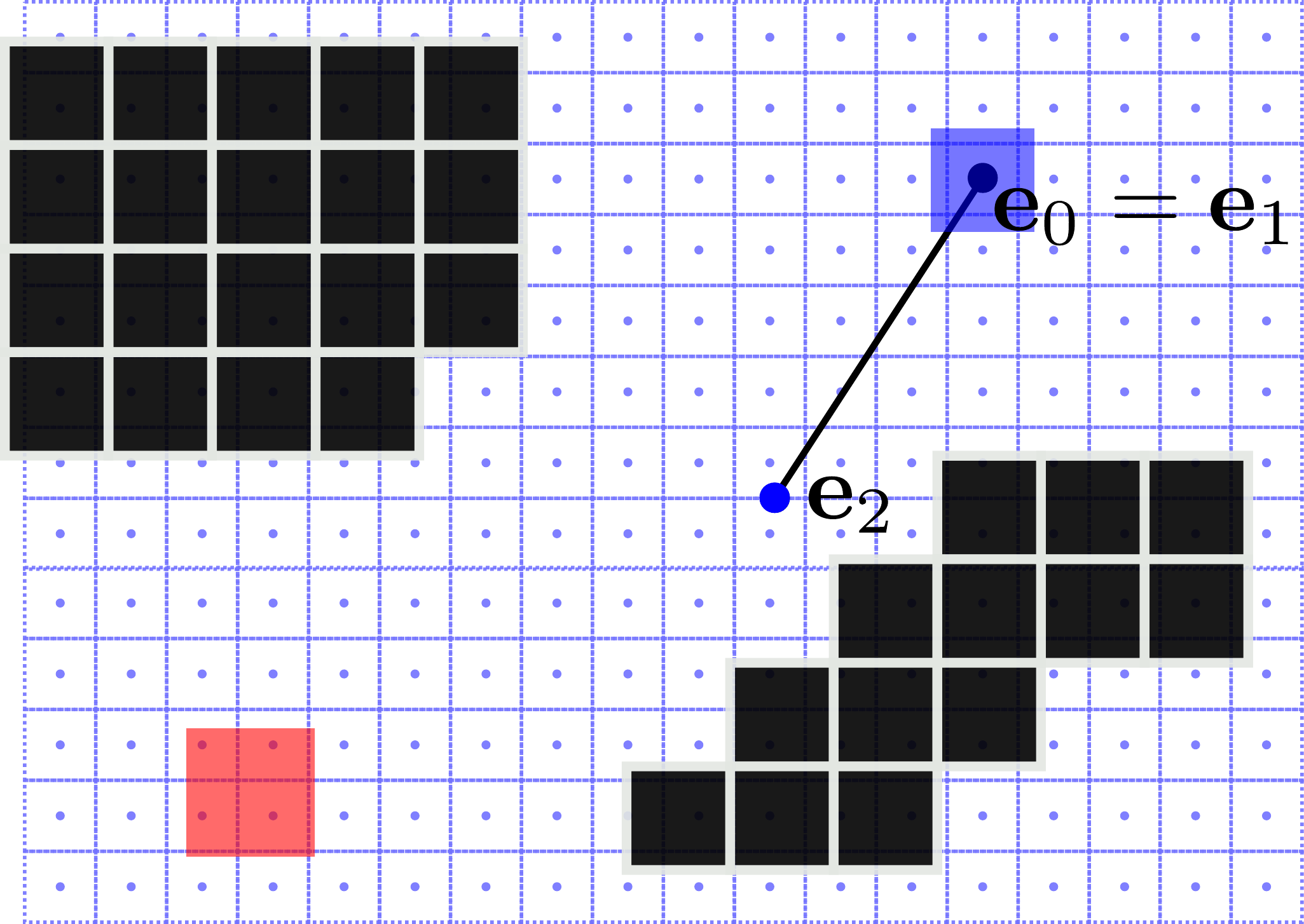}
     }
     \subfloat[Discrete Paths]{%
       \includegraphics[width=0.49\linewidth]{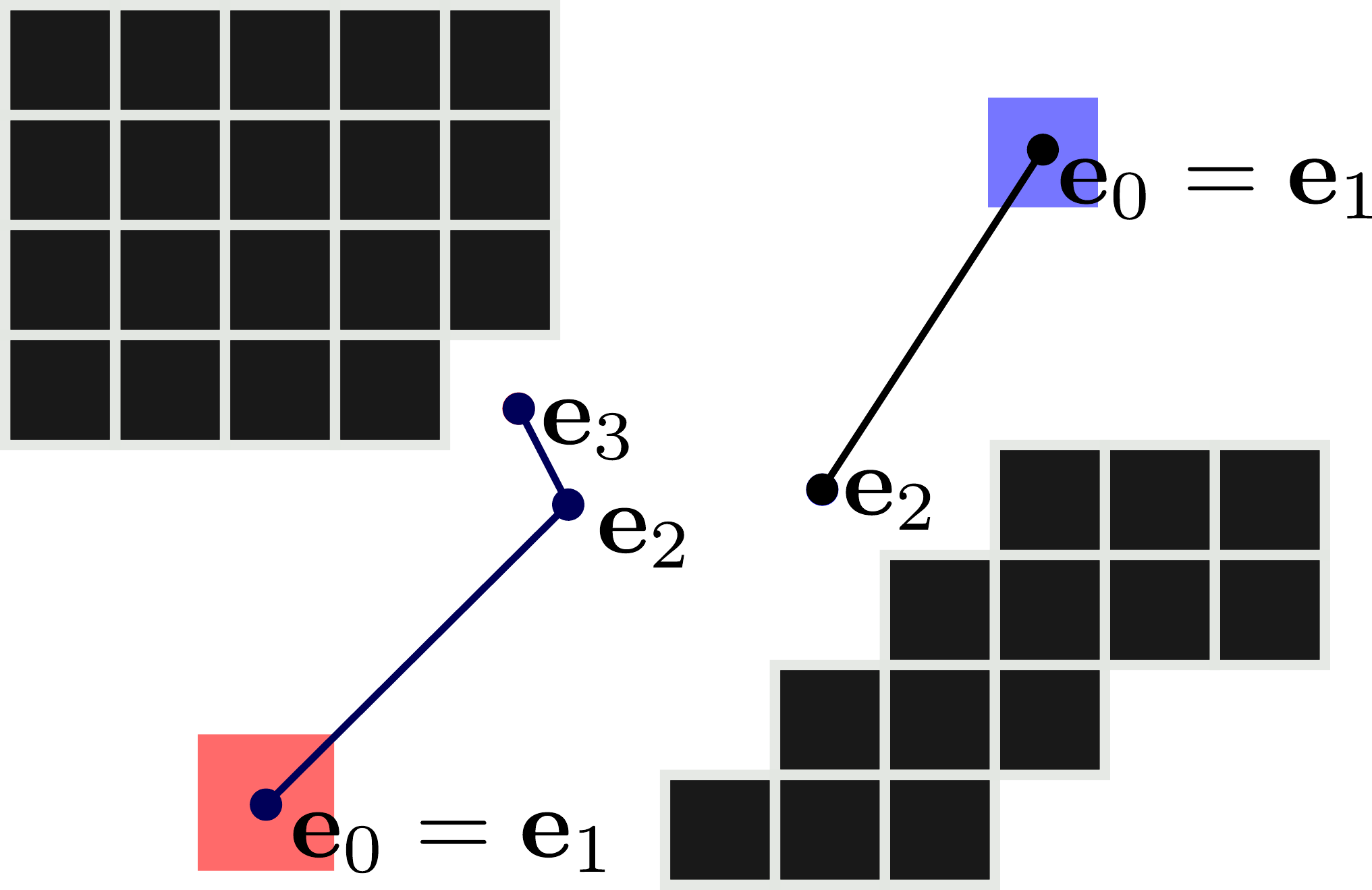}
     }
     \caption{\emph{Discrete Planning.} a) Goal positions of two robots computed at the goal selection stage are given. b) Red robot plans a discrete path from its current position to its goal position on a search grid that is aligned on its current position. c) Blue robot plans a discrete path from its current position to its goals position on a search grid that is aligned on its current position. Computed discrete paths are prepended with robot start positions to have $0$-length first segments. d) The resulting discrete paths are given.}
     \label{Figure:DiscretePlanning}
\end{figure}

We collapse consecutive FORWARD actions in the resulting plan and extract discrete segments (Line~\ref{Line:ExtractSegments}, Algorithm~\ref{Algorithm:DiscretePlanning}). 
Each $\text{(ROTATE)}\text{(FORWARD)}^+$ sequence and REACHGOAL becomes a separate discrete segment.
Let $\{\ve_1, \ldots, \ve_{L}\}$ be the endpoints of discrete segments.
We prepend the first endpoint to the endpoint sequence in order to have a $0$-length first segment for reasons that we will explain in Section~\ref{Section:TrajectoryOptimization} (Line ~\ref{Line:PrepentFirstPoint}, Algorithm~\ref{Algorithm:DiscretePlanning}).
The resulting $L$ segments are described by $L+1$ endpoints $\{\ve_0, \ldots, \ve_{L}\}$ where $\ve_0 = \ve_1$.
Example discrete paths for two robots are shown in Fig.~\ref{Figure:DiscretePlanning}.

Next, we assign durations to each segment.
The total duration of the segments is computed using the ego robot's maximum velocity $\gamma^1$, the timestamp $T'$ that the goal position should be reached by, and the current timestamp $\tilde{T}$.
We use $T' - \tilde{T}$ as the desired duration of the plan.
%It is the duration left to reach to the goal position.
However, if $T' - \tilde{T}$ is negative (i.e. $T' < \tilde{T}$,  meaning that the goal position should been reached in the past), or $T' - \tilde{T}$ is small such that the robot cannot traverse the discrete segments even with its maximum velocity $\gamma^1$, we adjust the desired duration to a feasible one, specifically %.
%In those cases, we set the feasible duration to 
the total length of segments divided by the maximum velocity (Line~\ref{Line:FeasibleDuration}, Algorithm~\ref{Algorithm:DiscretePlanning}).
We distribute the feasible duration to the segments except for the first one, proportional to their lengths (Loop at line~\ref{Loop:DurationAssignmentStart}, Algorithm~\ref{Algorithm:DiscretePlanning}).
We set the duration of the first segment, which has zero length, to the safety duration $s$ (Line~\ref{Line:FirstSegment}, Algorithm~\ref{Algorithm:DiscretePlanning}); the reason for this will be explained in Section~\ref{Section:TrajectoryOptimization}.

The outputs of  discrete planning are segments described by endpoints $\{\ve_0, \ldots, \ve_L\}$ with assigned durations $\{T_1, \ldots, T_L\}$ that are used in the trajectory optimization stage. 

\subsection{Trajectory Optimization}\label{Section:TrajectoryOptimization}

In the trajectory optimization stage (Algorithm~\ref{Algorithm:TrajectoryOptimization}), we formulate a convex quadratic optimization problem (QP) to compute a piecewise trajectory $\vf(t)$ by smoothing discrete segments.
The computed trajectory is collision-free and continuous up to the desired degree of derivative.
However, it may be dynamically infeasible (i.e., derivative magnitudes may exceed the maximum allowed derivative magnitudes $\gamma^k$); this is resolved during the subsequent temporal rescaling stage.

The decision variables of the optimization problem are the control points of an $L$-piece spline where each piece is a B\`ezier curve. 
The duration of each piece is assigned during discrete planning. Let $T = \sum_{i=1}^{L}T_i$ denote the total duration of the planned trajectory.
The degree of the B\'ezier curves is tuned with the parameter $h$.
Let $\vPxy{i}{j} \in \mathbb{R}^d$ be the $j^{th}$ control point of the $i^{th}$ piece where $i\in\{1,\ldots,L\}, j\in\{0,\ldots,h\}$.
Let $\mP = \{\vPxy{i}{j} \mid i\in\{1,\ldots,L\}, j\in\{0,\ldots,h\}\}$ be the set of all control points.

\subsubsection{Constraints}

There are $4$ types of constraints on the trajectory; 
all are linear in the decision variables $\mP$. 

\emph{1) Workspace constraints:} 
We shift each bounding hyperplane of workspace $\mW$ (there are a finite number of such hyperplanes as $\mW$ is a convex polytope) to account for the robot's collision shape $\mR$, such that when the robot's position is on the safe side of the shifted hyperplane, the entire robot is on the safe side of the original hyperplane (Line~\ref{Line:WorkspaceBuffering}, Algorithm~\ref{Algorithm:TrajectoryOptimization}).

\begin{algorithm}[H]
\Input{$\{\ve_0, \ldots, \ve_L\}:$ Endpoints of discrete segments}
\Input{$\{T_1, \ldots, T_L\}:$ Durations of discrete segments}
\Input{$\mS:$ Set of robot shapes}
\Input{$\mO:$ Set of obstacles}
\Input{$\vp:$ Current position}
\TaskInput{$\mR:$ Collision shape function}
\TaskInput{$\mW:$ Workspace}
\TaskInput{$c:$ Degree of derivative up to which resulting trajectory must be continuous}
\Parameter{$h:$ Degree of B\'ezier curves}
\Parameter{$\tilde{o}: $ Obstacle check distance}
\Parameter{$\tilde{r}: $ Robot check distance}
% % \Parameter{$\tilde{o}:$ Obstacle check distance}
% % \Parameter{$\tilde{r}:$ Robot check distance}
\Parameter{$\tilde{p}:$ Preferred distance to objects}
\Parameter{$\alpha:$ Preferred distance cost weight}
\Parameter{$\vtheta:$ Endpoint cost weights}
\Parameter{$\vlambda:$ Integrated derivative cost weights}
\Returns{Potentially dynamically infeasible trajectory}
\setcounter{AlgoLine}{0}

$QP$ is a quadratic program with variables $\mP$

$\Upsilon_\mW \gets \text{BUFFER-WORKSPACE}(\mW, \mR)$\;\label{Line:WorkspaceBuffering}

addWorkspaceConstraints($QP$, $\Upsilon_\mW$)\label{Line:WorkspaceConstraints}

$\Upsilon_i \gets \emptyset\ \forall i\in\{1,\ldots,L\}$

\For{$\forall \mS_j\in\mS\ $min-dist$(\mS_j, \mR(\vp)) \leq \tilde{r}$}{\label{Loop:RobotRobotCA}
$\Upsilon_1 \gets \Upsilon_1 \cup \{\text{BUFFERED-SVM}(\mS_j, \mR(\vp), \mR)\}$
}

\For{$i=1 \to L$}{\label{Loop:RobotObstacleCA}
    $\mathcal{\zeta}_i \gets$ region swept by $\mR$ from $\ve_{i-1}$ to $\ve_i$
    
    \For{$\forall \mQ\in\mO\ \text{min-dist}(\mQ, \mathcal{\zeta}_i) \leq \tilde{o}$}{
    $\Upsilon_i \gets \Upsilon_i \cup \{\text{BUFFERED-SVM}(\mQ, 
    \mathcal{\zeta}_i, \mR)\}$
    }
}

addCollAvoidanceConstraints($QP$, $\Upsilon_1,\ldots,\Upsilon_L$)

addContinuityConstraints($QP$, $c$, $\vp$, $T_1,\ldots,T_L$)\label{Line:ContinuityConstraints}

addEnergyCostTerm($QP$, $\vlambda$, $T_1,\ldots,T_L$)\label{Line:EnergyCost}

addDeviationCostTerm($QP$, $\vtheta$, $\ve_1, \ldots, \ve_L$)\label{Line:DeviationCost}

$\overline{\Upsilon}_1 \gets \text{SHIFT-HYPERPLANES}(\Upsilon_1, \tilde{p})$\label{Line:PreferredShift}

addPreferredDistanceCostTerm($QP$, $\overline{\Upsilon}_1, \alpha$)\label{Line:PreferredCost}

$\vf(t) \gets \text{SOLVE-QP}(QP)$\label{Line:SolveQP}

\Return $\vf(t)$

\caption{TRAJECTORY-OPTIMIZATION}
\label{Algorithm:TrajectoryOptimization}
\end{algorithm}

Let $\Upsilon_{\mW}$ be the set of shifted hyperplanes of $\mW$.
We constrain each control point of the trajectory to be on the valid sides of the shifted hyperplanes (Line~\ref{Line:WorkspaceConstraints}, Algorithm~\ref{Algorithm:TrajectoryOptimization}).
Since B\'ezier curves are contained in the convex hulls of their control points, constraining control points to be in a convex set automatically constrains the B\'ezier curves to stay within the same convex set.

\emph{2) Robot-robot collision avoidance constraints:}
For robot-robot collision avoidance, recall %we utilize the fact 
that each robot replans with the same period $\delta t$ and planning is synchronized between robots.
    
At each iteration, the ego robot computes its SVM cell within the SVM tessellation of robots using hard-margin SVMs.
SVM tessellation is similar to Voronoi tesselation, the only difference is that  pairwise SVM hyperplanes are computed between collision shapes instead of Voronoi hyperplanes.
We choose SVM tessellation because i) hard-margin SVM is convex, hence pairs of robots can compute the same exact hyperplane under the assumption of perfect sensing, ii) it allows for a richer set of collision shapes than basic Voronoi cells, which is valid only for hyperspherical objects, and iii) SVM cells are always convex unlike generalized Voronoi cells.

We buffer the ego robot's SVM cell to account for its collision shape $\mR$, and constrain the first piece of the trajectory to stay inside the buffered SVM cell (BSVM) (loop at Line~\ref{Loop:RobotRobotCA}, Algorithm~\ref{Algorithm:TrajectoryOptimization}). 
Only the first piece of the trajectory is constrained to remain in the buffered SVM cell, since the entire planning pipeline is run after $\delta t$, which is smaller than the duration $s$ of the first piece. At that time, planning begins at a new location, generating a new first piece that must remain in the new buffered SVM cell. 

Buffering is achieved by changing the offset of the hyperplane.
$\mR(\vx)$ is the shape of the robot when placed at $\vx$, defined as $\mR(\vx) = \{\vx\}\oplus \mR_0$, where $\mR_0$ is the shape of the robot when placed the origin and $\oplus$ is the Minkowski sum operator.
Given a hyperplane with normal $\mH_\vn$ and offset $\mH_a$, we find $\mH_{a'}$ that ensures $\mR(\vx)$ is on the negative side of the hyperplane with normal $\mH_\vn$ and offset $\mH_a$ whenever $\vx$ is on the negative side of the hyperplane with normal $\mH_\vn$ and offset $\mH_{a'}$, and vice versa, by setting $\mH_{a'} = \mH_a + \max_{\vy\in \mR_0}\mH_\vn \cdot\vy$.
The following shows that whenever $\mR(\vx)$ is on the negative side of the hyperplane $(\mH_\vn, \mH_a)$, $\vx$ is on the negative side of the hyperplane $(\mH_\vn, \mH_{a'})$.
The converse can be shown by following the steps backwards.
\begin{align*}
    &\forall \vy \in \mR(\vx)\ \mH_\vn\cdot \vy + \mH_a \leq 0\\
    &\implies \max_{\vy \in \mR(\vx)}\mH_\vn\cdot \vy + \mH_a \leq 0\\
    &\implies \max_{\vy \in \{\vx\} \oplus \mR_0}\mH_\vn\cdot \vy + \mH_a \leq 0\\
    &\implies \max_{\vy \in \mR_0}\mH_\vn\cdot (\vy+\vx) + \mH_a \leq 0\\
    &\implies \mH_\vn\cdot \vx + \mH_a + \max_{\vy\in \mR_0}\mH_\vn \cdot\vy \leq 0
\end{align*}

Since the duration of the first piece was set to the safety duration $s\geq \delta t$ in discrete planning, the robot stays within its buffered SVM cell for at least $\delta t$. 
Moreover, since planning is synchronized across all robots, the pairwise SVM hyperplanes they compute will match, thus the buffered SVM cells of robots are disjoint.
This ensures robot-robot collision avoidance until the next planning iteration. % assuming that each robot is cooperative.

\begin{figure*}
    \centering
     \subfloat[Discrete Paths]{%
       \includegraphics[width=0.31\linewidth]{7-discrete-plans.pdf}
     }\hspace{0.01\linewidth}
     \subfloat[Robot-robot Collision Avoidance Constraints\label{Figure:RobotRobotCA}]{%
       \includegraphics[width=0.31\linewidth]{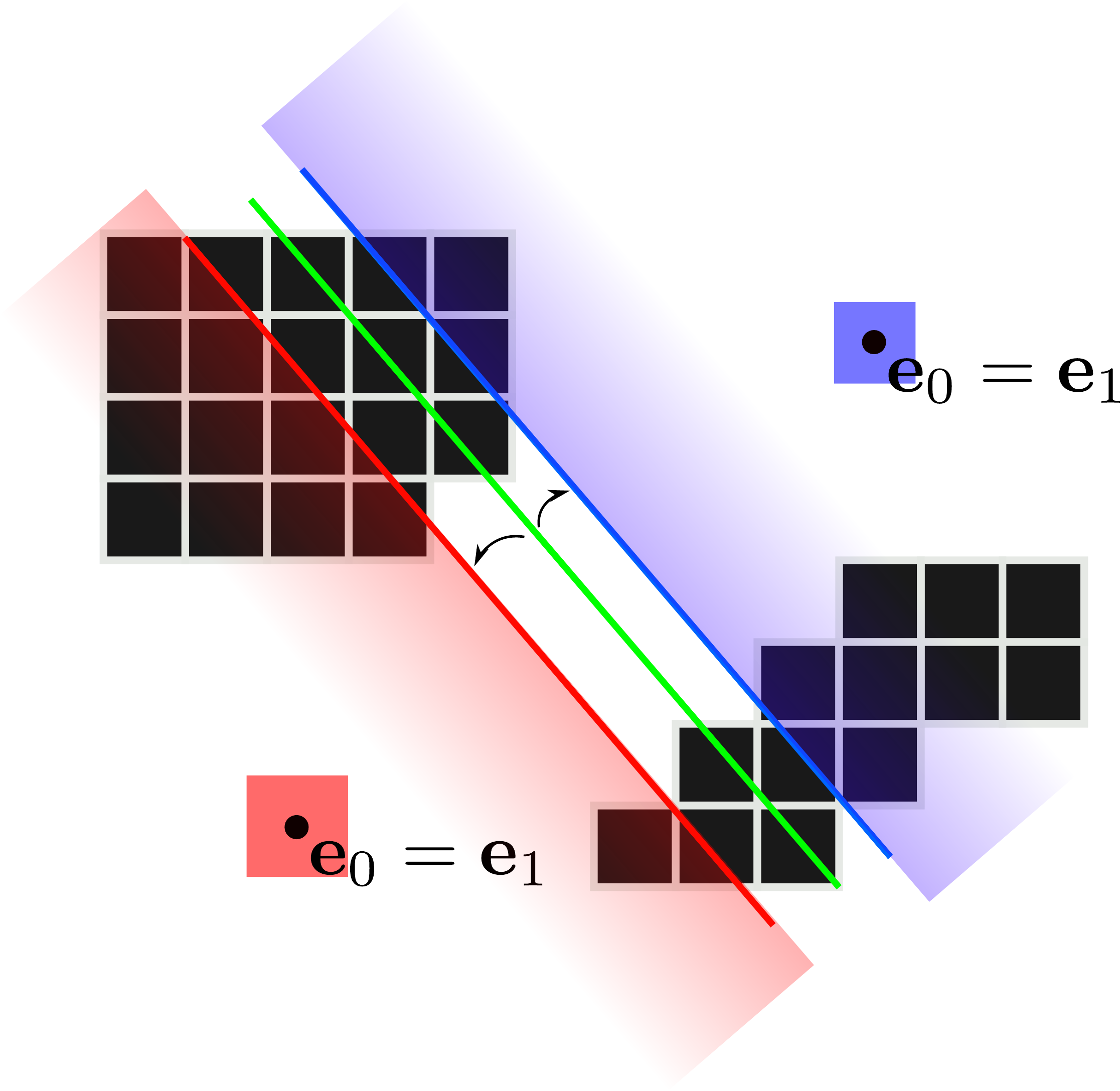}
     }\hspace{0.01\linewidth}
     \subfloat[Active Set of Robot-obstacle Collision Avoidance Constraints for the Second Piece of the Blue Robot\label{Figure:RobotObstacleCA1}]{%
       \includegraphics[width=0.31\linewidth]{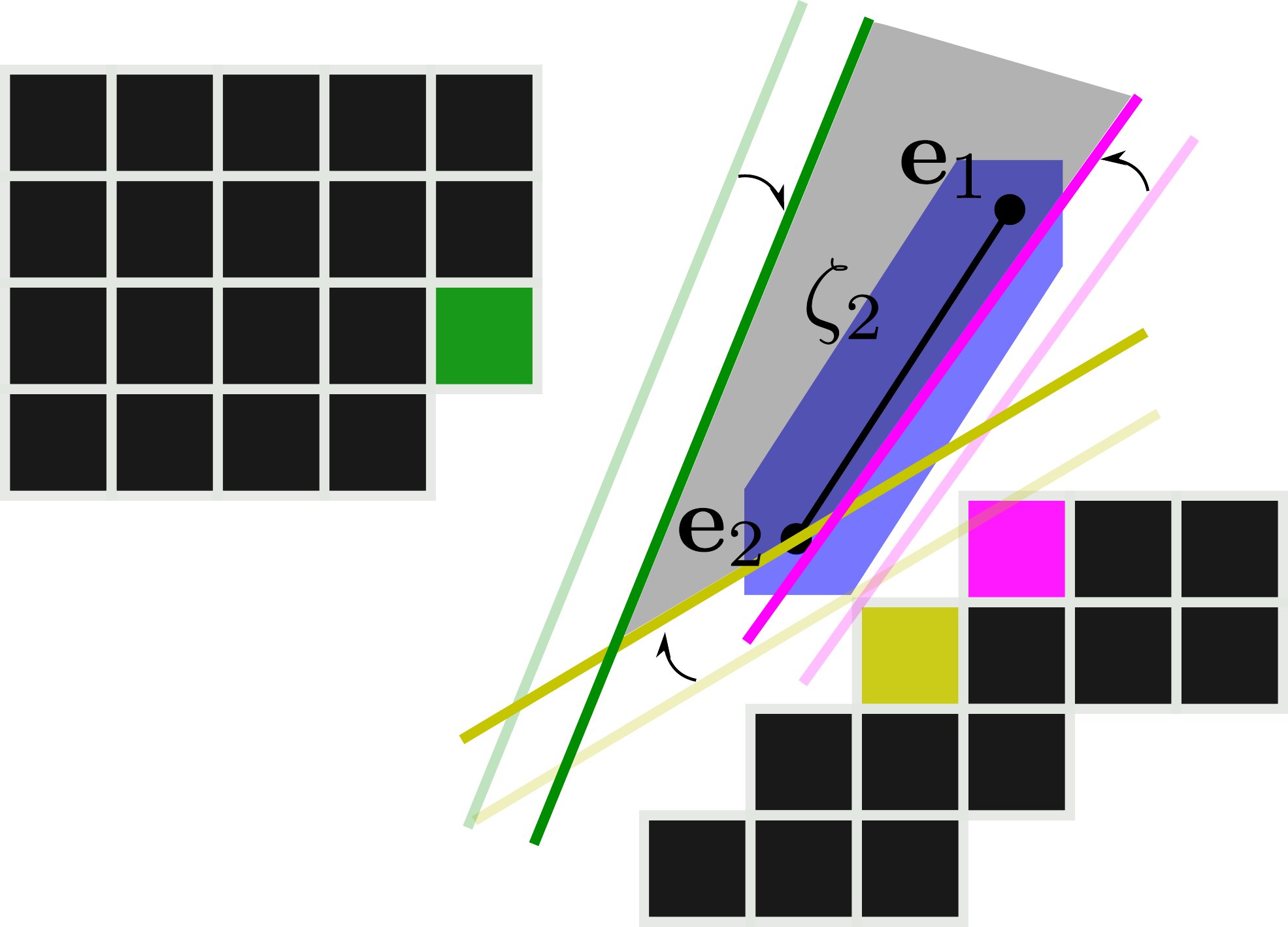}
     }\hfill
     \subfloat[Active Set of Robot-obstacle Collision Avoidance Constraints for the Second Piece of the Red Robot\label{Figure:RobotObstacleCA2}]{%
       \includegraphics[width=0.31\linewidth]{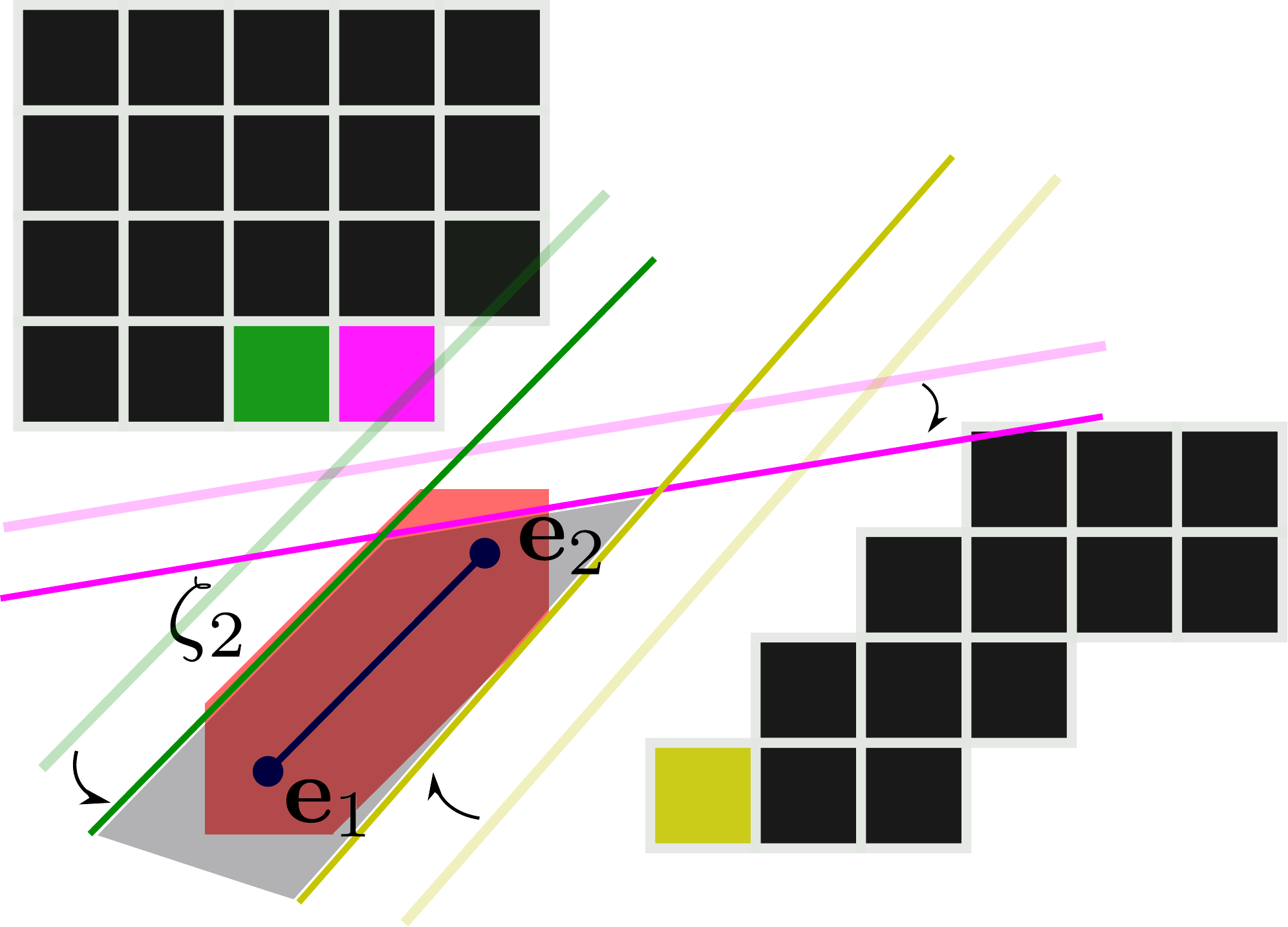}
     }\hspace{0.01\linewidth}
    \subfloat[Active Set of Robot-obstacle Collision Avoidance Constraints for the Third Piece of the Red Robot\label{Figure:RobotObstacleCA3}]{%
       \includegraphics[width=0.31\linewidth]{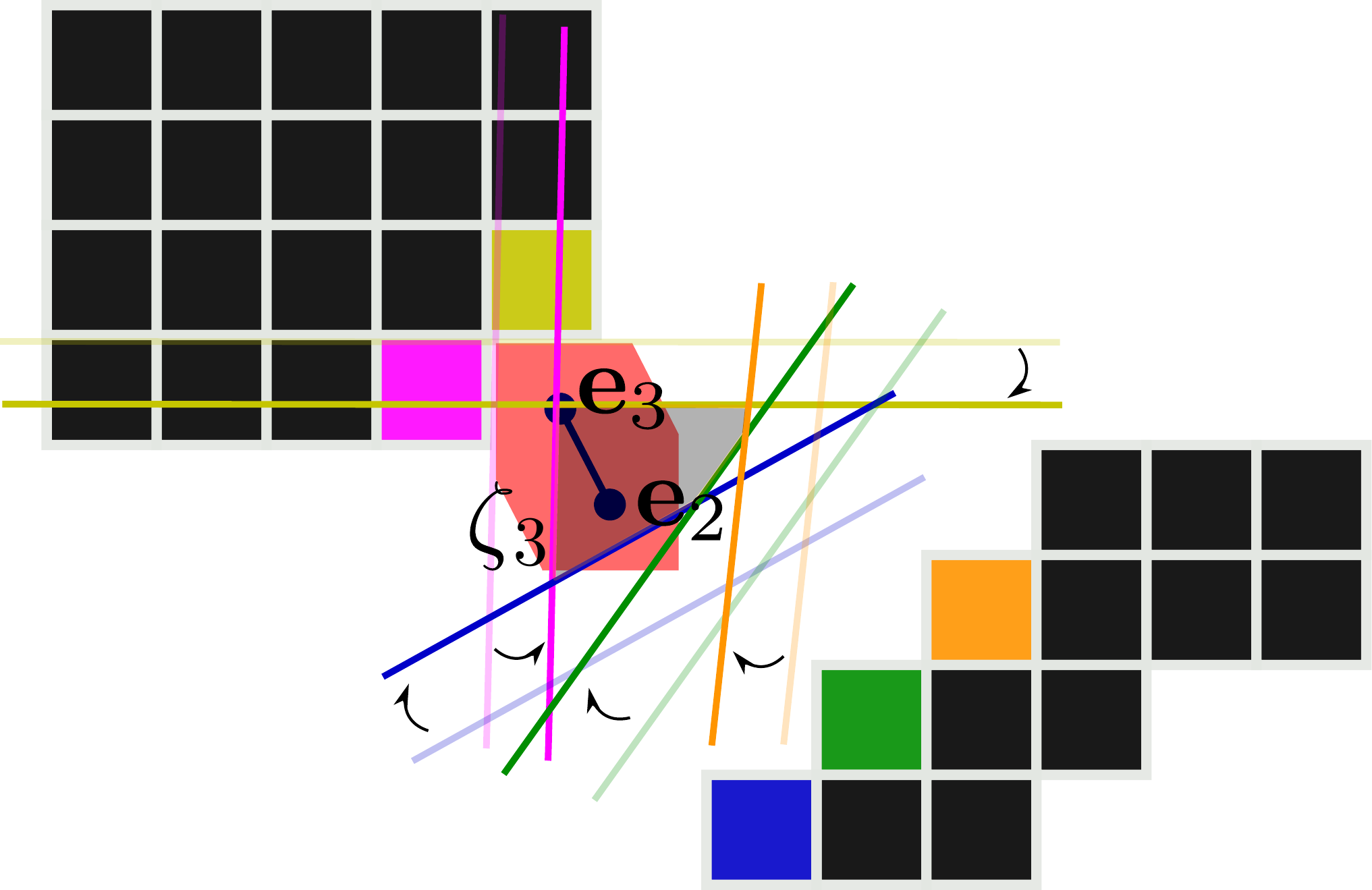}
     }\hspace{0.01\linewidth}
     \subfloat[Computed Trajectories \label{Figure:Trajectories}]{%
       \includegraphics[width=0.31\linewidth]{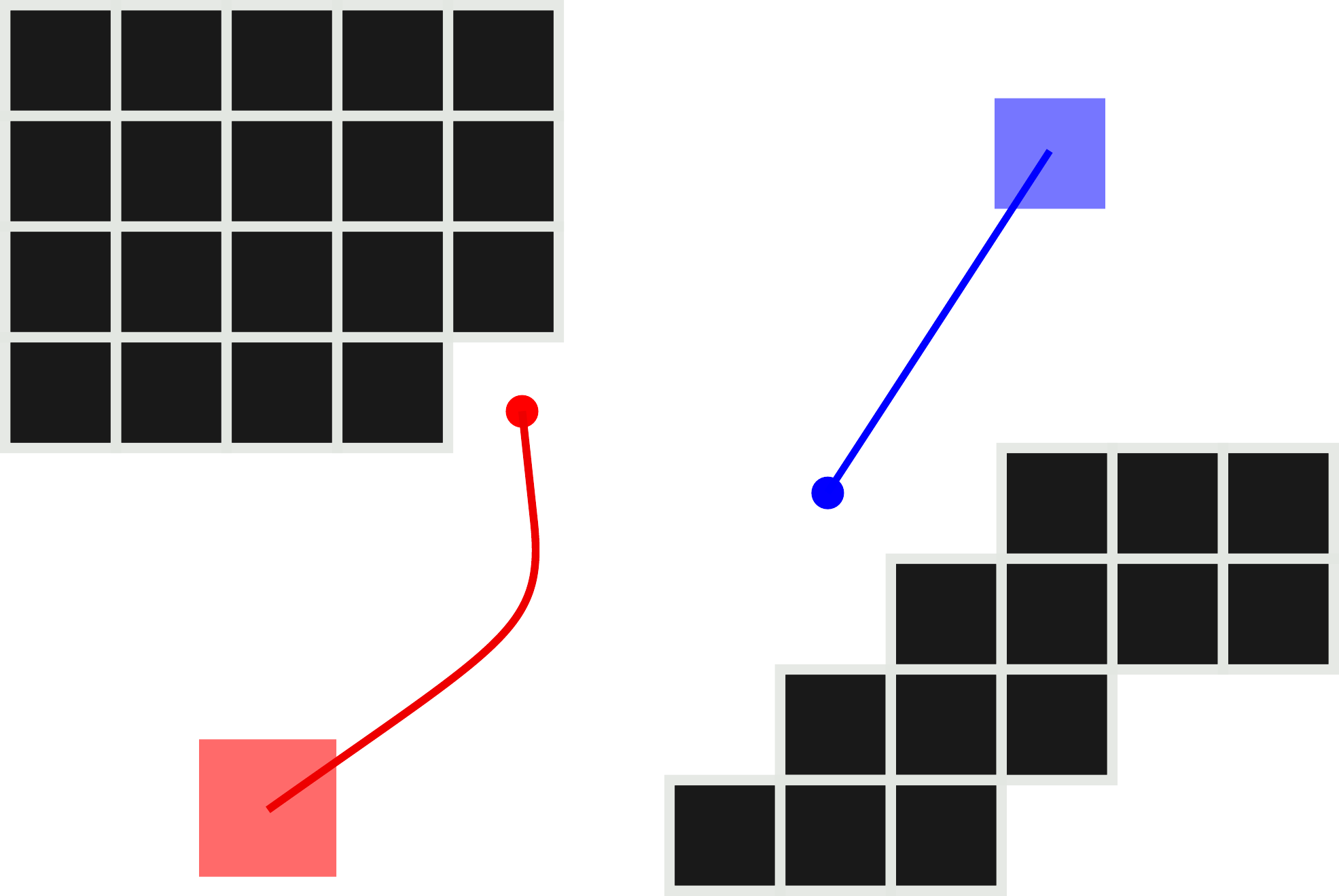}
     }\hfill
     \caption{\emph{Trajectory Optimization.} a) Discrete segments of two robots computed at the discrete planning stage. b) Robot-robot collision avoidance constraints are computed using BSVMs. The green hyperplane is the SVM hyperplane between two robots. Each robot shifts the SVM hyperplane to account for its geometry, and constrains the first piece of the trajectory with the resulting BSVM. c-d-e) Active set of robot-obstacle collision avoidance constraints for three different pieces (one belonging to the blue robot, two belonging to the red robot). In each figure, the region swept by the robot while traversing the segment is shown in robot's color. SVM hyperplanes between the swept region and the obstacles are given as light-colored lines. SVM hyperplanes are buffered to account for the robot's collision shape and shown as dark-colored lines (BSVMs). The shift operations are shown as arrows. Obstacles and constraints generated from them are colored using the same colors. For each piece, the feasible region that is induced by the robot-obstacle collision avoidance constraints is colored in gray. f) Trajectories computed by the trajectory optimization stage are given.}
     \label{Figure:TrajectoryOptimization}
\end{figure*}

Computed SVMs and BSVMs are shown in Fig.~\ref{Figure:RobotRobotCA} for a two robot case.

To ensure that the number of constraints of the optimization problem does not grow indefinitely, SVM hyperplanes are only computed against those robots that are at most $\tilde{r}$ away from the ego robot.
This does not result in unsafe trajectories so long as $\tilde{r}$ is more than the total maximum distance that can be traversed by two robots while following the first pieces of their trajectories, for which an upper bound is $\max_{i,j\in\{1,\ldots,N\}}(\gamma_i^1s_i+\gamma_j^1s_j)$, where $s_i$ and $s_j$ are the durations of the first pieces of the trajectories of robots $i$ and $j$ respectively.

\emph{3) Robot-obstacle collision avoidance constraints}:
Buffered SVM hyperplanes are used for robot-obstacle collision avoidance as well.
Let $\mathcal{\zeta}_i \subset \mathbb{R}^d$ be the region swept by the ego robot while traversing the $i^{th}$ segment from $\ve_{i-1}$ to $\ve_{i}$.
We compute the SVM hyperplane between $\mathcal{\zeta}_i$ and each object in $\mO$, buffer it as explained before to account for robot's collision shape, and constrain the $i^{th}$ piece by the resulting buffered SVM (Loop at line~\ref{Loop:RobotObstacleCA}, Algorithm~\ref{Algorithm:TrajectoryOptimization}).
This ensures that trajectory pieces do not cause collisions with obstacles.

The use of SVMs for collision avoidance against obstacles is a choice of convenience, as we already use them for robot-robot collision avoidance. For robot-obstacle collision avoidance, one can use any separating hyperplane between $\mathcal{\zeta}_i$ and objects in $\mO$, while in the case of robot-robot collision avoidance, pairs of robots must compute matching hyperplanes using the same algorithm.

Elements of the active set of SVM and BSVM hyperplanes for robot-obstacle collision avoidance are shown in an example scenario in Fig.~\ref{Figure:RobotObstacleCA1},~\ref{Figure:RobotObstacleCA2}, and~\ref{Figure:RobotObstacleCA3} for three different pieces.

Similar to robot-robot collision avoidance, to ensure that the number of constraints of the optimization problem does not grow indefinitely, we only compute SVM hyperplanes between $\mathcal{\zeta}_i$ and obstacles that are not more than $\tilde{o}$ away from $\mathcal{\zeta}_i$.

Let $\Upsilon_i$ be the set of buffered SVM hyperplanes that constrain the $i^{th}$ piece $\forall i\in\{1,\ldots,L\}$.
$\Upsilon_1$ contains both robot-robot and robot-obstacle collision avoidance hyperplanes while $\Upsilon_j\ \forall j\in\{2,\ldots,L\}$ contain only robot-obstacle collision avoidance hyperplanes. 
This is because the first piece is the only piece that we constrain with robot-robot collision avoidance hyperplanes because it is the only piece that will be executed until the next planning iteration.

\emph{4) Continuity constraints}:
We add two types of continuity constraints: i) continuity constraints between planning iterations, and ii) continuity constraints between trajectory pieces (Line~\ref{Line:ContinuityConstraints}, Algorithm~\ref{Algorithm:TrajectoryOptimization}).

To enforce continuity between planning iterations, we add constraints that enforce 
\begin{align*}
\frac{d^j\vf(0)}{dt^j} = \frac{d^j\vp}{dt^j}\ \forall j\in\{0,\ldots,c\} 
\end{align*}
where $c$ is the task input denoting the continuity degree up to which the resulting trajectory must be continuous and $\vp$ is the current position.

To enforce continuity between trajectory pieces, we add constraints that enforce
\begin{align*}
    \frac{d^j\vfx{i}(T_i)}{dt^j} &= \frac{d^j\vfx{i+1}(0)}{dt^j}\\ 
    &\forall i\in\{1,\ldots,L-1\}\ \forall j\in\{0,\ldots,c\}
\end{align*}
where $\vfx{i}(t)$ is the $i^{th}$ piece of the trajectory.

Remark~\ref{Remark:SVMFeasibility} discusses that the SVM problems generated during trajectory optimization are feasible, i.e., the trajectory optimization stage will always succeed constructing the QP.

\begin{remark}\label{Remark:SVMFeasibility}
All SVM problems generated for robot-robot and robot-obstacle collision avoidance from the discrete path outputted by the discrete planning stage are feasible.
\begin{proof}[Reasoning.]\renewcommand{\qedsymbol}{}
The discrete planning stage outputs a discrete path such that a robot with collision shape $\mR$ following the path does not collide with any obstacles in $\mO$ or any other robots in $\mS$.
It also ensures that $\vp=\ve_0$ since the search starts from the robot's current position $\vp$.
Hence, $\mR(\vp)$ does not intersect with any $\mS_j \in \mS$.
Since each robot is assumed to be convex, there exists at least one hyperplane that separates $\mR(\vp)$ from $\mS_j$ for each $j$ by the separating hyperplane theorem.
SVM is an optimal separating hyperplane according to a cost function.
Therefore, SVM problems between $\mR(\vp)$ and $\mS_j\in\mS$ for robot-robot collision avoidance are feasible.

Since the robot moving along the discrete segments is collision free, $\mathcal{\zeta}_i$ is collision free for all $i$.
Also, since it is a sweep of a convex shape along a line segment, $\mathcal{\zeta}_i$ is convex as shown in Lemma~\ref{Lemma:CvxSweep} in Appendix~\ref{Appendix:SweepConvex}.
Similar to the previous argument, since $\mathcal{\zeta}_i$ is collision-free and convex and all obstacles in the environment are convex, each SVM problem between $\mathcal{\zeta}_i$ and $\mQ\in\mO$ for robot-obstacle collision avoidance is feasible by the separating hyperplane theorem.
\end{proof}
\end{remark}

Workspace, robot-robot collision avoidance, robot-obstacle collision avoidance, and position continuity constraints are kinematic constraints. Higher-order continuity constraints are dynamic constraints.
Remark~\ref{Remark:KinematicFeasibility} discusses the ensured feasibility of the kinematic constraints.
The feasibility of the dynamic constraints cannot be ensured for arbitrary degrees of continuity.

\begin{remark}
\label{Remark:KinematicFeasibility}
Kinematic constraints of the optimization problem generated from a discrete path output from the discrete planning stage are feasible for the same path when the degree of B\'ezier curves $h\geq 1$.
\begin{proof}[Reasoning.]\renewcommand{\qedsymbol}{}
Any B\'ezier curve with degree $h\geq 1$ can represent a discrete segment by setting half of the points to the start of the segment and other half to the end of the segment.
Hence, we will only show that a discrete path output from discrete planning stage by itself satisfies the kinematic constraints generated.
Remember that discrete path output from the discrete planning stage has the property $\vp=\ve_0=\ve_1$.

Each robot-robot SVM problem is feasible (see Remark~\ref{Remark:SVMFeasibility}).
Let $\mH_\vn$ be the normal and $\mH_a$ be the offset of any of the robot-robot SVMs.
It is shifted by setting the offset to $\mH_{a'} = \mH_a + \max_{\vy\in \mR_0}\mH_\vn \cdot\vy$.
The point $\vp$ satisfies $\mH_\vn \cdot \vp + \mH_{a'} \leq 0$ because $\mR(\vp)$ is on the negative side of the SVM.
The robot-robot BSVM hyperplanes are used to constrain the first piece of the trajectory, and setting the first piece as a $0$-length segment with $\vp=\ve_0=\ve_1$ satisfies the robot-robot collision avoidance constraints.

Each robot-obstacle SVM problem is feasible (see Remark~\ref{Remark:SVMFeasibility}).
Let $\mH_\vn$ be the normal and $\mH_a$ be the offset of any of the robot-obstacle SVMs that is between $\mathcal{\zeta}_i$ and an obstacle.
It is shifted by setting the offset to $\mH_{a'} = \mH_a + \max_{\vy\in \mR_0}\mH_\vn \cdot\vy$.
All points on the line segment $\vp_i(t) = \ve_{i-1} + t(\ve_i - \ve_{i-1}), t\in[0,1]$ from $\ve_{i-1}$ to $\ve_i$ satisfy the BSVM constraint because $\mR(\vp_i(t))$ is on the negative side of the SVM hyperplane $\forall t\in[0,1]$.
Since each SVM hyperplane between $\mathcal{\zeta}_i$ and obstacles is only used to constrain the $i^{th}$ piece, constraints generated by it are feasible for the segment from $\ve_{i-1}$ to $\ve_i$.

The feasibility of the workspace constraints are trivial since the robot moving along discrete segments is contained in the workspace, and we shift bounding hyperplanes of the workspace in the same way as SVM hyperplanes.
Hence, the discrete path satisfies the workspace constraints.

Initial point position continuity of the robot is satisfied by the given discrete segments since $\vp=\ve_0$.
Position continuity between segments are trivially satisfied by the given discrete segments, since the discrete path is position continuous by its definition.
\end{proof}
\end{remark}

\subsubsection{Cost Function}

The cost function of the optimization problem has $3$ terms: i) energy usage, ii) deviation from the discrete path, and iii) preferred distance to objects.

We use the sum of integrated squared derivative magnitudes as a metric for energy usage (Line~\ref{Line:EnergyCost}, Algortihm~\ref{Algorithm:TrajectoryOptimization}), similar to the works of ~\cite{richter2013planning} and~\cite{honig2018quadswarms}.
Parameters $\vlambda = \{\lambda_j\}$ are the weights for integrated squared derivative magnitudes, where $\lambda_j$ is the weight for $j^{th}$ degree of derivative.
The energy term $\mJ_{energy}(\mP)$ is given as
\begin{align*}
    \mJ_{energy}(\mP) = \sum_{\lambda_j \in \vlambda} \lambda_j \int_0^T\normtwo{\frac{d^j\vf(t)}{dt^j}}^2dt.
\end{align*} 

We use squared Euclidean distances between trajectory piece endpoints and discrete segment endpoints as a metric for deviation from the discrete path (Line~\ref{Line:DeviationCost}, Algorithm~\ref{Algorithm:TrajectoryOptimization}).
Remember that each B\'ezier curve piece $i$ ends at its last control point $\vPxy{i}{h}$.
Parameters $\vtheta =\{\theta_i\}\in \mathbb{R}^L$ are the weights for each segment endpoint.
The deviation term $\mJ_{dev}(\mP)$ is given as
\begin{align*}
    \mJ_{dev}(\mP) = \sum_{i\in\{1,\ldots,L\}} \theta_i\normtwo{\vPxy{i}{h} - \ve_i}^2.
\end{align*}

The last term of the cost function models the preferred distance to objects.
We use this term to discourage robots from getting closer to other objects in the environment; this increases the numerical stability of the algorithm by driving robots away from tight states.

We shift each hyperplane in $\Upsilon_1$ (i.e., buffered SVM hyperplanes constraining the first piece) by the preferred distance to objects, $\tilde{p}$, to create the preferred hyperplanes $\overline{\Upsilon}_1$ (Line~\ref{Line:PreferredShift}, Algorithm~\ref{Algorithm:TrajectoryOptimization}).
Since each robot replans with the period $\delta t$, we add a cost term that drives the robot closer to the preferred hyperplanes at the replanning period (Line~\ref{Line:PreferredCost}, Algorithm~\ref{Algorithm:TrajectoryOptimization}).
We take the sum of squared distances between $\vf(\delta t)$ and hyperplanes in $\overline{\Upsilon}_1$:
\begin{align*}
    \mJ_{pref}(\mP) = \alpha\sum_{\mH \in \overline{\Upsilon}_1} \normtwo{\mH_{\vn}^\top \vf(\delta t) + \mH_a}^2
\end{align*}
where $\mH_{\vn}$ is the normal and $\mH_a$ is the offset of hyperplane $\mH$, and $\alpha$ is the weight of $\mJ_{pref}$ term.
Notice that this term is defined over the control points of first piece since $\delta t\leq s = T_1$, supporting the utilization of $\Upsilon_1$ only.

The overall trajectory optimization problem is:
\begin{align*}
    \min_{\mP}\ &\mJ_{energy}(\mP) + \mJ_{dev}(\mP) + \mJ_{pref}(\mP) \text{ s.t}.\\
    & \mH_{\vn}^\top\vPxy{i}{j} + \mH_{a} \leq 0 \ \mkern25mu\forall i\in\{1, \ldots, L\}\\
    &\mkern168mu\forall j\in\{0,\ldots,h\} \\
    &\mkern168mu\forall\mH\in\Upsilon_i \cup \Upsilon_\mW\\
    &\frac{d^j\vf(0)}{dt^j} = \frac{d^j\vp}{dt^j}\mkern59mu\forall j\in\{0,\ldots,c\} \\
    &\frac{d^j\vfx{i}(T_i)}{dt^j} = \frac{d^j\vfx{i+1}(0)}{dt^j}\ \forall i\in\{1,\ldots,L-1\}\ \\
    &\mkern169mu\forall j\in\{0,\ldots,c\}.
\end{align*}

Notice that we formulate continuity and the safety of the trajectory using hard constraints.
This ensures that the resulting trajectory is kinematically safe and continuous up to degree $c$ if the optimization succeeds for all robots and planning is synchronized.

\subsection{Temporal Rescaling}

At the temporal rescaling stage, we check whether the dynamic limits of the robot are violated.
If the resulting trajectory is valid, i.e. derivative magnitudes of the trajectory are less than or equal to the maximum derivative magnitudes $\gamma^k\ \forall k\in\{1,\ldots,K\}$, the trajectory is returned as the output of RLSS.
If not, temporal rescaling is applied to the trajectory similar to~\cite{honig2018quadswarms} and~\cite{park2020relbern} by increasing the durations of the pieces so that the trajectory is valid.
We scale the durations of pieces by multiplying them with a constant parameter greater than $1$ until the dynamic limits are obeyed.

We choose to enforce dynamic feasibility outside of the trajectory optimization problem as a post processing step because i) the output of the trajectory optimization stage is often dynamically feasible, hence rescaling is rarely needed, and ii) adding piece durations as variables to the optimization problem would make it a non quadratic program, potentially decreasing performance.

\section{Evaluation}

Here, using synchronized simulations, we first evaluate our algorithm's performance when different parameters are used in Section~\ref{Section:EffectsOfSelectedParamters}.
Second, we conduct an ablation study to show the effects of two important steps, namely the prepend operation of the discrete planning stage and the preferred distance cost term of the trajectory optimization stage, to the performance of the algorithm in Section~\ref{Section:AblationStudy}.
Third, we compare our algorithm to two state-of-the-art baseline planners in Section~\ref{Section:BaselineComparison}.
Finally, we show our algorithm's applicability to real robots by using it on quadrotors and differential drive robots in Section~\ref{Section:RealRobotExperiments}.

We conduct our simulations on a laptop computer with Intel i7-8565U @ $\SI{1.80}{GHz}$ running Ubuntu 20.04.
We implement our algorithm for a single core because of implementation simplicity and fairness to the baseline planners, which are not parallelized.
The memory usage of each simulation is $\SI{30}{MB}$ on average for our algorithm.

In synchronized simulations, we compute trajectories for each robot using the same snapshot of the environment, move robots perfectly according to computed trajectories for the re-planning period, and replan.
The effects of planning iterations taking longer than the re-planning period are not modeled in the synchronized simulations.
We show that all algorithms (RLSS and the baselines) can work in $\SI{1}{Hz} - \SI{10}{Hz}$ on the hardware we use, and assert that more powerful computers can be used to shorten planning times.
In addition, parallelization of the A* search on GPUs is possible.
For example, \cite{zhou2015massively} show the possibility of 6 -- 7 x speedup in A* search for pathfinding problems on GPUs.
Moreover, parallelization of quadratic program solving is possible through i) running multiple competing solvers in multiple cores and returning the answer from the first one that solves the problem, or ii) parallelizing individual solvers.
For instance, IBM ILOG CPLEX Optimizer\footnote{\url{https://www.ibm.com/docs/en/icos/22.1.0?topic=optimizers-concurrent-optimizer-in-parallel}} and Gurobi Optimizer\footnote{\url{https://www.gurobi.com/documentation/9.5/refman/concurrent_environments.html}} support running multiple competing solvers concurrently.
IBM ILOG CPLEX Optimizer supports a parallel barrier optimizer\footnote{\url{https://www.ibm.com/docs/en/icos/22.1.0?topic=optimizers-using-parallel-barrier-optimizer}}, which parallelizes a barrier algorithm for solving QPs.
~\cite{gondzio2009parallel} propose a parallel interior point method solver that exploits nested block structure of problems.
Performance improvements of these approaches are problem dependent, and we do not investigate how much the performance of our trajectory optimization stage could be improved with such methods.

In all experiments, $32$ robots are placed in a circle formation of radius $\SI{20}{m}$ in 3D, and the task is to swap all robots to the antipodal points on the circle.
%that are in the opposite sides of the circle.
The workspace $\mW$ is set to an axis aligned bounding box from $\begin{bmatrix}\SI{-25}{m} & \SI{-25}{m} & \SI{0}{m}\end{bmatrix}^\top$ to $\begin{bmatrix}\SI{25}{m} & \SI{25}{m} & \SI{5}{m}\end{bmatrix}^\top$.
Robot collision shapes are modeled as axis aligned cubes with $\SI{0.2}{m}$ edge lengths.
The desired planning horizon $\tau$ is set to $\SI{5}{s}$.
The safety distance $D$ of goal selection is set to $\SI{0.2}{m}$.
Robots have velocity limit $\gamma^1 = \SI{3.67}{\frac{m}{s}}$, and accelaration limit $\gamma^2 = \SI{4.88}{\frac{m}{s^2}}$, which are chosen arbitrarily.
The safety duration $s$ is set to $\SI{0.11}{s}$ and re-planning period $\delta t$ is set to $\SI{0.1}{s}$.
We set integrated derivative cost weights $\lambda_1 = 2.0$ for velocity and $\lambda_2 = 2.8$ for acceleration.
We set endpoint cost weights to $\theta_1 = 0$, $\theta_2 = 150$, $\theta_3 = 240$, $\theta_i = 300\ \forall i \geq 4$.
Setting $\theta_1=0$ allows the optimization to stretch the first $0$-length segment freely, and setting other $\theta$s incrementally increases the importance of tail segments.
We set the preferred distance to objects to $\tilde{p}=\SI{0.6}{m}$ and the preferred distance cost weight $\alpha = 0.3$.
We use the OcTree data structure from the octomap library~\citep{hornung2013octomap} to represent the environment. 
Each leaf-level occupied axis aligned box of the OcTree is used as a separate obstacle in all algorithms.
OcTree allows fast axis aligned box queries which return obstacles that intersects with a given axis aligned box, an operation we use extensively in our implementation.
For example, we use axis aligned box queries in discrete planning as broadphase collision checkers to find the obstacles that are close to the volume swept by the ego robot while traversing a given segment, and check collisions between the robot traversing the segment and only the obstacles returned from the query.
% Also, we used axis aligned box queries during constraint generation in the trajectory optimization stage.
Also, to generate robot-obstacle collision avoidance constraints, we execute axis aligned box queries around the segments to retrieve nearby obstacles.
We generate BSVM constraints only against nearby obstacles that are no more than $\tilde{o}$ away from the robot traversing the segment.
Other popular approaches for environment representations include i) Euclidean signed distance fields (ESDFs)~\citep{oleynikova2017voxblox}, which support fast distance queries to nearest obstacles, ii) 3D circular buffers~\citep{usenko2017real}, which aim to limit memory usage of maps and supports fast occupancy checks, and iii) Gaussian mixture models~\citep{meadhra2019gmm}, which continuously represent occupancy instead of discretizing the environment as the former approaches do.
None of these representations are as suitable as OcTrees for RLSS since they do not allow fast querying of obstacles in the vicinity of segments.
We use the IBM ILOG CPLEX Optimizer\footnote{\url{https://www.ibm.com/analytics/cplex-optimizer}} to solve our optimization problems.
In all experiments, robots that are closer than $\SI{0.25}{m}$ to their goal positions are considered as goal reaching robots.
If a robot that has not reached its goal does not change its position more than $\SI{1}{cm}$ in the last $\SI{1}{s}$ of the simulation, it is considered as a deadlocked robot.
At any point of the simulation, if each robot is either at the goal or deadlocked, we conclude the simulation.

\subsection{Effects of Selected Parameters}\label{Section:EffectsOfSelectedParamters}

We evaluate the performance of our algorithm when $4$ important parameters are changed: step size $\sigma$ of the search grid, degree $h$ of B\'ezier curves, obstacle check distance $\tilde{o}$, and robot check distance $\tilde{r}$.
Step size $\sigma$ of the search grid is the parameter that affects discrete planning performance most because it determines the amount of movement at each step during the $A^*$ search.
The degree $h$ of B\'ezier curves is important in trajectory optimization because it determines the number of decision variables.
The obstacle check distance $\tilde{o}$ and robot check distance $\tilde{r}$ determine the number of collision avoidance constraints in the optimization problem.

In all of the parameter evaluations, we use a random 3D forest environment with OcTree resolution $\SI{0.5}{m}$ in which $\SI{10}{\%}$ of the environment is occupied.
There are $2332$ leaf-level boxes in OcTree, translating to $2332$ obstacles in total.
We set the desired trajectory of each robot to the straight line segment that connects the robot's start and goal positions.
We set the duration of the segment to the length of the line segment divided by the maximum velocity of the robot.
We enforce continuity up to velocity, hence set $c=1$.

In our experiments, our algorithm does not result in any collisions or deadlocks.

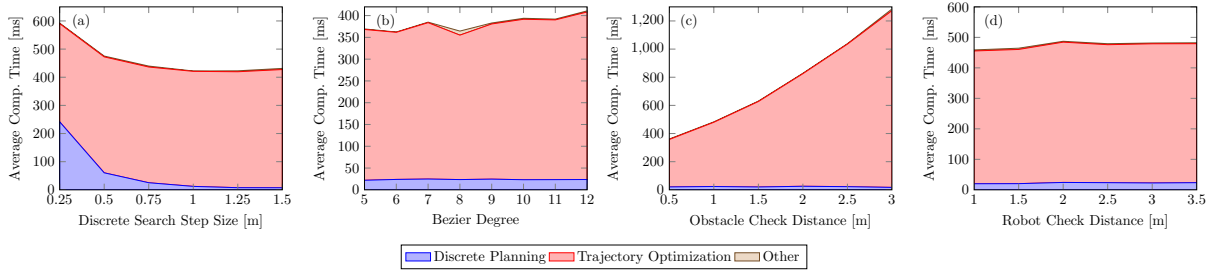
\begin{figure*}
    \centering
    \resizebox{\textwidth}{!}{\begin{tikzpicture}[]
        \begin{axis}[
        in=x,
        table/col sep=comma,
        xlabel={Discrete Search Step Size [m]},
        ylabel={Average Comp. Time [ms]},
        xmin={0.25},
        xmax={1.5},
        ymin=0,
        ymax=650,
        xtick=data,
        max space between ticks=20,
        stack plots=y,%   
        area style,
        legend style={at={(0.5,-0.25)},anchor=north},
        name=ss
        ]
            \addplot table [mark=none,x=step_size,y=discrete_search] {step_size_duration.csv}
            \closedcycle;
            \addlegendentry{Discrete Planning};
            % \closedcycle;
            \addplot table [mark=none,x=step_size,y=traj_opt] {step_size_duration.csv}
            \closedcycle;
            \addlegendentry{Trajectory Optimization};
            \addplot table [mark=none,x=step_size,y=other] {step_size_duration.csv}
            \closedcycle;
            \addlegendentry{Other};
            \legend{}
        \end{axis}
        \begin{axis}[
        in=x,
        xshift=2cm,
        table/col sep=comma,
        xlabel={Bezier Degree},
        ylabel={Average Comp. Time [ms]},
        xmin={5},
        xmax={12},
        ymin=0,
        ymax=420,
        max space between ticks=20,
        stack plots=y,%   
        area style,
        at={(ss.south east)},
        name=bd
        ]
            \addplot table [mark=none,x=bezier_degree,y=discrete_search] {bezier_duration.csv}
            \closedcycle;
            \addlegendentry{Discrete Planning};
            % \closedcycle;
            \addplot table [mark=none,x=bezier_degree,y=traj_opt] {bezier_duration.csv}
            \closedcycle;
            \addlegendentry{Trajectory Optimization};
            \addplot table [mark=none,x=bezier_degree,y=other] {bezier_duration.csv}
            \closedcycle;
            \addlegendentry{Other};
            \legend{}
        \end{axis}
        \begin{axis}[
        in=x,
        table/col sep=comma,
        xshift=2cm,
        xlabel={Obstacle Check Distance [m]},
        ylabel={Average Comp. Time [ms]},
        xmin={0.5},
        xmax={3.0},
        ymin=0,
        ymax=1300,
        xtick=data,
        max space between ticks=20,
        stack plots=y,%   
        area style,
        at={(bd.south east)},
        name=oc
        ]
            \addplot table [mark=none,x=obs_check_distance,y=discrete_search] {obs_check_distance_duration.csv}
            \closedcycle;
            \addlegendentry{Discrete Planning};
            % \closedcycle;
            \addplot table [mark=none,x=obs_check_distance,y=traj_opt] {obs_check_distance_duration.csv}
            \closedcycle;
            \addlegendentry{Trajectory Optimization};
            \addplot table [mark=none,x=obs_check_distance,y=other] {obs_check_distance_duration.csv}
            \closedcycle;
            \addlegendentry{Other};
            \legend{}
        \end{axis}
        \begin{axis}[
        in=x,
        table/col sep=comma,
        xshift=2cm,
        xlabel={Robot Check Distance [m]},
        ylabel={Average Comp. Time [ms]},
        xmin={1.0},
        xmax={3.5},
        ymin=0,
        ymax=600,
        xtick=data,
        max space between ticks=20,
        stack plots=y,%   
        area style,
        at={(oc.south east)},
        legend style={at={(-0.75,-0.3)},anchor=north east,legend columns=3},
        ]
            \addplot table [mark=none,x=robot_check_distance,y=discrete_search] {robot_check_distance_duration.csv}
            \closedcycle;
            \addlegendentry{Discrete Planning};
            % \closedcycle;
            \addplot table [mark=none,x=robot_check_distance,y=traj_opt] {robot_check_distance_duration.csv}
            \closedcycle;
            \addlegendentry{Trajectory Optimization};
            \addplot table [mark=none,x=robot_check_distance,y=other] {robot_check_distance_duration.csv}
            \closedcycle;
            \addlegendentry{Other};
        \end{axis}
        \node[text width=0cm] at (0.3,4.1) {(a)};
        \node[text width=0cm] at (7.7,4.1) {(b)};
        \node[text width=0cm] at (15.1,4.1) {(c)};
        \node[text width=0cm] at (22.5,4.1) {(d)};
    \end{tikzpicture}}
    \caption{Average computation time per stage are given when different parameters are used. Goal selection and temporal rescaling steps are given as "other" since they take a small amount time compared to other two stages. All experiments are done in a 3D random forest environment with $\SI{10}{\%}$ occupancy.}
    \label{Figure:AverageCompTime}
\end{figure*}

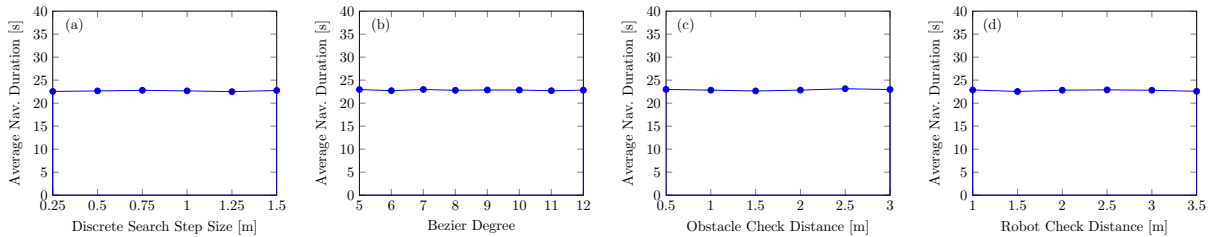
\begin{figure*}
    \centering
    \resizebox{\textwidth}{!}{\begin{tikzpicture}[]
        \begin{axis}[
        in=x,
        table/col sep=comma,
        xlabel={Discrete Search Step Size [m]},
        ylabel={Average Nav. Duration [s]},
        xmin={0.25},
        xmax={1.5},
        xtick=data,
        ymin=0,
        ymax=40,
        max space between ticks=20,
        stack plots=y,
        name=ss
        ]
            \addplot table [mark=dot,x=step_size,y=nav_dur] {step_size_nav_dur.csv}
            \closedcycle;
        \end{axis}
        \begin{axis}[
        in=x,
        xshift=2cm,
        table/col sep=comma,
        xlabel={Bezier Degree},
        ylabel={Average Nav. Duration [s]},
        xmin={5},
        xmax={12},
        ymin=0,
        ymax=40,
        max space between ticks=20,
        stack plots=y,
        at={(ss.south east)},
        name=bd
        ]
            \addplot table [mark=dot,x=bezier_degree,y=nav_dur] {bezier_navigation_duration.csv}
            \closedcycle;
        \end{axis}
        \begin{axis}[
        in=x,
        xshift=2cm,
        table/col sep=comma,
        xlabel={Obstacle Check Distance [m]},
        ylabel={Average Nav. Duration [s]},
        xmin={0.5},
        xmax={3},
        xtick=data,
        ymin=0,
        ymax=40,
        max space between ticks=20,
        stack plots=y,
        at={(bd.south east)},
        name=oc
        ]
            \addplot table [mark=dot,x=obstacle_check_distance,y=nav_dur] {obs_check_distance_nav_dur.csv}
            \closedcycle;
        \end{axis}
        \begin{axis}[
        in=x,
        xshift=2cm,
        table/col sep=comma,
        xlabel={Robot Check Distance [m]},
        ylabel={Average Nav. Duration [s]},
        xmin={1.0},
        xmax={3.5},
        xtick=data,
        ymin=0,
        ymax=40,
        max space between ticks=20,
        stack plots=y,% 
        at={(oc.south east)},
        ]
            \addplot table [mark=dot,x=robot_check_distance,y=nav_dur] {robot_check_distance_nav_dur.csv}
            \closedcycle;
        \end{axis}
        
        \node[text width=0cm] at (0.3,4.1) {(a)};
        \node[text width=0cm] at (7.7,4.1) {(b)};
        \node[text width=0cm] at (15.1,4.1) {(c)};
        \node[text width=0cm] at (22.5,4.1) {(d)};
    \end{tikzpicture}}
    \caption{Average navigation duration of robots from their start positions to their goal positions are given when the selected parameters are changed. In all cases, average navigation duration is not affected by the changes in the selected parameters in the chosen ranges.}
    \label{Figure:AverageNavDuration}
\end{figure*}

We report average computation time per iteration (Fig.~\ref{Figure:AverageCompTime}) and average navigation duration of robots (Fig.~\ref{Figure:AverageNavDuration}).
Average navigation duration is the summed total navigation time for all robots divided by the number of robots.

\subsubsection{Step Size $\sigma$ of Discrete Search}\label{Section:StepSize}

We evaluate our algorithm's performance when step size $\sigma$ is changed.
We set $\sigma$ to values between $\SI{0.25}{m}$ to $\SI{1.5}{m}$ with $\SI{0.25}{m}$ increments.
We set obstacle check distance $\tilde{o}=\SI{1.0}{m}$, robot check distance $\tilde{r}=\SI{2.0}{m}$, and degree of B\'ezier curves $h=12$ in all cases, which are determined by premiliminary experiments and the results of the experiments in Sections~\ref{Section:BezierCurvesDegrees},~\ref{Section:ObstacleCheckDistance}, and~\ref{Section:RobotCheckDistance}.
The results are summarized in Fig.~\ref{Figure:AverageCompTime}a and Fig.~\ref{Figure:AverageNavDuration}a.

As the step size gets smaller, discrete search takes more time; but the algorithm can still work in about \SI{2}{Hz} even when $\sigma = \SI{0.25}{m}$.
The average navigation duration of robots are close to $\SI{22.5}{s}$ in each case, suggesting the robustness of the algorithm to the changes in this parameter.
At $\sigma\geq0.75$, the time discrete planning takes is less than $\SI{6}{\%}$ of the time trajectory optimization takes.

We also run the algorithm with $\sigma = \SI{3}{m}, \sigma = \SI{6}{m},$ and $\sigma = \SI{12}{m}$, which decreases the flexibility of the discrete search considerably.
In all of those cases, discrete search results in fluctuations, and some robots get stuck in livelocks, in which they move between same set of positions without reaching to their goal positions.

\subsubsection{Degree $h$ of B\'ezier Curves}\label{Section:BezierCurvesDegrees}

Next, we evaluate our algorithm's performance when the degree $h$ of B\'ezier curves is changed.
We set $h$ to values in $\{5,\ldots,12\}$.
We set step size $\sigma=\SI{0.77}{m}$, obstacle check distance $\tilde{o}=\SI{1.0}{m}$, and robot check distance $\tilde{r}=\SI{2.0}{m}$, which are determined by premiliminary experiments and the results of the experiments in Sections~\ref{Section:StepSize},~\ref{Section:ObstacleCheckDistance}, and~\ref{Section:RobotCheckDistance}.
The results are summarized in Fig.~\ref{Figure:AverageCompTime}b and Fig.~\ref{Figure:AverageNavDuration}b.

Even if the degree of the B\'ezier curves determine the number of decision variables of the trajectory optimization, the computation time increase of the trajectory optimization stage is not more than $\SI{10}{\%}$ between degree $5$ B\'ezier curves and degree $12$ B\'ezier curves. 
Also, average navigation duration of robots are close to $\SI{22.5}{m}$ in each case, suggesting the robustness of the algorithm to the changes in this parameter as well.

\subsubsection{Obstacle Check Distance $\tilde{o}$}\label{Section:ObstacleCheckDistance}

Next, we evaluate our algorithm's performance when the obstacle check distance $\tilde{o}$ is changed.
We set $\tilde{o}$ to values between $\SI{0.5}{m}$ and $\SI{3}{m}$ with $\SI{0.5}{m}$ increments.
Since the replanning period $\delta t = \SI{0.1}{s}$, and maximum velocity $\gamma^1 = 3.67\frac{m}{s}$, the maximum amount of distance that can be traversed by a robot until the next planning iteration is $\SI{0.367}{m}$.
The obstacle check distance must be more than this value for safety.
We set step size $\sigma=\SI{0.77}{m}$, robot check distance $\tilde{r}=\SI{2.0}{m}$, and degree of B\'ezier curves $h=12$, which are determined by premiliminary experiments and the results of the experiments in Sections~\ref{Section:StepSize},~\ref{Section:BezierCurvesDegrees}, and~\ref{Section:RobotCheckDistance}.
The results are summarized in Fig.~\ref{Figure:AverageCompTime}c and Fig.~\ref{Figure:AverageNavDuration}c.

The obstacle check distance is the most important parameter that determines the speed of trajectory optimization, and hence the planning pipeline. 
As $\tilde{o}$ increases, the number of SVM computations and the number of constraints in the optimization problem increases, which results in increased computation time.
Average navigation durations of the robots are close to $\SI{22.5}{m}$ in all cases, suggesting the robustness of the algorithm to this parameter.
We explain the reason of this robustness as follows. 
All obstacles are considered during discrete search and $\tilde{o}$ determines the obstacles that are considered during trajectory optimization.
Therefore, the path suggested by the discrete search is already very good, and obstacle avoidance behavior of trajectory optimization is only important when the discrete path is close to obstacles.
In those cases, all obstacle check distances capture the obstacles in the vicinity of the path.
Therefore, the quality of the planned trajectories does not increase as $\tilde{o}$ increases.

\subsubsection{Robot Check Distance $\tilde{r}$}\label{Section:RobotCheckDistance}

Last, we evaluate our algorithm's performance when the robot check distance $\tilde{r}$ is changed.
We set $\tilde{r}$ to values between $\SI{1}{m}$ and $\SI{3.5}{m}$ with $\SI{0.5}{m}$ increments.
Robot check distance must be at least twice the amount of distance that can be traversed by the robot in one planning iteration, i.e. $\SI{0.734}{m}$, because two robots may be travelling towards each other with their maximum speed in the worst case.
We set step size $\sigma=\SI{0.77}{m}$, obstacle check distance $\tilde{o}=\SI{1.0}{m}$, and degree of B\'ezier curves $h=12$, which are determined by premiliminary experiments and the results of the experiments in Sections~\ref{Section:StepSize},~\ref{Section:BezierCurvesDegrees}, and~\ref{Section:ObstacleCheckDistance}.

The speed of the algorithm is not affected by the robot check distance considerably, because there are $32$ robots in the environment, and from the perspective of the ego robot, there are at most $31$ constraints generated from other robots. 
Since there are more than $2000$ obstacles in the environment, effects of the obstacle check distance are more drastic than robot check distance.
Similar to other cases, average navigation duration of the robots is not affected by the choice of $\tilde{r}$ because constraints generated by the robots far away do not actually constrain the trajectory of the ego robot since the ego robot cannot move faster than its maximum speed and hence the first piece of the trajectory is never affected by those constraints.

Overall, RLSS does not result in collisions or deadlocks when parameters are not set to extreme values.
In addition, changes in parameters, outside of extreme ranges, do not result in significant changes in the average navigation durations.
These suggest that RLSS does not need extensive parameter tuning.

\subsection{Ablation Study}\label{Section:AblationStudy}
We investigate the effects of i) the prepend operation of the discrete planning stage (Line~\ref{Line:PrepentFirstPoint}, Algorithm ~\ref{Algorithm:DiscretePlanning}) and ii) the preferred distance cost term $\mJ_{pref}$ of the trajectory optimization stage to the performance of the algorithm.
The prepend operation enables the kinematic feasibility of the generated optimization problem as shown in Remark~\ref{Remark:KinematicFeasibility}.
The preferred distance cost term increases the numerical stability of the algorithm by encouraging robots to create a gap between themselves and other objects.

We consider four versions our algorithm: RLSS, RLSS without the prepend operation (RLSS w/o prepend), RLSS without the preferred distance cost term (RLSS w/o pref. dist.) and RLSS with neither.
We set step size $\sigma=\SI{0.77}{m}$, obstacle check distance $\tilde{o}=\SI{1.0}{m}$, robot check distance $\tilde{r}=\SI{2.0}{m}$, degree of B\'ezier curves $h=12$, and $c=1$ (continuity up to velocity).
Robots navigate in 3D maze like environments.
The desired trajectories are set to straight line segments connecting robot start positions to goal positions and the durations of the segments are set to the length of the line segments divided by the maximum speed of the robots.
During simulation, robots continue using their existing plans when planning fails.
Also, colliding robots continue navigating and are not removed from the experiment.

\begin{table}[tb]
    \centering
    \caption{The results of the ablation study. We compare four different versions of RLSS. We ablate i) the prepend operation of the discrete planning stage and ii) the preferred distance cost term of the trajectory optimization stage. The details of the metrics are given in Section~\ref{Section:AblationStudy}. The reported values are means and standard deviations (given in parentheses) of $10$ experiments in random maze-like environments. Both prepend operation and preferred distance cost term are important for the effectiveness of the algorithm.}
    \resizebox{\linewidth}{!}{\begin{tabular}{|c|c|c|c|}
         \cline{2-4} \multicolumn{1}{c|}{} & Fail Rate &\# Coll. & Avg. Nav. Dur [s]\\
         \hline RLSS & \textbf{0.89 (1.45) / 7904 (297)} & \textbf{0 (0)} &  \textbf{24.68 (0.93)}\\
         \hline RLSS w/o pref. dist.& 4.78 (9.81) / 8194 (489) & 0.67 (1) & 25.58 (1.53)\\
         \hline RLSS w/o prepend & 1556 (120) / 12220 (538) & 6.78 (3.73) &  38.17 (1.68)\\
         \hline RLSS with neither & 1569 (101) / 14778 (412) & 8.67 (2.87) & 46.16 (1.29) \\
         \hline 
    \end{tabular}}
    \label{Table:Ablation}
\end{table}

We generate ten 3D-maze like environments and list the average and standard deviation values for our metrics in Table~\ref{Table:Ablation}. 
% The results of the experiments are summarized in Table~\ref{Table:Ablation}.
We report the failure rate of all algorithms in the form of the ratio of number of failures to the number of planning iterations (\emph{Fail Rate} column in Table~\ref{Table:Ablation}), the number of robots that are involved in at least one collision during navigation (\emph{\# Coll.} column in Table~\ref{Table:Ablation}) and the average navigation duration of all robots (\emph{Avg. Nav. Dur.} column in Table~\ref{Table:Ablation}).

The failure rate of RLSS is $0.01\%$ on average.
RLSS w/o pref. dist. fails $0.06\%$ of the time.
RLSS w/o prepend fails $12.73\%$ of the time.
This is drastically more than RLSS w/o pref. dist because our prepend operation ensures the kinematic feasibility of the optimization problem while our preferred distance cost term tackles numerical instabilities only.
RLSS with neither results in a failure rate of $10.62\%$.
Interestingly, RLSS with the preferred distance cost term but without the prepend operation (RLSS w/o prepend) results in a higher failure rate than RLSS with neither.
We do not investigate the root cause of this since failure rates in both cases are a lot higher than of RLSS.

The effects of failures are seen in the next two metrics.
RLSS results in no collisions.
The number of colliding robots increase to $0.67$ in RLSS w/o pref. dist, $6.78$ in RLSS w/o prepend and $8.67$ in RLSS with neither on average.
The average navigation duration of robots is lowest in RLSS.
It increases by $3.65\%$ in RLSS w/o pref. dis, $54.66\%$ in RLSS w/o prepend and $87.03\%$ in RLSS with neither compared to RLSS.

These results show that the prepend operation is more important than the preferred distance cost term for the success of the algorithm.
Nevertheless, RLSS needs both to be safe and effective.

\subsection{Comparisons with Baseline Planners}\label{Section:BaselineComparison}

We compare the performance of our planner to two baseline planners that do not require communication in 3D experiments.
We set step size $\sigma=\SI{0.77}{m}$, obstacle check distance $\tilde{o}=\SI{1.0}{m}$, robot check distance $\tilde{r}=\SI{2.0}{m}$, and degree of B\'ezier curves $h=12$ in RLSS in all cases.

\subsubsection{Extended Buffered Voronoi Cell (eBVC) Planner}

The first baseling planner is a MPC-style planner based on buffered Voronoi cells introduced by~\cite{zhou2017bvc}, which we call BVC.
In the BVC approach, each robot computes its Voronoi cell within the Voronoi tesselation of the environment.
This is done by using the position information of other robots.
Robots buffer their Voronoi cells to account for their collision shapes and plan their trajectories within their corresponding buffered Voronoi cells.
Similar to RLSS, BVC does not require any communication between robots, requires perfect sensing of robot positions in the environment, and the resulting trajectories are safe only when planning is synchronized between robots.
They also require that each robot stays within its buffered Voronoi cell until the next planning iterations.
Unlike RLSS, BVC is based on buffered Voronoi cells and it cannot work with arbitrary convex objects.
Instead, robots are modeled as hyperspheres in BVC.

The original formulation presented in the BVC article only allows position state for the robots, which cannot model a rich set of dynamics, including double, triple, or higher order integrators.
We extend their formulation to all discrete linear time invariant systems with position output.
We define the systems with three matrices $\vA, \vB, \vC$.
Since the output of the system is the position of the robot, it cannot depend on the current input, hence $\vD = \vzero$.
This allows us to formulate the problem for a richer set of dynamics and have constraints on higher order derivatives than robot velocity.

BVC as published also does not consider obstacles in the environment.
We extend their formulation to static obstacles by modeling obstacles as robots.
Since obstacles are static, they stay within their buffered Voronoi cells at all times.
Since we model obstacles as robots, extended BVC does not require the ability of distinguishing robots from obstacles.

Our extended BVC formulation, which we call eBVC, is as follows: \\
\resizebox{\linewidth}{!}{
\begin{minipage}{\linewidth}
% \begin{align*}
%     \setcounter{equation}{0}
%     &\min_{\vu_0, \ldots, \vu_{M-1}} \sum_{i=1}^M \lambda_i \normtwo{\vp_i- \vd(\tilde{T} + M\Delta t)}^2 + \sum_{i=0}^{M-1}\theta_i \normtwo{\vu_i}^2 s.t.\\
%     &\ \ \vx_{i+1} = \vA \vx_{i} + \vB \vu_{i}\mkern3mu \forall i\in\{0,\ldots,M-1\}\\
%     &\ \ \vp_{i} = \vC \vx_{i} \mkern76mu\forall i \in \{0,\ldots,M\}\\
%     &\ \ \vp_{i} \in \mV \mkern98mu\forall i \in \{0,\ldots,M\}\\
%     &\ \ \vp_i \in \mW \mkern92mu\forall i \in \{0,\ldots,M\}\\
%     &\ \ \vu_{min} \preceq \vu_i \preceq \vu_{max}\ \forall i \in \{0, \ldots, M-1\}\\
%     &\ \ \vx_{min} \preceq \vx_i \preceq \vx_{max}\mkern9mu \forall i \in \{0, \ldots, M\}
% \end{align*}
\begin{align*}
    \setcounter{equation}{0}
    &\min_{\vu_0, \ldots, \vu_{M-1}} \sum_{i=1}^M \lambda_i \normtwo{\vp_i- \vd(\tilde{T} + M\Delta t)}^2+ \sum_{i=0}^{M-1}\theta_i \normtwo{\vu_i}^2 s.t.\\
    &\ \ \vx_{i+1} = \vA \vx_{i} + \vB \vu_{i} \tag*{$\forall i\in\{0,\ldots,M-1\}$}\\
    &\ \ \vp_{i} = \vC \vx_{i} \tag*{$\forall i \in \{0,\ldots,M\}$}\\
    &\ \ \vp_{i} \in \mV \tag*{$\forall i \in \{0,\ldots,M\}$}\\
    &\ \ \vp_i \in \mW \tag*{$\forall i \in \{0,\ldots,M\}$}\\
    &\ \ \vu_{min} \preceq \vu_i \preceq \vu_{max}  \tag*{$\forall i \in \{0, \ldots, M-1\}$}\\
    &\ \ \vx_{min} \preceq \vx_i \preceq \vx_{max}\tag*{$\forall i \in \{0, \ldots, M\}$}
\end{align*}
\end{minipage}
}
where $\tilde{T}$ is the current timestamp, $\vd(t)$ is the desired trajectory for the robot, $\Delta t$ is the discretization timestep of the system, $M$ is the number of steps to plan for, $\vu_i$ is the input to apply from timestep $i$ to $i+1$, $\vx_i$ is the state at timestep $i$, $\vp_i$ is the position at timestep $i$, $\mV$ is the buffered Voronoi cell of the robot, $\vu_{min}$ and $\vu_{max}$ are the limits for the inputs, $\vx_{min}$ and $\vx_{max}$ are the limits for the states (which can be used to bound velocity in a double integrator system, or velocity and acceleration in a triple integrator system for example), and $\mW$ is the workspace of the robot.
$M\Delta t$ is the planning horizon.
A robot plans toward the position $\vd(\tilde{T} + M\Delta t)$, which is the position of the robot after the planning horizon if it could follow the desired trajectory perfectly.
$\vx_0$ is the current state of the robot.
The first term of the cost function penalizes deviation from the goal position for the final and each intermediate position with different weights $\lambda_i$.
The second term of the cost function is the input cost that penalizes input magnitudes with different weights $\theta_i$.
We apply the first input of the solution for duration $\Delta t$, and replan at the next timestep.

We use our own implementation of eBVC as explained above during the comparisons.

\subsubsection{Relative Safe Flight Corridor (RSFC) Planner}

The second planner we compare against is presented by~\cite{park2021rsfc}, in which piecewise B\'ezier curves are computed, executed for a short duration, and replanning is done at the next iteration similar to our work.
It utilizes the fact that the difference of two B\'ezier curves is another B\'ezier curve by constraining these relative B\'ezier curves to be inside safe regions (relative safe flight corridors, or RSFCs) defined according to robot collision shapes.
We call this algorithm RSFC for short.
RSFC does not require any communication between robots.
It utilizes both positions and velocities of other objects in the environment, hence requires more sensing information than our algorithm.
Velocities are used to predict the trajectories of other robots as piecewise B\'ezier curves.
While it can handle dynamic obstacles as well, we use it in static environments in our comparisons, since RLSS does not handle dynamic obstacles explicitly.

We use the authors' implementation of RSFC during our comparisons.

\subsubsection{Experiments \& Results}

We compare RLSS against eBVC and RSFC in $10$ different experiments differing in required degree of continuity, desired trajectories, and map of the environment.
In all experiments, $32$ robots are placed in a circle formation with radius $\SI{20}{m}$ in 3D.
The task is to swap the positions of robots to the antipodal points on the circle.
The results of the experiments are summarized in Table~\ref{Table:Comparisons}.

\begin{table*}[tb]
\centering
\caption{The results of the comparisons of RLSS, eBVC, and RSFC are summarized. Each experiment differ in the map used during navigation, prior map used during desired trajectory computation, and the required degree of continuity. RLSS results in longer navigation durations than eBVC and RSFC on average; but eBVC suffers from deadlocks in environments with obstacles and RSFC results in collisions in environments with obstacles.}
\resizebox{\textwidth}{!}{\begin{tabular}{|c|c|c|c|c|c|c|c|c|c|}
    \hline Experiment & Map & Prior Map & Continuity & Algorithm & \# Deadlocks & \# Coll. Robots & Avg. Collision Dur. [s] & Avg. Nav. Dur. [s] & Avg. Comp. Time [ms] \\
    \hline \multirow{2}{*}{1} & \multirow{5}{*}{empty} & \multirow{5}{*}{empty} & \multirow{2}{*}{velocity} &  \multicolumn{1}{c}{RLSS} & \multicolumn{1}{c}{0} &\multicolumn{1}{c}{0}& \multicolumn{1}{c}{NA} & \multicolumn{1}{c}{22.37} & \multicolumn{1}{c|}{107}\\
    & & & & \multicolumn{1}{c}{eBVC} & \multicolumn{1}{c}{0}&\multicolumn{1}{c}{0} & \multicolumn{1}{c}{NA} & \multicolumn{1}{c}{\textbf{18.50}} & \multicolumn{1}{c|}{\textbf{106}}\\
    \cline{1-1}\cline{4-10}\multirow{3}{*}{2}& & &\multirow{3}{*}{acceleration} & \multicolumn{1}{c}{RLSS} & \multicolumn{1}{c}{0}&\multicolumn{1}{c}{0} & \multicolumn{1}{c}{NA} & \multicolumn{1}{c}{22.12} & \multicolumn{1}{c|}{110}\\
    & & & & \multicolumn{1}{c}{eBVC} & \multicolumn{1}{c}{0} &\multicolumn{1}{c}{0}& \multicolumn{1}{c}{NA} & \multicolumn{1}{c}{20.65} & \multicolumn{1}{c|}{161}\\
    & & & & \multicolumn{1}{c}{RSFC} & \multicolumn{1}{c}{0} & \multicolumn{1}{c}{0}&\multicolumn{1}{c}{NA} & \multicolumn{1}{c}{\textbf{19.83}} & \multicolumn{1}{c|}{\textbf{47}}\\
    
    \hline \hline \multirow{2}{*}{3} & \multirow{4}{*}{forest} & \multirow{4}{*}{forest} & \multirow{2}{*}{velocity} &  \multicolumn{1}{c}{RLSS} & \multicolumn{1}{c}{\textbf{0}} &\multicolumn{1}{c}{\textbf{0}}& \multicolumn{1}{c}{\textbf{NA}} & \multicolumn{1}{c}{23.11} & \multicolumn{1}{c|}{\textbf{182}}\\
    & & & & \multicolumn{1}{c}{eBVC} & \multicolumn{1}{c}{10} & \multicolumn{1}{c}{7}&\multicolumn{1}{c}{0.53} & \multicolumn{1}{c}{\textbf{18.94}} & \multicolumn{1}{c|}{387}\\
    \cline{1-1}\cline{4-10}\multirow{2}{*}{4}& & &\multirow{2}{*}{acceleration} & \multicolumn{1}{c}{RLSS} & \multicolumn{1}{c}{\textbf{0}} & \multicolumn{1}{c}{\textbf{0}}&\multicolumn{1}{c}{\textbf{NA}} & \multicolumn{1}{c}{23.06} & \multicolumn{1}{c|}{\textbf{147}}\\
    & & & & \multicolumn{1}{c}{eBVC} & \multicolumn{1}{c}{8} &\multicolumn{1}{c}{13}& \multicolumn{1}{c}{0.83} & \multicolumn{1}{c}{\textbf{21.24}} & \multicolumn{1}{c|}{763}\\
    
    \hline \hline \multirow{2}{*}{5} & \multirow{5}{*}{forest} & \multirow{5}{*}{empty} & \multirow{2}{*}{velocity} &  \multicolumn{1}{c}{RLSS} & \multicolumn{1}{c}{\textbf{0}}&\multicolumn{1}{c}{\textbf{0}} & \multicolumn{1}{c}{\textbf{NA}} & \multicolumn{1}{c}{22.62} & \multicolumn{1}{c|}{\textbf{192}}\\
    & & & & \multicolumn{1}{c}{eBVC} & \multicolumn{1}{c}{8} &\multicolumn{1}{c}{24}& \multicolumn{1}{c}{1.35} & \multicolumn{1}{c}{\textbf{18.54}} & \multicolumn{1}{c|}{186}\\
    \cline{1-1}\cline{4-10}\cline{4-10}\multirow{3}{*}{6}& & &\multirow{3}{*}{acceleration} & \multicolumn{1}{c}{RLSS} & \multicolumn{1}{c}{\textbf{0}} &\multicolumn{1}{c}{\textbf{0}}& \multicolumn{1}{c}{\textbf{NA}} & \multicolumn{1}{c}{22.72} & \multicolumn{1}{c|}{244}\\
    & & & & \multicolumn{1}{c}{eBVC} & \multicolumn{1}{c}{11} &\multicolumn{1}{c}{15}& \multicolumn{1}{c}{0.71} & \multicolumn{1}{c}{\textbf{21.28}} & \multicolumn{1}{c|}{970}\\
    & & & & \multicolumn{1}{c}{RSFC}& \multicolumn{1}{c}{\textbf{0}} & \multicolumn{1}{c}{6} & \multicolumn{1}{c}{0.42} & \multicolumn{1}{c}{21.82} & \multicolumn{1}{c|}{\textbf{201}}\\
    
    \hline \hline \multirow{2}{*}{7} & \multirow{4}{*}{maze} & \multirow{4}{*}{maze} & \multirow{2}{*}{velocity} &  \multicolumn{1}{c}{RLSS} & \multicolumn{1}{c}{\textbf{0}} & \multicolumn{1}{c}{\textbf{0}} & \multicolumn{1}{c}{\textbf{NA}} & \multicolumn{1}{c}{25.09} & \multicolumn{1}{c|}{160}\\
    & & & & \multicolumn{1}{c}{eBVC} & \multicolumn{1}{c}{21} & \multicolumn{1}{c}{9} & \multicolumn{1}{c}{0.67} & \multicolumn{1}{c}{\textbf{20.12}} & \multicolumn{1}{c|}{\textbf{145}}\\
    \cline{1-1}\cline{4-10}\cline{4-10}\multirow{2}{*}{8}& & &\multirow{2}{*}{acceleration} & \multicolumn{1}{c}{RLSS} & \multicolumn{1}{c}{\textbf{0}} & \multicolumn{1}{c}{\textbf{0}} & \multicolumn{1}{c}{\textbf{NA}} & \multicolumn{1}{c}{25.32} & \multicolumn{1}{c|}{\textbf{170}}\\
    & & & & \multicolumn{1}{c}{eBVC} & \multicolumn{1}{c}{17} & \multicolumn{1}{c}{13} & \multicolumn{1}{c}{1.49} & \multicolumn{1}{c}{\textbf{23.80}} & \multicolumn{1}{c|}{264}\\
    
    \hline \hline \multirow{2}{*}{9} & \multirow{5}{*}{maze} & \multirow{5}{*}{empty} & \multirow{2}{*}{velocity} & \multicolumn{1}{c}{RLSS} & \multicolumn{1}{c}{\textbf{0}} & \multicolumn{1}{c}{\textbf{0}} & \multicolumn{1}{c}{\textbf{NA}} & \multicolumn{1}{c}{\textbf{27.98}} & \multicolumn{1}{c|}{410}\\
    & & & & \multicolumn{1}{c}{eBVC} & \multicolumn{1}{c}{32} & \multicolumn{1}{c}{0} & \multicolumn{1}{c}{NA} & \multicolumn{1}{c}{NA} & \multicolumn{1}{c|}{\textbf{176}}\\
    \cline{1-1}\cline{4-10}\cline{4-10}\multirow{3}{*}{10}& & &\multirow{3}{*}{acceleration} & \multicolumn{1}{c}{RLSS} & \multicolumn{1}{c}{\textbf{0}} & \multicolumn{1}{c}{\textbf{0}} & \multicolumn{1}{c}{\textbf{NA}} & \multicolumn{1}{c}{32.04} & \multicolumn{1}{c|}{386}\\
    & & & & \multicolumn{1}{c}{eBVC} & \multicolumn{1}{c}{30} & \multicolumn{1}{c}{2} & \multicolumn{1}{c}{3.26} & \multicolumn{1}{c}{\textbf{21.75}} & \multicolumn{1}{c|}{407}\\
    & & & & \multicolumn{1}{c}{RSFC}& \multicolumn{1}{c}{\textbf{0}} & \multicolumn{1}{c}{22} & \multicolumn{1}{c}{0.34} & \multicolumn{1}{c}{28.23} & \multicolumn{1}{c|}{\textbf{80}}\\
    \hline
\end{tabular}}
\label{Table:Comparisons}
\end{table*}

\begin{figure}
     \centering
     \subfloat[Experiment $4$ Desired Trajectories (Side View)\label{Figure:Exp4Desired}]{
     \centering
       \includegraphics[width=0.45\linewidth]{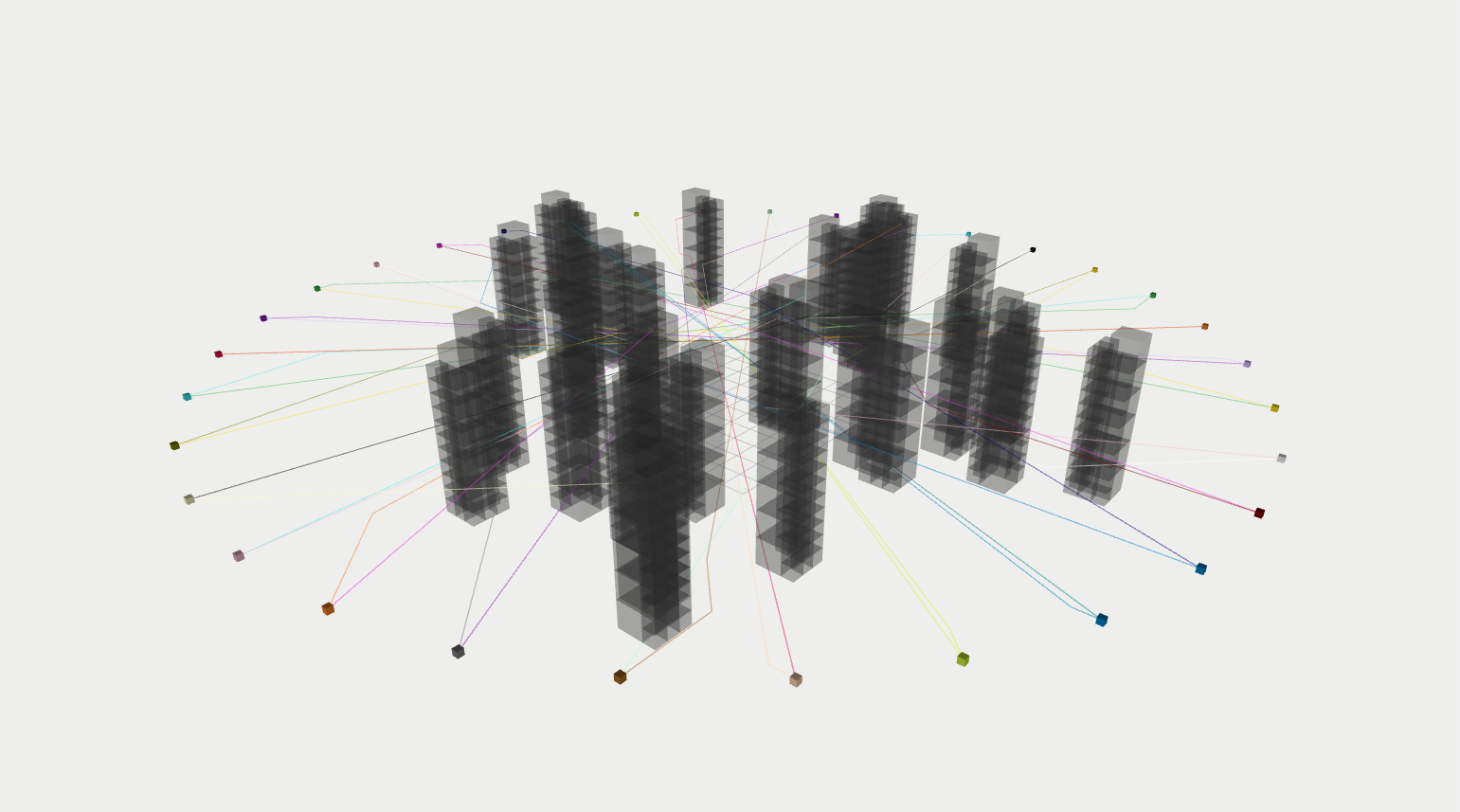}
     }
     \hfill
     \subfloat[Experiment $7$ Desired Trajectories (Side View)\label{Figure:Exp7Desired}]{
     \centering
       \includegraphics[width=0.45\linewidth]{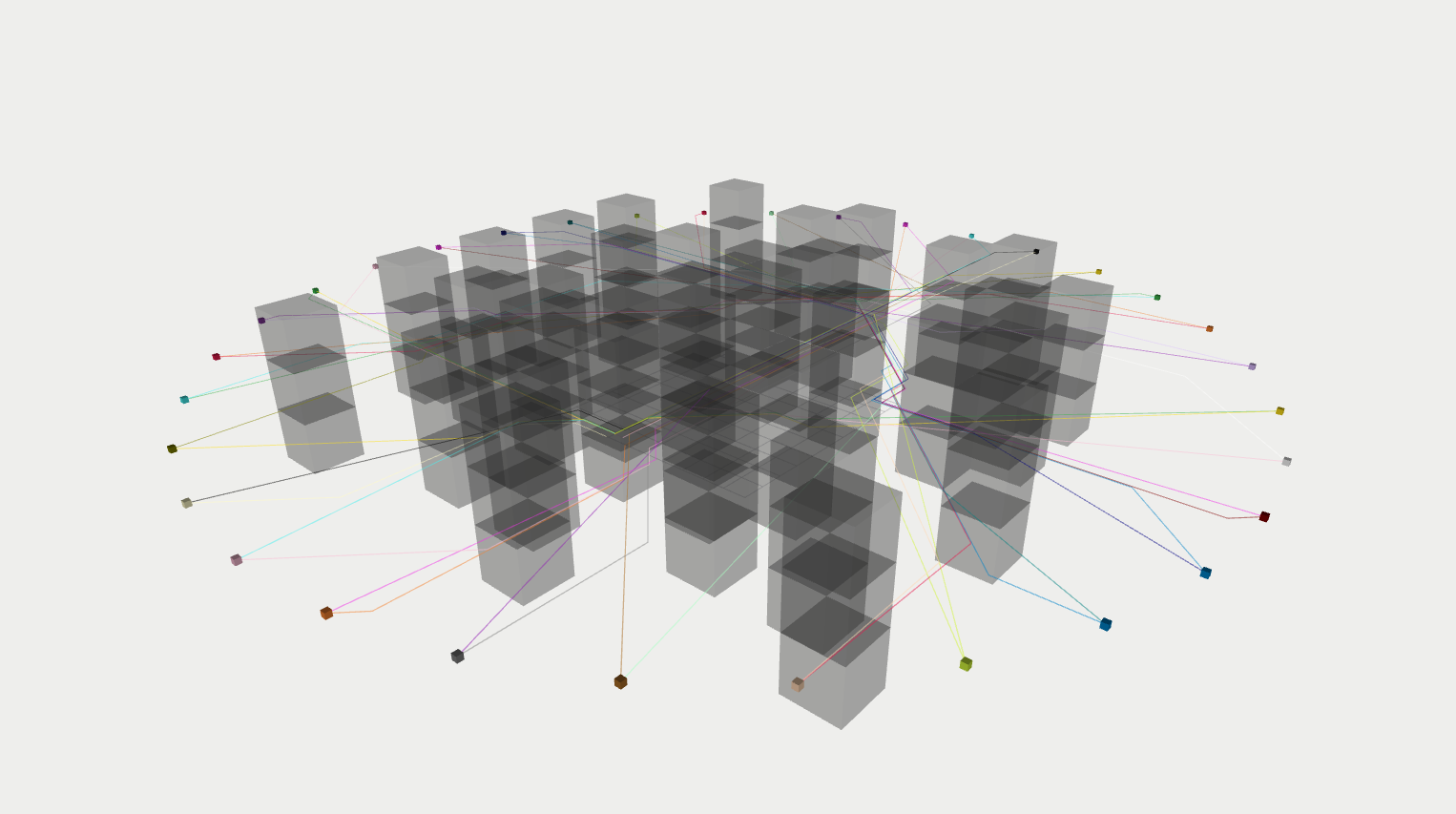}
     }
     \hfill
     \subfloat[Experiment $4$ Executed Trajectories (RLSS / Top View)\label{Figure:Exp4Executed}]{
     \centering
       \includegraphics[width=0.45\linewidth]{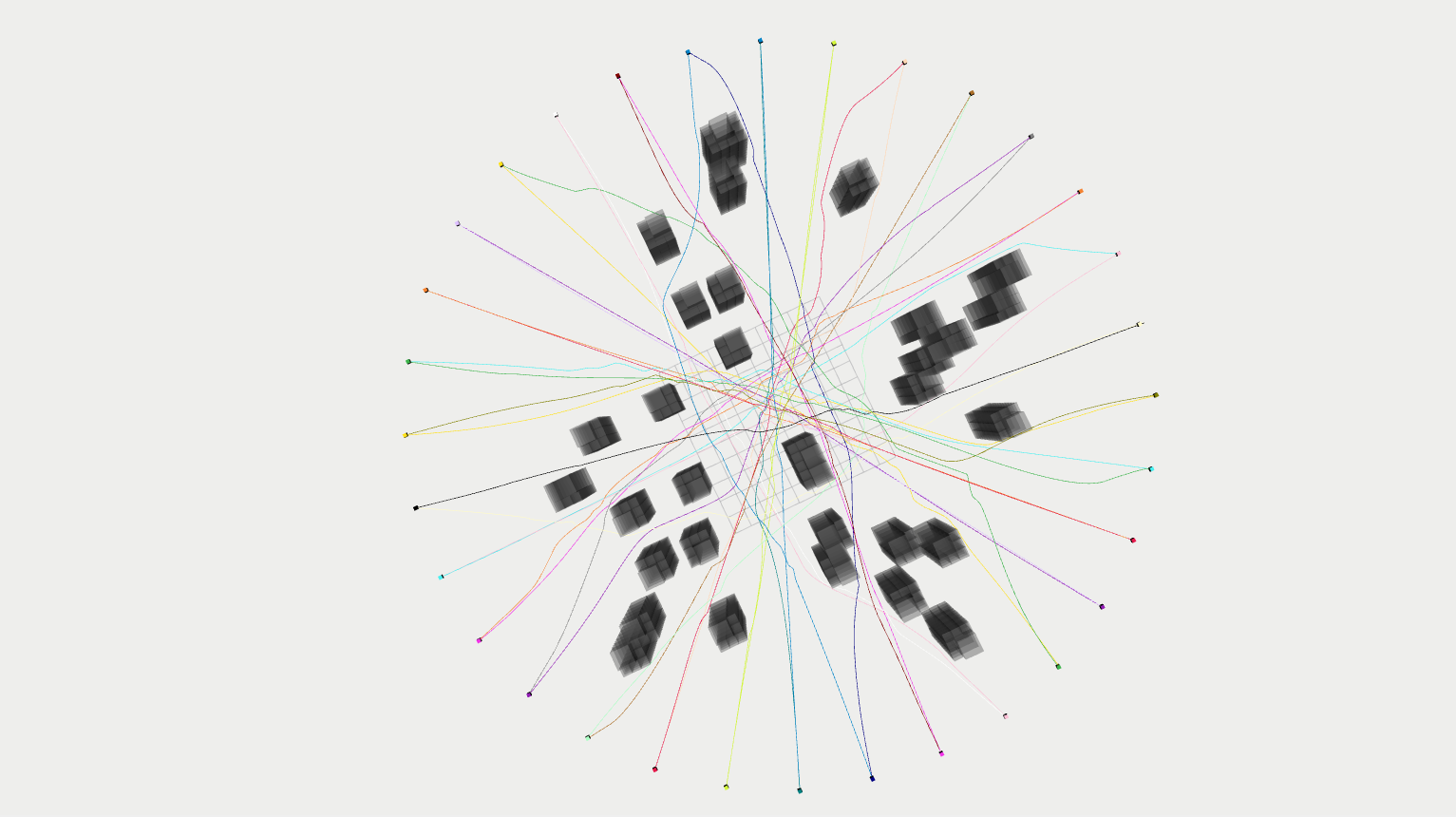}
     }
     \hfill
     \subfloat[Experiment $7$ Executed Trajectories (RLSS / Top View)\label{Figure:Exp7Executed}]{
     \centering
       \includegraphics[width=0.45\linewidth]{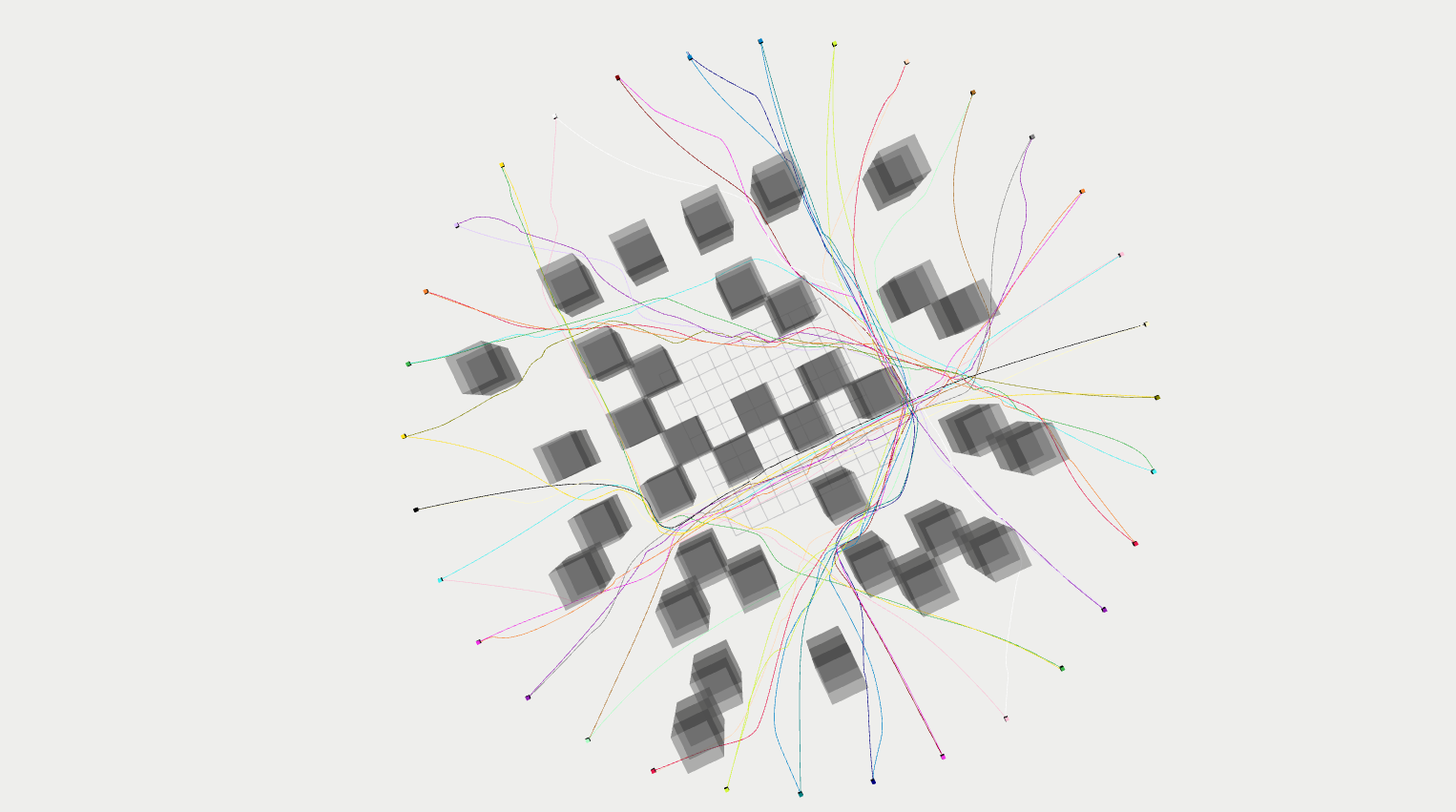}
     }
     \caption{Desired trajectories and the used forest map of experiment $4$ is given in (a). Same forest map is used in each forest experiment. Desired trajectories and the used maze map of experiment $7$ is given in (b). Same maze map is used in each maze experiment. (c) shows the executed trajectories of robots running RLSS from top in experiment $4$. (d) shows the executed trajectories of robots running RLSS from top in experiment $7$.}
     \label{Figure:Exp4And7}
\end{figure}

There are $3$ maps we use: empty, forest (Fig.~\ref{Figure:Exp4Desired}), and maze (Fig.~\ref{Figure:Exp7Desired}), listed in the map column of Table~\ref{Table:Comparisons}.
In the empty map, there are no obstacles in the environment.
The forest map is a random forest with $\SI{10}{\%}$ occupancy; it has a radius of $\SI{15}{m}$ and each tree is a cylinder with radius $\SI{0.5}{m}$.
The maze map is a maze-like environment with choke regions.

We compute the desired trajectories of the robots by running a single-agent shortest path using the discrete planning stage of RLSS.
We run single-agent shortest path on a prior map, which is set to either a full map of the environment or to an empty map; this is listed in the prior map column of Table~\ref{Table:Comparisons}.
When the prior map is empty, single-agent shortest paths are straight line segments connecting robot start positions to robot goal positions.

We set two different continuity requirements: velocity and acceleration, listed in the continuity column of Table~\ref{Table:Comparisons}.
When velocity continuity is required, the system of eBVC is a double integrator with position output.
When acceleration continuity is required, the system of eBVC is a triple integrator with position output.

We compare RSFC to RLSS and eBVC only with acceleration continuity requirement because the authors' implementation hard codes the acceleration continuity requirement, while in theory it can work with any degree.
Also, we compare RSFC only in the case of an empty prior map in order not to change RSFC's source code, while in theory it can be guided with arbitrary trajectories.

We plan for $\SI{5}{s}$ long trajectories in every $\SI{0.1}{s}$ with all algorithms.
In eBVC, we plan for $M=50$ steps with discretization timestep $\Delta t= \SI{0.1}{s}$.
We set state and input upper and lower bounds in eBVC in order to obey the dynamic limits of the robots.
We set distance to goal weights $\lambda_1=120$, $\lambda_i = 20\ \forall i\geq2$ in eBVC, putting more importance on the position of the robot after $1$ timestep.
We set $\theta_i=1\ \forall i$ in eBVC.

Both eBVC and RSFC require spherical obstacles.
We use the smallest spheres containing each leaf-level box of the OcTree structure as obstacles in eBVC and RSFC.
Similar to RLSS, we use robot and obstacle check distance to limit the number of obstacles considered at each iteration.
We set both obstacle and robot check distance $\tilde{o} = \tilde{r} = \SI{2.0}{m}$ in eBVC, and set $\tilde{o} = \tilde{r} = \SI{5.0}{m}$ in RSFC, since smaller values in RSFC result in a high number of collisions and higher values for the parameters do not improve the success of the algorithms. %because it suffers greatly for smaller values in terms of collisions.
Robots are modeled as spheres in eBVC and RSFC as well.
We set robot shapes to spheres with radius $\SI{0.173}{m}$, which are the smallest spheres containing the actual robot shapes.
We count collisions only when contained leaf-level OcTree boxes and contained robot shapes intersect.
This gives both eBVC and RSFC buffer zones for collisions.

We report the number of deadlocking robots (\emph{\# Deadlocks} column in Table~\ref{Table:Comparisons}), number of robots that are involved in at least one collision (\emph{\# Coll. Robots} column in Table~\ref{Table:Comparisons}), collision duration of robots that are involved in collisions averaged over robots (\emph{Avg. Collision Dur.} column in Table~\ref{Table:Comparisons}), average navigation duration of non-deadlocking robots from their start positions to goal positions (\emph{Avg. Nav. Dur.} column in Table~\ref{Table:Comparisons}), and computation time per iteration averaged over each planning iteration of each robot (\emph{Avg. Comp. Time} column in Table~\ref{Table:Comparisons}).
We continue the navigation of colliding robots and do not remove them from the experiment.

Executed trajectories of robots running RLSS are shown in Fig.~\ref{Figure:Exp4Executed} for experiment $4$, and in Fig.~\ref{Figure:Exp7Executed} for experiment $7$ as examples.

RLSS does not result in any deadlocks or collisions in all cases.
eBVC has a significant number of deadlocks and RSFC results in collisions in experiments with obstacles.

When there are no obstacles in the environment, e.g., in experiments $1$ and $2$, no algorithm results in deadlocks or collisions.
When velocity continuity is required, e.g., in experiment $1$, the average navigation duration of robots running eBVC is $\SI{18}{\%}$ lower than those that run RLSS.
Both eBVC and RLSS run close to $\SI{9}{Hz}$ on average.
When acceleration continuity is required, e.g., in experiment $2$, the average navigation duration of robots running RSFC is $\SI{10}{\%}$ lower than those that run RLSS; and the average navigation duration for eBVC is $\SI{7}{\%}$ lower than that for RLSS.
RLSS runs at about $\SI{9}{Hz}$ on average, eBVC runs close to $\SI{6}{Hz}$ on average, and RSFC runs close to $\SI{21}{Hz}$ on average.
When there are no obstacles in the environment, the discrete search of RLSS results in unnecessary avoidance movements, which is the main reason for average navigation duration differences.

When there are obstacles in the environment, performance of both eBVC and RSFC  degrades in terms of the number of deadlocks and collisions.

In experiment $3$, even if the full prior map of the environment is given during initial discrete search with only velocity continuity, $10$ out of $32$ robots deadlock, and $7$ out of the remaining $22$ get involved in at least one collision when eBVC is used.
When acceleration continuity is required (experiment $4$), $8$ robots deadlock and $13$ other robots get involved in collisions, resulting in only $11$ robot reaching their goal volumes without collisions in eBVC.
RLSS both works faster than eBVC and results in no deadlocks or collisions in those cases.

When the prior map is not known in the forest environment (experiments $5$ and $6$), eBVC results in a lot of deadlocks and collisions.
All robots in experiment $5$ either deadlock or collide when eBVC is used.
In experiment $6$, RSFC does a lot better than eBVC.
RSFC results in no deadlocks while eBVC results in $11$ deadlocks.
$15$ out of remaining $21$ robots get involved in at least one collision when eBVC is used with an average collision duration of $\SI{0.71}{s}$.
$6$ out of $32$ robots running RSFC collide at least once with average collision duration of $\SI{0.42}{s}$.
RLSS does not result in any deadlocks or collisions.

When the environment is a complicated maze, the performance of both eBVC and RSFC degrades more.
With full prior map and velocity continuity (experiment $7$), $21$ out of $32$ robots deadlock, $9$ out of remaining $11$ are involved in collisions, leaving only $2$ reaching to their goal volumes without an incident when eBVC is used.
With full prior map and acceleration continuity (experiment $8$), $17$ out of $32$ robots deadlock, $13$ out of remaining $15$ are involved in collisions, again leaving only $2$ that reach their goal volumes without incident when eBVC is used.
RLSS does not result in any deadlocks or collisions in these scenarios.
When the prior map is empty, i.e. the desired trajectories are straight line segments, the performance of eBVC degrades even more. 
All robots deadlock in the case with velocity continuity (experiment $9$), while
$30$ out of $32$ robots deadlock and the rest get involved in collisions in the case with acceleration continuity (experiment $10$).
No robot running RSFC deadlocks but $22$ out of $32$ get involved in collisions at least once with $\SI{0.34}{s}$ average collision duration in experiment $10$ with acceleration continuity.
RLSS does not result in any deadlocks or collisions in those cases.

Overall, robots running RLSS have higher navigation durations than those that use eBVC or RSFC in all experiments.
While the higher navigation duration is not an important metric when other algorithms cause deadlocks or collisions, it is an important metric when they do not (experiments $1$ and $2$ with no obstacles in particular). 
eBVC does not have an integrated discrete planner, which is the reason for its good performance in terms of average navigation durations.
When robots are close to each other, RLSS uses the free space less effectively since discrete planning has a step size $\SI{0.77}{m}$.
eBVC does not rely on discrete planning and hence avoids other robots by executing small direction changes.
RSFC utilizes velocities of other robots on top of positions, which allows it to estimate the intents of robots more effectively, resulting in a better usage of the free space, and hence results in better navigation durations on average.
Since RLSS does not utilize communication and uses only positions of other robots, it cannot deduce the intents of other robots.
This results in fluctuations of plans between planning iterations, which increases the average navigation durations of robots.
Fluctuations of plans increase when the environment is dense as seen in the supplemental video\footnote{\url{https://youtu.be/Jrdvf2qyzrg}}. 
If the environment becomes overly constraining, e.g. tens of robots trying to pass through a narrow tube, these fluctuations may turn into livelocks.

However, when obstacles are introduced in the environment, the performance of RLSS is better than other algorithms.
eBVC suffers greatly from deadlocks.
RSFC does not result in deadlocks but results in collisions, even while using more information than RLSS (velocity and position instead of position only).

\textbf{A note about the statistical significance of the results:} In each experiment, we run each algorithm on single randomly generated forest-like or maze-like environment.
To show that the results are consistent for environment types, we generate $10$ forest-like environments with the same parameters for experiment $3$ and run RLSS and eBVC.
RLSS does not result in deadlocks or collisions in any of the cases.
The average navigation duration of robots averaged over $10$ environments is $\SI{19.15}{s}$ with standard deviation $\SI{0.24}{s}$ for eBVC and $\SI{23.59}{s}$ with standard deviation $\SI{0.25}{s}$ for RLSS.
The ratio between average navigation durations when eBVC or RLSS is used is consistent with the reported values given in Table~\ref{Table:Comparisons} for experiment $3$.

\subsection{Physical Robots}\label{Section:RealRobotExperiments}

\begin{figure}
    \centering
    %  \subfloat[]{%
    %   \includegraphics[width=0.31\linewidth]{crazyflienoobs/crazyflie_noobs_start00001.png}
    %  }
    %  \hfill
    % \subfloat[Heterogeneous team of differential drive robots are navigating through an environment without obstacles.\label{fig:phy-dd-1}]{%
    %   \includegraphics[width=0.48\linewidth]{diffdrive/diff_noobs_start00611.png}
    %  }
    %  ~
    \subfloat[Heterogeneous team of differential drive robots navigating through an environment without obstacles. A person changes the positions of the robots while RLSS is running.\label{fig:phy-dd-2}]{%
       \includegraphics[width=0.48\linewidth]{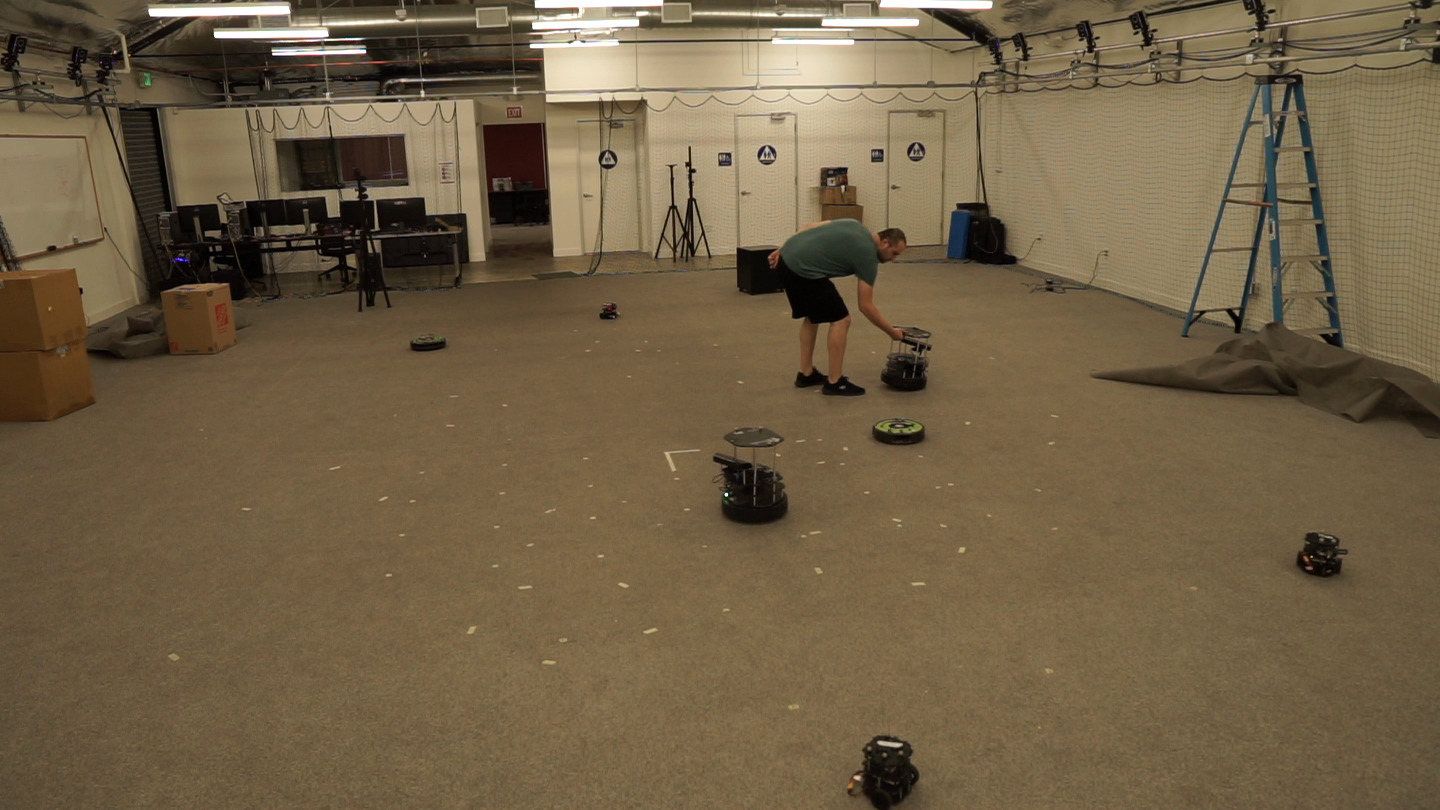}
     }
     ~
     \subfloat[6 Crazyflie 2.0s are navigating through an environment without obstacles.\label{fig:phy-cf-noobs-1}]{%
       \includegraphics[width=0.48\linewidth]{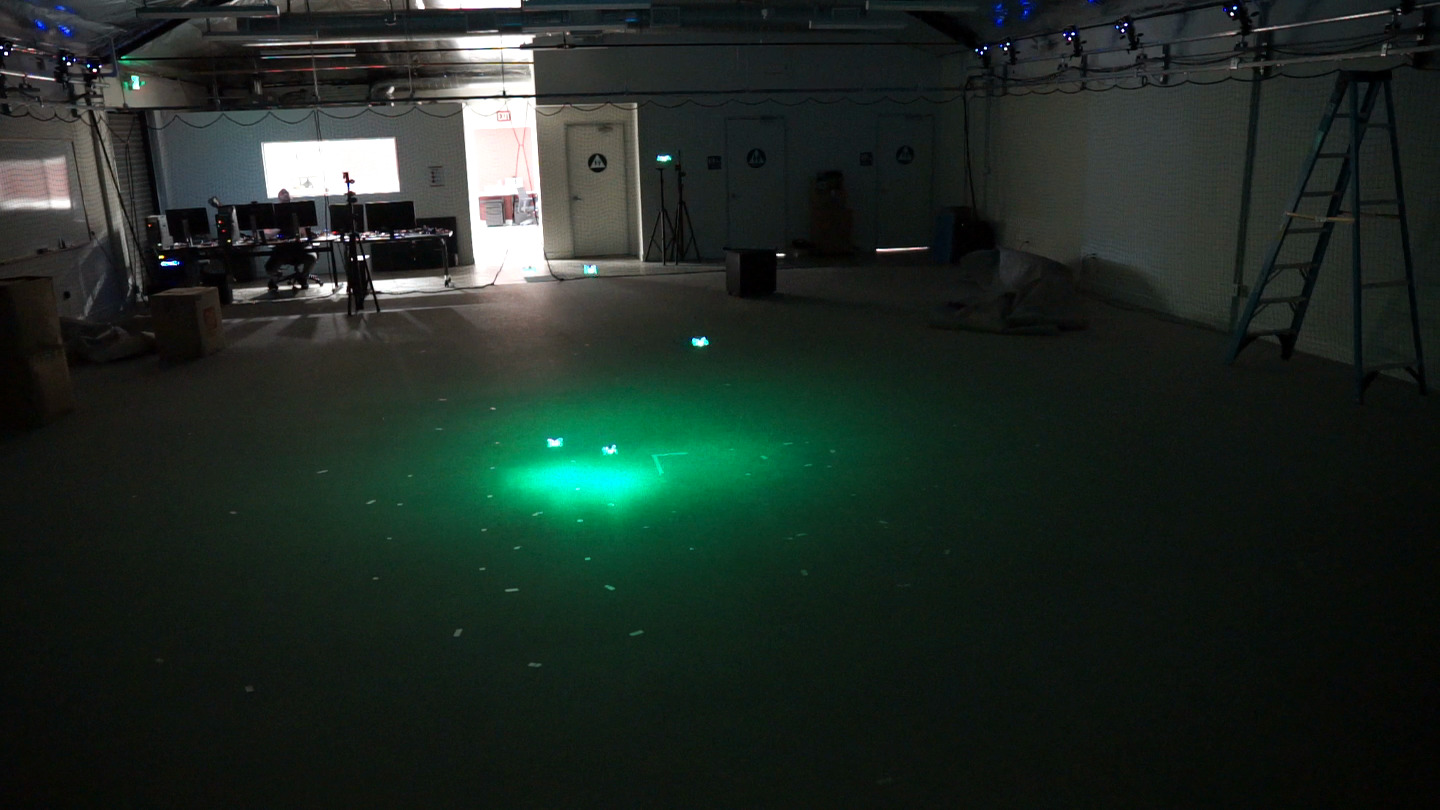}
     }
    %  ~
    % \subfloat[6 Crazyflie 2.0s in their final positions in an environment without obstacles.\label{fig:phy-cf-noobs-2}]{%
    %   \includegraphics[width=0.48\linewidth]{crazyflienoobs/crazyflie_noobs_start00570.png}
    %  }
     \hfill
     \subfloat[6 Crazyflie 2.0s are navigating through an environment with obstacles.\label{fig:phy-cf-obs-1}]{%
       \includegraphics[width=0.48\linewidth]{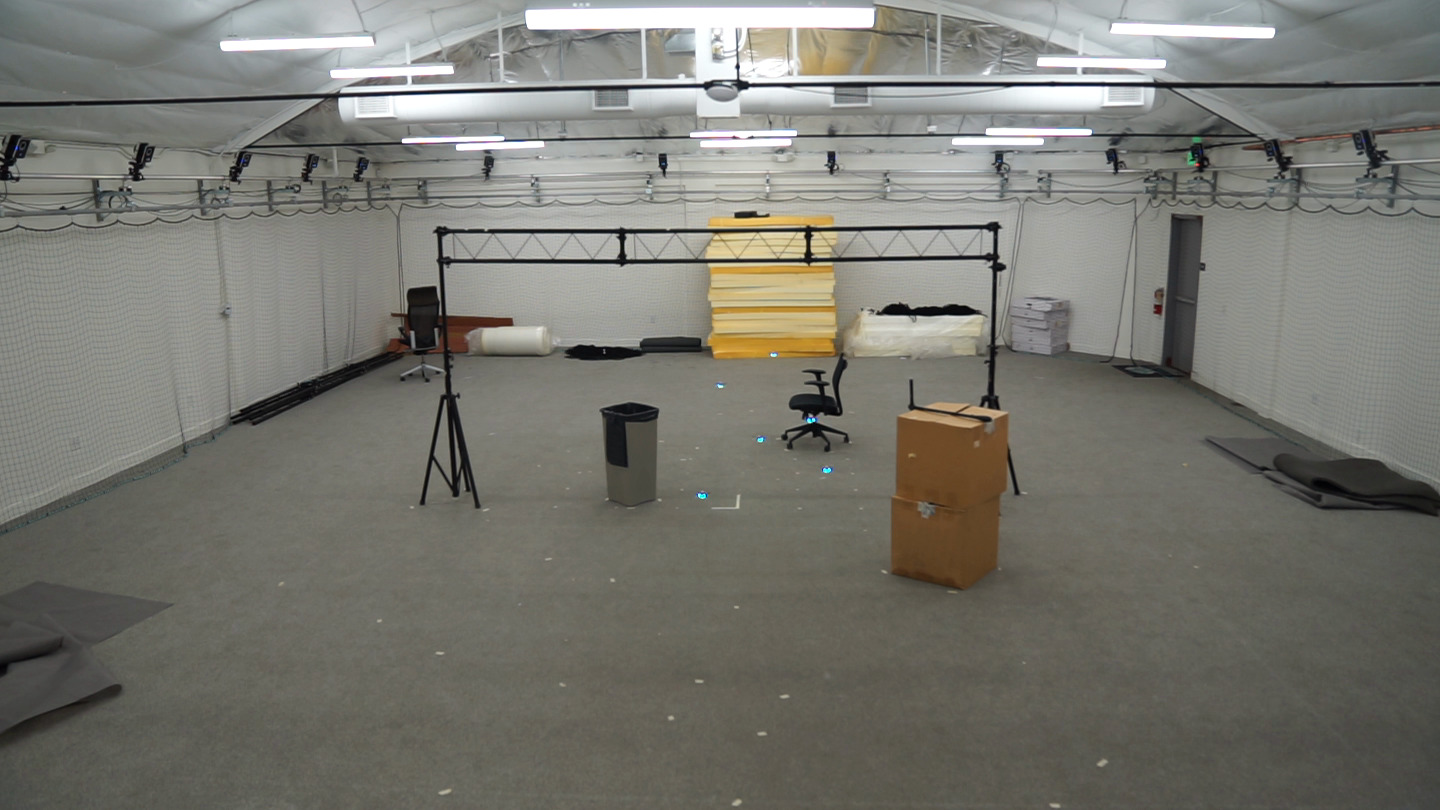}
     }
     ~
    \subfloat[6 Crazyflie 2.0s are navigating through an environment with obstacles. A person changes the positions of the robots while RLSS is running.\label{fig:phy-cf-obs-2}]{%
       \includegraphics[width=0.48\linewidth]{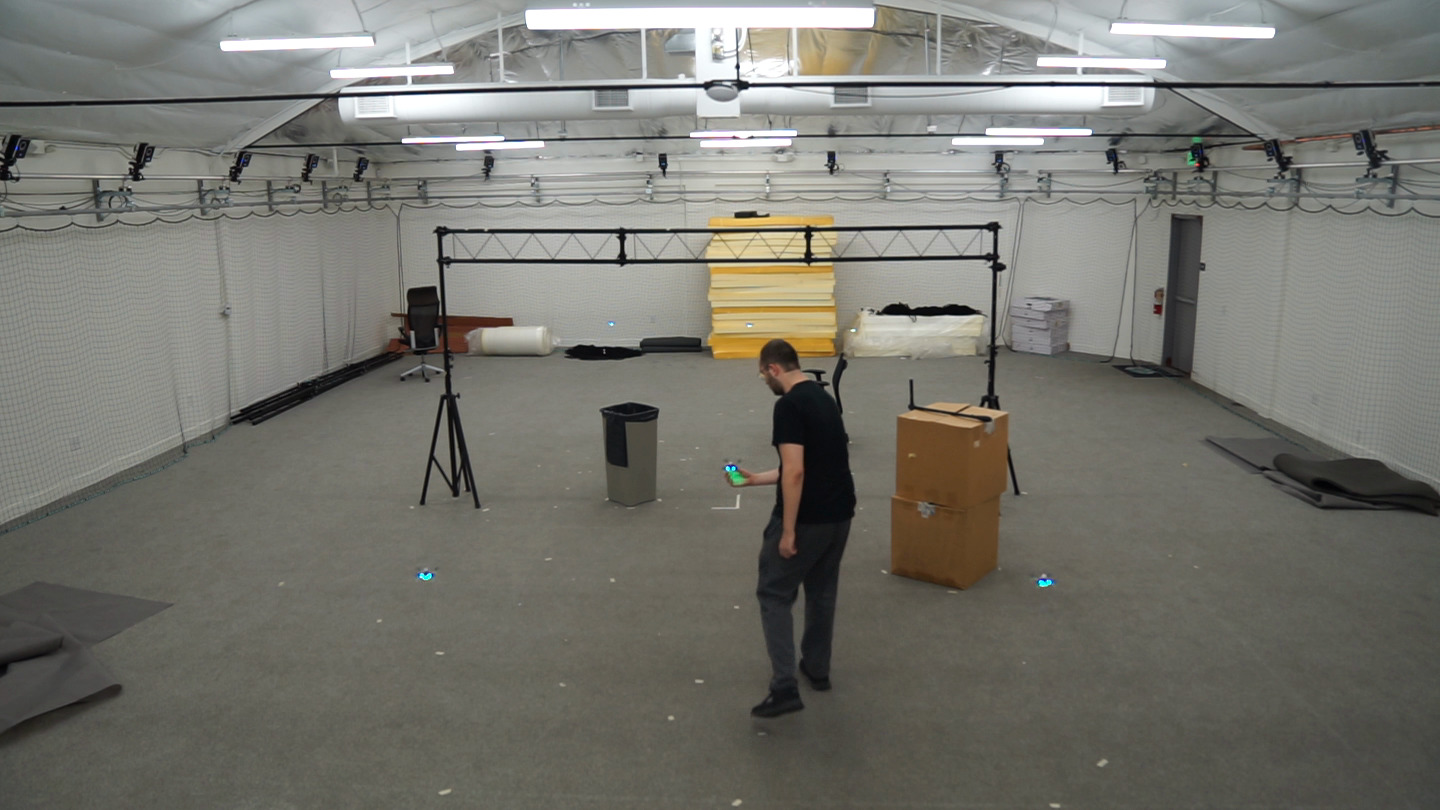}
     }
     
     \caption{Physical robot experiments using Turtlebot2s, Turtlebot3s, iRobot Create2s, and Crazyflie 2.0s. RLSS works in real-time under external disturbances.}
     \label{figure:physical-experiments}
\end{figure}

We implement and run RLSS on physical robots using iRobot Create2s, Turtlebot2s and Turtlebot 3s in 2D; and Crazyflie 2.0s in 3D.
We use a VICON motion tracking system for localization.
Robots do not sense, but receive the position of others using the VICON system.
iRobot Create2s, Turtlebot3s, and Turtlebot2s are equipped with ODROID XU4 and ODROID C1+ single board computers running ROS Kinetic on the Ubuntu 16.04 operating system.
In all cases the algorithm is run on a centralized base station computer using separate processes for each robot.
Therefore, unlike simulations, planning is not synchronized between robots on real robot implementations.
RLSS does not result in any deadlocks in collisions in these asynchronized deployments as well.
Commands are sent to 2D robots over a WiFi network, and to Crazyflie 2.0s directly over their custom radio.

We conduct external disturbance experiments with 2 iRobot Create2s, 3 Turtlebot3s, and 2 Turtlebot2s (Fig.~\ref{fig:phy-dd-2}).
A human changes the positions of some robots by moving them arbitrarily during execution several times.
In all cases, robots replan in real-time and avoid each other successfully.

We demonstrate the algorithm in 3D using 6 Crazyflie 2.0s.
We conduct an experiment without obstacles in which Crazyflies swap positions with straight lines as desired trajectories (Fig.~\ref{fig:phy-cf-noobs-1}).
In another experiment, we show that Crazyflies can navigate through an environment with obstacles (\Cref{fig:phy-cf-obs-1,fig:phy-cf-obs-2}).
In each case, we externally disturb the Crazyflies and show that they can replan in real-time.

The recordings for our physical robot experiments are included in the supplemental video\footnote{Since we define robots' goals as single points, i.e. sets of measure zero, in physical experiments, robots keep missing their goals slightly.
This results in a spinning behavior in 2D since robots continuously fix their positions by replanning.}.

\section{Conclusion}
% random place: perfect sensing is impossible, we can increase safety distances or inflate objects, future work?
In this article, we present RLSS, a real-time decentralized long horizon trajectory planning algorithm for the navigation of multiple robots in shared environments with obstacles that provides guarantees on collision avoidance if the resulting problems are feasible. 
The generated optimization problem to compute a smooth trajectory is convex and kinematically feasible.
It does not require any communication between robots, requires only position sensing of obstacles and robots in the environment, and robots to be able to distinguish other robots from obstacles.
With its comparatively minimal sensing requirements and no reliance on communication, it presents a new baseline for algorithms that require communication and sensing/prediction of higher order state components of other robots. 
The algorithm considers the dynamic limits of the robots explicitly and enforces safety using hard constraints.

We show in synchronized simulation that RLSS performs better than two state-of-the-art planning algorithms (eBVC and RSFC), one of which requires velocity sensing on top of position sensing, in environments with obstacles in terms of number of deadlocks and number of colliding robots.
In our experiments, RLSS does not result in any deadlocks or collisions, while eBVC suffers from deadlocks and RSFC results in collisions (while RLSS provides theoretical guarantees on collision avoidance when it succeeds, it does not provide theoretical guarantees on deadlock avoidance).
When there are no obstacles in the environment, RSFC and eBVC outperform RLSS in terms of average navigation duration by $\SIrange{7}{20}{\%}$.

In future work, we would like to extend our planner to  asynchronously planning robots where each robot starts and finishes planning at different unknown timestamps by utilizing communication.
Also, we would like to incorporate the noise in the sensing systems within our algorithm in order to solve the cases with imperfect sensing in a principled way.

\begin{appendices}

\section{Sweep of a Convex Shape Along a Line Segment is Convex}\label{Appendix:SweepConvex}

\begin{lemma}\label{Lemma:CvxSweep}
Let $\mR(\vx) = \{\vx\} \oplus \mR_0$ be the Minkowski sum of $\{\vx\} \in \mathbb R^d$ and $\mR_0 \subseteq \mathbb R^d$, and let $\mR_0$ be a convex set.
Let $\vp(t) = \va + t(\vb-\va)$ where $t\in[0,1]$ be the line segment from $\va\in\mathbb R^d$ to $\vb\in \mathbb R^d$.
Then, the swept volume of $\mR$ along $\vp(t)$ is convex.
\begin{proof}
The swept volume $\mathcal{\zeta}$ of $\mR$ from $\va$ to $\vb$ can be defined as $\mathcal{\zeta} = \cup_{t=0}^1\mR(\vp(t)) = \cup_{t=0}^1\{\vp(t)\} \oplus \mR_0$.
Choose two points $\vq_1, \vq_2 \in \mathcal{\zeta}$.
$\exists t_1\in[0,1] \exists \tilde{\vq}_1\in \mR_0\ \vq_1 = \vp(t_1) + \tilde{\vq_1}$ and $\exists t_2\in[0,1] \exists\tilde{\vq}_2\in \mR_0\ \vq_2 = \vp(t_2) + \tilde{\vq_2}$. Let $\vq' = (1-t')\vq_1 + t'\vq_2$ be a point on the line segment connecting $\vq_1$ and $\vq_2$ where $t'\in[0,1]$.
\begin{align*}
\vq' &= (1-t')(\vp(t_1) + \tilde{\vq}_1) + t'(\vp(t_2) + \tilde{\vq}_2)\\
&= (1-t')(\va + t_1(\vb - \va)+ \tilde{\vq}_1) \\
&\ \ + t'(\va + t_2(\vb - \va) + \tilde{\vq}_2)\\
&= (1-t')\tilde{\vq}_1 + t'\tilde{\vq}_2 \\
&\ \ + \va + (t_1(1-t')+t_2t')(\vb-\va)
\end{align*}
$(t_1(1-t')+t_2t')\in[0,1]$ because it is a convex combination of $t_1,t_2\in[0,1]$.
Therefore $\va + (t_1(1-t')+t_2t')(\vb-\va) \in \vp(t)$.
Also, $(1-t')\tilde{\vq}_1 + t'\tilde{\vq}_2 \in \mR_0$, because it is a convex combination of $\tilde{\vq}_1,\tilde{\vq}_2\in\mR_0$ and $\mR_0$ is convex.
Therefore, $\vq'\in\mathcal{\zeta}$ since $\exists t\in[0,1]\ \vq' \in \{\vp(t)\}\oplus \mR_0$.
Hence, $\mathcal{\zeta}$ is convex.
\end{proof}
\end{lemma}

\end{appendices}

\section*{Compliance with Ethical Standards}
% \subsection*{Disclosure of a Potential Conflict of Interest}

\textbf{Disclosure of a potential conflict of interest.} Bask{\i}n \c{S}enba\c{s}lar, the first author, has moved to Prof. Gaurav S. Sukhatme's group in January 2022 after the initial submission of this paper.
Prof. Sukhatme is the editor-in-chief of Autonomous Robots journal and the advisor of Bask{\i}n \c{S}enba\c{s}lar at the time of writing.
Prof. Sukhatme was not involved in ideation, writing and experiments of this paper in any way.
This paper is submitted to \emph{Robot Swarms in the Real World: from Design to Deployment} special issue of Autonomous Robots, which has Dr. Siddharth Mayya as the lead guest editor.

\bibliography{bibliography}% common bib file


\end{document}